\documentclass[10pt,reqno]{article}
\usepackage{arxiv}

\usepackage{microtype}
\usepackage{graphicx}
\usepackage{tcolorbox}
\usepackage{subcaption}
\usepackage{booktabs}
\usepackage{multirow}
\usepackage{makecell}
\usepackage{hyperref}
\usepackage{amsmath}
\usepackage{amssymb}
\usepackage{mathtools}
\usepackage{amsthm}
\usepackage{authblk}

\newcommand\norm[1]{\|#1\|}
\newcommand{\abs}[1]{\left\lvert#1\right\rvert}

\theoremstyle{plain}
\newtheorem{theorem}{Theorem}[section]
\newtheorem{proposition}[theorem]{Proposition}
\newtheorem{lemma}[theorem]{Lemma}
\newtheorem{remark}[theorem]{Remark}

\newtheorem{corollary}[theorem]{Corollary}
\theoremstyle{definition}

\theoremstyle{remark}

\definecolor{tumblue}{RGB}{0, 101, 189}
\hypersetup{
   colorlinks=true,
   citecolor=tumblue,
   linkcolor=tumblue,
   filecolor=tumblue,
   urlcolor=tumblue,
}

\title{LieAugmenter: Equivariant Learning by \\
Discovering Symmetries with Learnable Augmentations}

\author[1,2]{Eduardo Santos-Escriche}
\author[1,2]{Ya-Wei Eileen Lin}
\author[1,2,3,4]{Stefanie Jegelka}

\affil[1]{Technical University of Munich (School of CIT)}
\affil[2]{Munich Center for Machine Learning}
\affil[3]{Munich Data Science Institute}
\affil[4]{MIT (Department of EECS and CSAIL)}

\begin{document}

\maketitle

\begin{abstract}
Data augmentation is a powerful mechanism in equivariant machine learning, encouraging symmetry by training networks to produce consistent outputs under transformed inputs.
Yet, effective augmentation typically requires the underlying symmetry to be specified \emph{a priori}, which can limit generalization when symmetries are unknown or only approximately valid.
To address this, we introduce \emph{LieAugmenter}, an end-to-end framework that discovers task-relevant continuous symmetries through \emph{learnable augmentations}.
Specifically, the augmentation generator is parameterized using the theory of Lie groups and trained jointly with the prediction network using the augmented views. 
The learned augmentations are task-adaptive,  enabling effective and interpretable symmetry discovery.
We provide a theoretical analysis of identifiability and show that our method yields symmetry-respecting models for the identified groups. 
Empirically, LieAugmenter outperforms baselines on image classification, as well as on the prediction of $N$-body dynamics and molecular properties. 
In addition, it can also provide an interpretable signature for detecting the absence of symmetries.
\end{abstract}

\section{Introduction}

Equivariant machine learning \citep{bronstein2017geometric, villar2021scalars, van2020mdp} has emerged as a principled approach for incorporating known symmetries into learning systems, leading to improved sample efficiency \citep{wang2022surprising, tahmasebi2023sample},  generalization \citep{elesedy2021provably, sokolic2017generalization}, and interpretability \citep{crabbe2023evaluating, wang2021self} across a wide range of domains \citep{bronstein2021geometric, kondor2018generalization}.
When the target function respects a symmetry, e.g., translation equivariance in vision \citep{lecun1989backpropagation, worrall2017harmonic}, rotational equivariance in molecular modeling \citep{batzner20223, schutt2021equivariant}, or Euclidean symmetry in physical systems \citep{smidt2021finding}, explicitly encoding the symmetry can substantially reduce the effective hypothesis space \citep{cohenGroupEquivariantConvolutional2016}. 
Among the various paradigms for incorporating symmetry, \emph{data augmentation} \citep{shorten2019survey, mumuni2022data} is arguably the most widely used in practice: it is architecture-agnostic, straightforward to integrate into existing pipelines, and naturally accommodates \emph{approximate} symmetries \citep{van2022relaxing, petrache2023approximation, tahmasebi2025achieving}.
Yet, similar to equivariant architectures which impose hard equivariance by design \citep{maron2018invariant, maron2019universality, weiler2018learning,  liao2023equiformer}, 
augmentation-based approaches typically require the symmetry to be specified a priori to be effective.
In practice, the relevant symmetry is often unknown, partially present, or holds only approximately \citep{gross1996role, cubuk2018autoaugment, davies2016computational, mitra2006partial, mitra2013symmetry,  liu2022machine}.

\begin{figure}[t]
  \centering
  \includegraphics[width=0.6\linewidth]{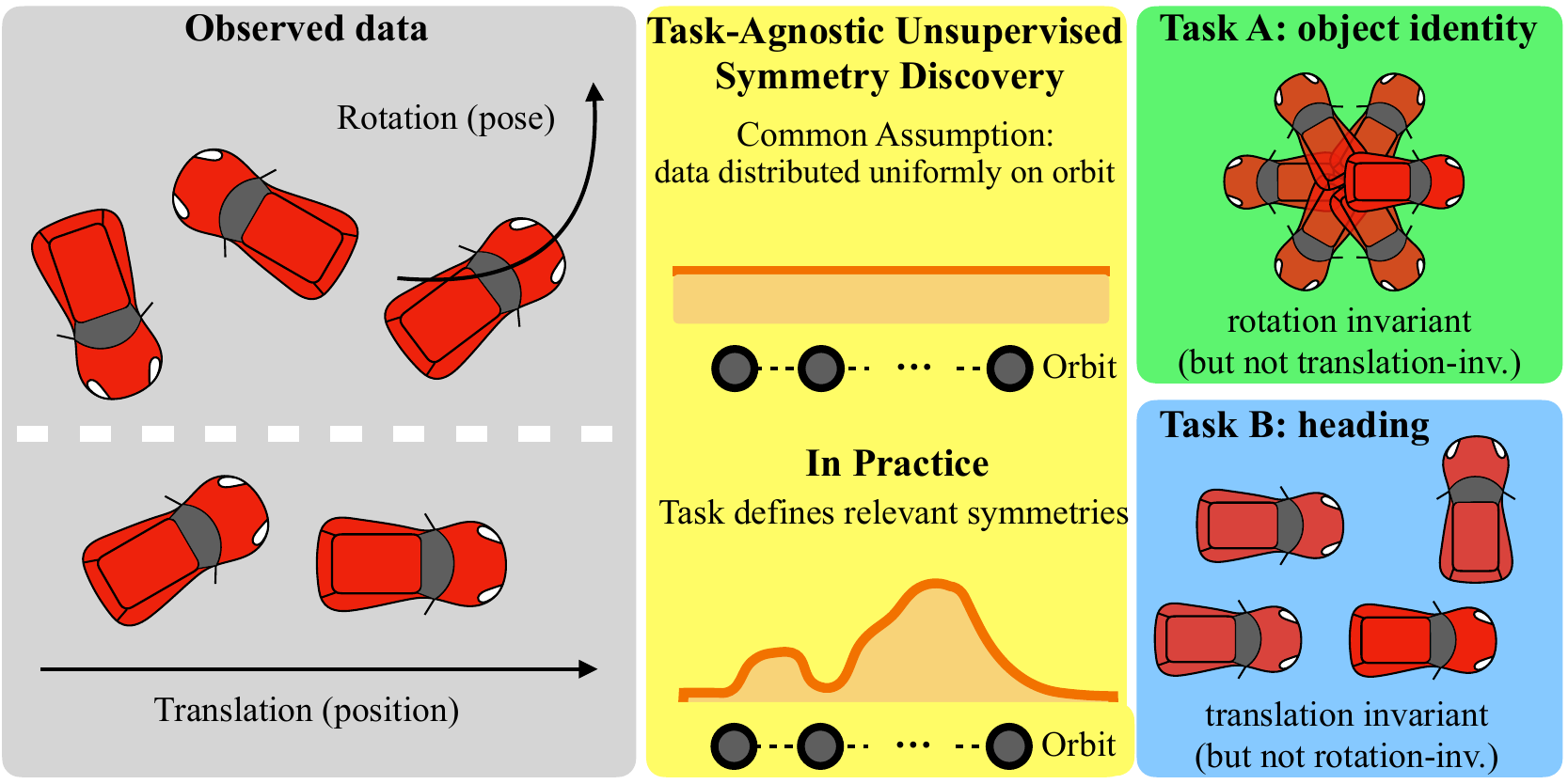}
  \caption{
Data exhibits latent transformations, e.g., rotation and translation, but which of these transformations should be treated as symmetries is often unknown. Symmetry discovery infers this. Existing unsupervised symmetry discovery methods can have two drawbacks: (i) they often assume the data is uniformly distributed across an orbit; this is not always true in practice \citep{lawrence2025augment}; (ii) symmetries can differ across target tasks.
  }
  \label{fig:task_dependend_sym}
\end{figure}

These challenges motivate \emph{symmetry discovery}: learning transformations from data rather than specifying them by hand \citep{zhou2020meta, bentonLearningInvariancesNeural2020, yang2023latent, dehmamy2021automatic, ko2024learning}. 
A growing line of work explores unsupervised or weakly supervised discovery of continuous symmetries by learning latent group actions, e.g., 
via Lie algebra-based architectures, adversarial distribution matching in data space, latent-space formulations for nonlinear actions, or post-hoc extraction from trained predictors \citep{yangGenerativeAdversarialSymmetry2023, moskalev2022liegg, hu2025symmetry, shaw2024symmetry, krippendorf2020detecting, van2023learning}.
Yet, two limitations commonly arise.
First, many discovery objectives target distributional symmetry, encouraging transformed samples to match the observed input distribution  \citep{yangGenerativeAdversarialSymmetry2023, desaiSymmetryGANSymmetryDiscovery2022, hu2025symmetry}. 
Prior methods find symmetries by assuming that the data distribution is invariant under a transformation (i.e., uniform over group orbits). However, this criterion can fail to capture symmetries that are relevant for the downstream task, as symmetries can be task-dependent (see Figure~\ref{fig:task_dependend_sym}), and imposing the wrong symmetry can degrade performance \citep{bentonLearningInvariancesNeural2020, rommel2022deep, miao2022learning}.
Second, many symmetry discovery methods are used post hoc or as a separate preprocessing step: a symmetry (or Lie algebra) is estimated from unlabeled data or a trained model, and then a second model is trained or redesigned to exploit it \citep{moskalev2022liegg, yangGenerativeAdversarialSymmetry2023, hu2025symmetry}.
Such separation can complicate model selection and prevents task supervision from refining the symmetry estimate end-to-end.
Hence, this paper seeks to address the following question:

\begin{tcolorbox}
Can we train an end-to-end unconstrained architecture
to jointly achieve both interpretable symmetry discovery and approximately
equivariant predictions?
\end{tcolorbox}

\begin{figure*}[t]
  \centering
  \includegraphics[width=1\linewidth]{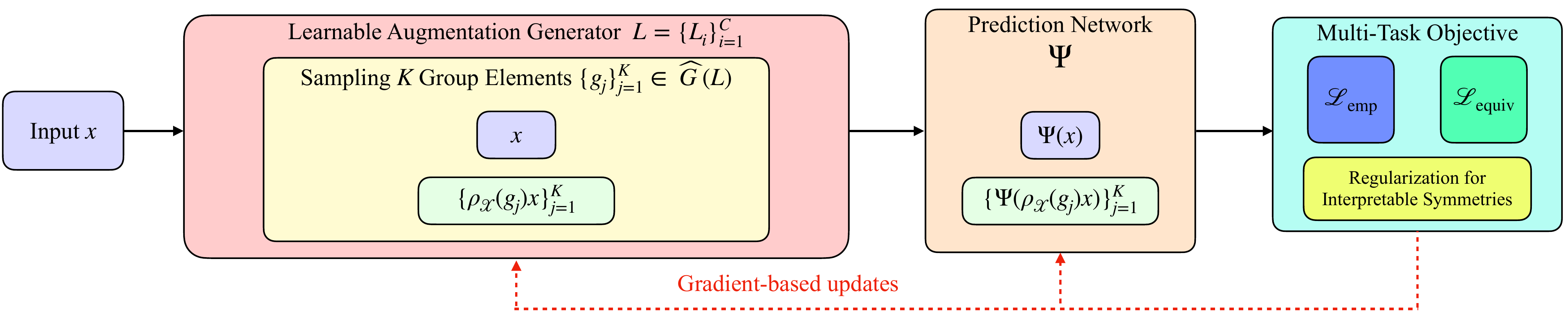}
  \caption{
  Illustration of LieAugmenter for task-dependent soft-equivariant models. 
LieAugmenter performs end-to-end task-dependent symmetry discovery by learning augmentation generators $L=\{L_i\}$ used to sample transformations $g_j\in \widehat{G}(L)$.
Sampling $K$ group elements $\{g_j\}_{j=1}^K$ induces $K$ learned augmentations $\{\rho_{\mathcal X}(g_j)x\}_{j=1}^K$, which are passed to the prediction network $\Psi$.
During training, $\Psi$ and $L$ are optimized jointly using the supervised task loss and the equivariance loss computed on the learned augmentations. 
  }
  \label{fig:semola}
\end{figure*}

\paragraph{Our Approach: Task-Adaptive Symmetry Discovery via Learnable Lie Augmentations.}
We propose \emph{LieAugmenter}, an end-to-end framework that formulates symmetry discovery as learning task-dependent data augmentations over a continuous transformation family.
LieAugmenter parameterizes candidate symmetries using a set of learnable Lie algebra generators.
During training, we sample group elements and apply the corresponding transformations to inputs.
We jointly optimize (i) the parameters of the augmentation generator and (ii) a symmetry-agnostic prediction network that receives the augmented data to learn equivariant predictions. The multi-task loss includes a supervised prediction loss and an equivariance loss that encourages equivariance under sampled transformations.
Figure~\ref{fig:semola} illustrates our model.
It is worth noting that rather than imposing equivariance architecturally, LieAugmenter promotes approximate equivariance through the end-to-end loss, allowing the learned transformations to adapt to the task while yielding an interpretable Lie algebra basis for the discovered symmetry.

We theoretically characterize the symmetry discovery mechanism by establishing an identifiability result for learned transformations. We also show that the approximate learned equivariance improves robustness under transformation-shifted test distributions. 
Empirically, we demonstrate the effectiveness of our approach on image classification, $N$-body dynamics prediction, and molecular property prediction. 
LieAugmenter learns interpretable symmetries \emph{end-to-end} and achieves strong predictive performance compared to oracle augmentation and two-stage symmetry discovery baselines.
Moreover, our method can detect the absence of symmetry for datasets with no nontrivial continuous symmetry and provides an interpretable \emph{no-symmetry} signature.

\section{Related Work}\label{sec:related_work}

We briefly review the most relevant Lie-based symmetry discovery 
methods related to our approach. For a more comprehensive and detailed discussion of related literature, see Appendix~\ref{app:Additional_RelatedWork}.
Augerino \citep{bentonLearningInvariancesNeural2020} and LieGAN \citep{yangGenerativeAdversarialSymmetry2023} are the most related symmetry discovery methods to ours
since they also parameterize learnable continuous symmetries with Lie algebra bases.
Different from LieAugmenter, Augerino is a supervised method solely based on the task loss, and the discovered symmetry is restricted to a subgroup of a pre-specified group.
To avoid having to pre-specify this group, the Augerino+ variant \citep{yangGenerativeAdversarialSymmetry2023} fixes the search space to the general linear group.
LieGAN learns fully unknown groups via generative adversarial training on the input distribution. 
However, it is an unsupervised symmetry discovery approach that is task-agnostic. 
It discovers symmetries as transformations that preserve the data distribution, making its performance sensitive to symmetry acting uniformly on the data \citep{hu2025symmetry}, and hindering its adaptability to different tasks. 
In addition, to use the learned symmetry in a task, an additional downstream model is needed.
Our method, in contrast to these approaches, is end-to-end and it incorporates an equivariance loss in the objective to promote task-adaptive approximate symmetries.
\section{Preliminaries}\label{sec:background}

We briefly review relevant background. For additional details and used notation, we refer to Appendix~\ref{app:Additional_Background}. 

\paragraph{Matrix Lie Group and Lie Algebras.}
A matrix Lie group $G$ is a subgroup of the general linear group $\mathrm{GL}(d, \mathbb{R})$ that is also a smooth manifold s.t. multiplication and inversion are smooth maps \citep{baker2003matrix, erdmann2006introduction}. 
In this paper, we primarily consider connected matrix Lie groups.
The associated Lie algebra $\mathfrak{g} = T_{\mathrm{Id}} G$ is the tangent space of $G$ at the identity element. 
Each element $A\in \mathfrak{g}$ is an infinitesimal generator of one-parameter subgroups $t \mapsto \exp(tA)$ for $t\in \mathbb{R}$. 
Let $\{L_i\}_{i=1}^{\dim \mathfrak{g}}$ be a basis of $\mathfrak{g}$.
For matrix Lie groups, the exponential map $\exp:\mathfrak{g}\to G$ is given by the matrix exponential, which is a local diffeomorphism in a neighborhood of the identity.
If $G$ is connected and compact,  $\exp$ is surjective \citep{kirillov2008introduction, hall2013lie}, hence every $g\in G$ can be written as $g=\exp(\sum_i w_i L_i)$ with $w_i\in\mathbb{R}$. 

\paragraph{Group Actions and Equivariance.}
We consider inputs $x\in\mathcal{X}\subseteq\mathbb{R}^n$ and outputs $y\in\mathcal{Y}\subseteq\mathbb{R}^m$. 
A linear action of $G$ on $\mathcal{X}$ is specified by a representation $\rho_{\mathcal{X}}:G\to\mathrm{GL}(n)$, and similarly on $\mathcal{Y}$ by $\rho_{\mathcal{Y}}:G\to\mathrm{GL}(m)$. 
A function $f:\mathcal{X}\to\mathcal{Y}$ is $G$-equivariant if $f(\rho_{\mathcal{X}}(g)x) = \rho_{\mathcal{Y}}(g) f(x)$ for all $g\in G $ and $x\in \mathcal{X}$.
Invariance is a special case of equivariance,  where the output action is $\rho_{\mathcal{Y}}(g)=I\;\forall g\in G$. 
\section{LieAugmenter}\label{sec:LieAugmenter}

In this section, we introduce \emph{LieAugmenter} for jointly discovering continuous symmetries and training a prediction network using learnable Lie group augmentations under a multi-task objective (see Figure~\ref{fig:semola} for an illustration). 

\paragraph{Problem Setting.}
We consider a supervised learning setting with a labeled dataset $\mathcal{D} = \{(x_i, y_i)\}_{i=1}^N$,  where $x_i \in \mathcal{X} \subseteq \mathbb{R}^n$ and $\ y_i \in \mathcal{Y} \subseteq \mathbb{R}^m$, and aim to learn a predictor $f : \mathcal{X} \rightarrow \mathcal{Y}$.
We assume the data exhibits an unknown \emph{continuous} symmetry modeled by a connected matrix Lie group $G$ with linear representations $\rho_{\mathcal{X}}:G\to \mathrm{GL}(n)$ and $\rho_{\mathcal{Y}}:G\to \mathrm{GL}(m)$ (see Appendix~\ref{app:recoverable_sym_group} for examples of such groups). 
Our objective is two-fold: (i) \emph{symmetry discovery}, i.e., to identify an interpretable representation of the latent symmetry via its associated Lie algebra basis, and (ii) \emph{equivariant prediction}, i.e., to learn a function $f$ with strong generalization performance and small equivariance error with respect to the discovered symmetry:
\begin{equation*}
 f(\rho_{\mathcal{X}}(g) x) \approx \rho_{\mathcal{Y}}(g) f(x) \text{ for all }   g\in G \text{ and } x\in \mathcal{X}.
\end{equation*}

\subsection{Parameterizing Augmentations using Lie Groups}\label{sec:Parameterizing_Augmentations_using_Lie_Groups}

Central to our model is a learnable data augmentation procedure that learns a Lie algebra basis. 
For each input $x_i\in \mathcal{X}$ we sample a set of group elements based on the learned symmetry and generate the corresponding transformed inputs, which are then fed to the predictor.
To learn symmetry group $G$ from data, we need a differentiable parameterization of group elements that supports gradient-based optimization through the sampling procedure.
We thus adopt a reparameterization approach \citep{price2003useful, kingma2013auto} adapted to Lie groups \citep{falorsi2019reparameterizing}, which samples group elements through the associated Lie algebra. 

Let $\mathfrak{g}$ denote the Lie algebra associated with a connected matrix Lie group acting on the representation space of $\mathcal{X}$.
We parameterize a candidate Lie algebra using a learnable set of $C$ generators $L = \{L_i \in \mathbb{R}^{d \times d}\}_{i=1}^C$
\footnote{For vector-valued data, we take $d=n$ and learn generators in the ambient data representation, i.e., $L_i\in\mathbb{R}^{n\times n}$. For images, we parameterize transformations in coordinate space (e.g., homogeneous $d=3$), learn $L_i\in\mathbb{R}^{d\times d}$, and realize the induced action $\rho_{\mathcal{X}}(g)$ via differentiable grid warping (see Appendix~\ref{app:Generator_Dimension}).}
and sample coefficients $w=(w_1,\ldots,w_C)\in\mathbb{R}^C$ from a distribution $P(\gamma)$ with parameters $\gamma$.
Specifically, we consider a uniform distribution $P(\gamma) = \mathcal{U}[-\gamma, \gamma]$.
A group element is obtained by mapping the resulting Lie algebra element to the group via the matrix exponential, given by 
\begin{equation}
    g = \exp \left[\sum\nolimits_{i=1}^C w_i L_i \right], \  w_i \sim P(\gamma).
    \label{eq:group_sampling}
\end{equation}
By construction, $g$ lies in the matrix Lie subgroup $\widehat{G}(L)$ generated by $L$ (and $\widehat{G}(L)$ coincides with $G$ when the learned generators recover the true Lie algebra). The reparameterized sampling allows gradients to propagate to $L$.

Given an input $x_i$, we draw $K$ independent group elements $g_{i,2},\ldots,g_{i,K+1}$ according to Eq.~\eqref{eq:group_sampling} and form the corresponding augmentations using their representations
\begin{equation}
    x_{i,j} := \rho_{\mathcal{X}}(g_{i,j})\,x_i, \qquad  j \in [2, \ldots, K+1], 
\end{equation}
with the original input $x_{i,1}:=x_i$.
We then evaluate a symmetry-agnostic network $\Psi$ on $\{x_{i,j}\}_{j=1}^{K+1}$ to produce predictions for the original and transformed inputs, which are used by the training objective in the next subsection.

\subsection{Multi-Task Objective for Symmetry Discovery}

We now introduce the end-to-end training objective used to learn both the task network and the augmenter.
Specifically, we jointly learn (i) a \emph{symmetry-agnostic} prediction network $\Psi$ and (ii) the parameters of the learned augmentation generators.
LieAugmenter does not impose the equivariance architecturally; instead, the learned augmentation generators adapt to the task and the prediction model learns symmetry through a training objective that encourages equivariance of the task network $\Psi$ across sampled group transformations.

Our objective consists of two primary terms. 
The empirical prediction loss measures standard supervised performance on the original inputs, while the equivariance loss promotes symmetry of predictions under transformations sampled from the learned group.
In particular, if a sampled transformation does not reflect a valid symmetry of the prediction task, then enforcing equivariance under it leads to discrepancy between predictions and transformed targets, which increases the loss and steers the learned generators away from such transformations.
We detail these terms below.

\paragraph{Empirical Prediction Loss.} 
We define the empirical prediction loss on the original, untransformed inputs $x_{i,1}=x_i$.
For regression tasks, we use the mean squared error
\begin{equation}
    \mathcal{L}_{\mathrm{emp}} = \frac{1}{N}\sum_{i=1}^{N} \norm{y_i - \Psi(x_{i,1})}_2^2,
\end{equation} 
and for classification tasks we use the cross-entropy. 

\paragraph{Equivariance Loss.}
To encourage approximate equivariance w.r.t the sampled transformations, we define an equivariance loss that compares predictions on transformed inputs with transformed targets.
Let $g_{i,2},\ldots,g_{i,K+1}$ denote the $K$ group elements sampled for $x_i$. For regression tasks, we set 
\begin{equation}
\label{eq:Equivariance loss}
    \mathcal{L}_{\mathrm{equiv}}  =  \frac{1}{N}\sum_{i=1}^{N}\frac{1}{K}\sum_{j=2}^{K+1} \norm{\Psi (\rho_{\mathcal{X}}(g_{i,j}) x_i)  -  \rho_{\mathcal{Y}}(g_{i,j})y_i}_1, 
\end{equation}
and for classification tasks we use cross-entropy.

The equivariance loss in Eq.~\eqref{eq:Equivariance loss} links symmetry discovery and prediction: if the learned transformations are incompatible with the data, enforcing consistency across them increases the loss, which in turn discourages incorrect symmetry estimates.
In addition, this term can be viewed as a Monte Carlo estimate \citep{robert1999monte} of the expected equivariance violation under the learned sampling distribution over $g$. 
It is minimized when $\Psi(\rho_{\mathcal{X}}(g)x)\approx \rho_{\mathcal{Y}}(g)y$ for transformations supported by the data.

\paragraph{Main Objective.}
We balance predictive accuracy and equivariance with weights $\alpha,\beta \ge 0$:
\begin{equation} \label{eq:main_loss}
    \mathcal{L}_{\mathrm{main}} = \alpha\mathcal{L}_{\mathrm{emp}}
    + \beta \mathcal{L}_{\mathrm{equiv}}.
\end{equation}
The coefficient $\beta$ controls the equivariance constraint, which is important when the underlying symmetry is approximate.

\paragraph{Regularization for Interpretable Symmetries.} In addition to $\mathcal{L}_{\mathrm{main}}$, we include regularizers that promote non-trivial and interpretable symmetry representations. First, we discourage near-identity augmentations by penalizing similarity between an input and its transformed variants:
\begin{equation}
    \ell_{\mathrm{areg}} = \frac{1}{N}\sum_{i=1}^{N}\frac{1}{K}\sum_{j=2}^{K+1}
    \lvert \mathrm{CosSim}\left(x_i,\ \rho_{\mathcal{X}}(g_{i,j})\,x_i\right) \rvert.
\end{equation}
Second, we encourage sparse generators: 
\begin{equation}
    \ell_{\mathrm{bsreg}} = \sum\nolimits_{i=1}^{C} \norm{L_i}_1. 
\end{equation}
Finally, we promote orthogonality among generators by discouraging highly aligned basis elements:
\begin{equation}
    \ell_{\mathrm{bcreg}} = \sum\nolimits_{i=1}^C \sum\nolimits_{j=i+1}^C \lvert \mathrm{CosSim} (L_i, L_j) \rvert.
\end{equation}

\paragraph{Total Training Objective.}
The final objective is the weighted sum of the main loss and the regularization terms:
\begin{equation}
     \mathcal{L}_{\mathrm{total}}
   = 
    \alpha\mathcal{L}_{\mathrm{emp}}
     + 
    \beta\mathcal{L}_{\mathrm{equiv}}
    + 
    \lambda\ell_{\mathrm{areg}}
    + 
    \nu\ell_{\mathrm{bsreg}}
     +
    \eta\ell_{\mathrm{bcreg}},
    \label{eq:total_loss}
\end{equation}
where $\alpha,\beta,\lambda,\eta,\nu \ge  0$ control the contribution of each term. 

We conclude this section with two remarks. 
First, the multi-task objective in Eq.~\eqref{eq:total_loss} is designed to prevent degenerate solutions while promoting an interpretable, non-redundant basis (see Appendix~\ref{app:proof_identifiability}).
In Appendix~\ref{app:ablation}, we conduct an ablation study and sensitivity analysis over regularizations and hyperparameters, and find that our method is robust across a broad range of settings.
Second, in line with prior Lie group-based symmetry discovery methods \citep{bentonLearningInvariancesNeural2020, yang2023latent, yangGenerativeAdversarialSymmetry2023, moskalev2022liegg}, sampling via Eq.~\eqref{eq:group_sampling} is inherently limited  to the identity component of the Lie subgroup generated by the learned generators, and may not cover it fully when the exponential map is non-surjective (notably for many non-compact Lie groups) \citep{kirillov2008introduction, moskowitz2003exponential}.
Thus, disconnected symmetries (e.g., reflections) are not reachable without additional discrete structure. 
See Appendix~\ref{app:recoverable_sym_group} for further discussion of recoverable groups.
Nevertheless, Lie group actions in our method still capture a broad class of practical continuous symmetries across various application domains \citep{gross1996role, gilmore2006lie, thomas2018tensor, rao1998learning, finzi2020generalizing, falorsi2019reparameterizing, sola2018micro, kondor2018generalization, sohl2010unsupervised, cohen2014learning,  dehmamy2021automatic}.
Extending the framework to disconnected symmetries and to more fully address non-compact settings is an important direction for future work.
\section{Theoretical Properties of LieAugmenter}\label{sec:theoretical_results}

In this section, we present the theoretical properties of LieAugmenter.
We first state the Monte Carlo nature of the equivariance objective.
We then present the identifiability of the discovered symmetry. 
We further show that LieAugmenter leads to  approximately symmetry preserving networks with universal approximation properties and relate approximate equivariance to robustness under transformation shifts. 
Proofs and additional analysis are in Appendix~\ref{app:Theoretical_Analysis_of_LieAugmenter}. 

Let $Q_{L}$ be the distribution over group elements from Eq.~\eqref{eq:group_sampling} and let the data  $(X,Y)\sim P$.
We define the population analogues of the losses by  $\mathcal{L}_{\mathrm{emp}}^{\mathrm{pop}}(\Psi)= \mathbb{E}[\norm{Y-\Psi(X)}_2^2]$, $\mathcal{L}_{\mathrm{equiv}}^{\mathrm{pop}}(\Psi,L)=  \mathbb{E}[\mathbb{E}_{g\sim Q_{L}}
[\norm{\Psi(\rho_{\mathcal{X}}(g)X)-\rho_{\mathcal{Y}}(g)Y}_1]]$, and set $\mathcal{L}_{\mathrm{main}}^{\mathrm{pop}} =\alpha \mathcal{L}_{\mathrm{emp}}^{\mathrm{pop}}+\beta \mathcal{L}_{\mathrm{equiv}}^{\mathrm{pop}}$.

\begin{lemma}\label{lemma:monte_carlo}
    For fixed $(\Psi,L)$, the empirical equivariance term in Eq.~\eqref{eq:Equivariance loss} is a $K$-sample Monte Carlo estimator of $\mathcal{L}_{\mathrm{equiv}}^{\mathrm{pop}}(\Psi,L)$.
    Specifically, it is unbiased (up to dataset averaging) and its variance scales as $O(1/K)$.
\end{lemma}
Lemma~\ref{lemma:monte_carlo} formalizes that LieAugmenter minimizes an expected equivariance deviation under learned augmentations.
Concretely, for fixed augmenter parameters, 
$\mathcal{L}_{\mathrm{equiv}}$ averages the task loss over $K$ sampled augmentations. This corresponds to a partial augmentation scheme that approximates the group-averaging operator via Monte Carlo sampling. Empirically, we find that using $K=10$ attains strong performance (see Section~\ref{sec:experiments} and Appendix~\ref{app:additional_exp}).

\begin{theorem} \label{thm:identifiability}
    Let $\alpha, \beta>0$.
    Assume $Y=f^\star(X)$ almost surely and that $f^\star$ is the unique minimizer of $\mathcal{L}_{\mathrm{emp}}^{\mathrm{pop}}(\Psi)$ over the predictor class.
    In addition, suppose $\mathcal{L}_{\mathrm{equiv}}^{\mathrm{pop}}(f^\star,L)=0$.
    Let $\widehat{\mathfrak{g}}(L)=\mathrm{span}\{L_1,\dots,L_C\}$ and let $\widehat{G}(L)$ be the connected subgroup generated by $\widehat{\mathfrak{g}}(L)$.
    Suppose for every $g$ in a neighborhood of the identity in the candidate subgroup $\widehat{G}(L)$, $\rho_{\mathcal{X}}(g)\mathcal{X}\subseteq \mathcal{X}$. 
    Then any global minimizer $(\Psi^\star,L^\star)$ of $\mathcal{L}_{\mathrm{main}}^{\mathrm{pop}}(\Psi,L)$ satisfies
    $\Psi^\star=f^\star$ almost surely and $\widehat{G}(L^\star)\subseteq G$. 
    Moreover, if the connected symmetry group of $f^\star$ is exactly $G$ and $\dim(\widehat{G}(L^\star))=\dim(G)$, then $\widehat{G}(L^\star)=G$ and $\widehat{\mathfrak{g}}(L^\star)=\mathfrak{g}$. 
\end{theorem}

Theorem~\ref{thm:identifiability} explains the symmetry-discovery mechanism,  addressing (i) recovering the correct underlying function and (ii) recovering the symmetry.
With respect to discovered symmetries, the main objective admits multiple minimizers 
as any subgroup of $G$ yields zero equivariance loss when $\Psi=f^\star$. Therefore, the regularizers in Eq.~\eqref{eq:total_loss} are designed to break this degeneracy and promote an interpretable basis within the symmetry group.
We formalize the role of regularizers as a canonicalization principle in Appendix~\ref{app:proof_identifiability}. 

\begin{proposition} \label{prop:ua_main}
    Let $(X,Y)\sim P$ with $Y=f^\star(X)$ almost surely for a continuous target function $f^\star:\mathcal{X}\to\mathcal{Y}$ and let $f^\star$ be equivariant $ \forall g\in\mathrm{supp}(Q_{L})$. 
    Assume $\mathcal{X}$ is compact and that $\rho_{\mathcal{X}}(g)\mathcal{X}\subseteq \mathcal{X} \; \forall \; g\in \mathrm{supp}(Q_{L}), x\in \mathcal{X}$.
    Let the hypothesis class used for $\Psi$ be a universal approximator on $\mathcal{X}$.
    Then for any $\varepsilon>0$, there exists a network $\Psi$ such that 
    \begin{equation}
    \underset{x\in\mathcal{X}}{\sup}\abs{\Psi(x)-f^\star(x)}< \varepsilon \text{ and }\mathcal{L}_{\mathrm{equiv}}^{\mathrm{pop}}(\Psi,L)< \varepsilon.
    \end{equation}
\end{proposition}
Proposition~\ref{prop:ua_main} shows that LieAugmenter can be used with symmetryless networks while retaining universal approximation properties for equivariant targets. 
Specifically, the universal approximation properties for standard architectures \citep{LESHNO1993861, hornik1989multilayer, cybenko1989approximation} imply the existence of networks $\Psi$ without built-in equivariance that uniformly approximate an equivariant target $f^\star$. 
In our method, such approximations can be realized while simultaneously optimizing the induced equivariance deviation w.r.t. the learned transformations.

We now connect the approximate equivariance to out-of-distribution  (OOD) generalization under transformation shift.
We consider the test distribution to be obtained by applying a group element to both inputs and labels.

\begin{proposition}
    \label{prop:ood}
    For $g\in G$, define the transformation-shifted distribution $P^{(g)}$ by $(X',Y')=(\rho_{\mathcal{X}}(g)X,\rho_{\mathcal{Y}}(g)Y)$ with $(X,Y)\sim P$.
    Let $\ell:\mathcal{Y}\times\mathcal{Y}\to\mathbb{R}_+$ satisfy:
    (i) $\ell(\rho_{\mathcal{Y}}(g)a,\rho_{\mathcal{Y}}(g)b)=\ell(a,b) \; \forall  a,b\in\mathcal{Y}, g\in G$, and
    (ii) there exists $L_\ell>0$ s.t. $|\ell(u,y)-\ell(v,y)|\le L_\ell\norm{u-v}_1\; \forall \; u,v,y \in \mathcal{Y}$.
    Then for any network $\Psi$ and any $g\in G$, $\mathcal{R}_{P^{(g)}}(\Psi)
\le
\mathcal{R}_{P}(\Psi)
+
L_\ell\,\mathbb{E}[\norm{\Psi(\rho_{\mathcal{X}}(g)X)-\rho_{\mathcal{Y}}(g)\Psi(X)}_1]$, where $\mathcal{R}_{P}(\Psi):=\mathbb{E}
[\ell(\Psi(X),Y)]$.
\end{proposition}

Proposition~\ref{prop:ood} measures how stable the prediction network is under a transformation shift. 
Specifically, this bound is controlled by the objective optimized in LieAugmenter (see Appendix~\ref{app:proof_ood}).
We empirically show the advantages of our method under OOD settings in Section~\ref{sec:experiments} and Appendix~\ref{app:additional_exp}.
\section{Experiments}\label{sec:experiments}

We evaluate LieAugmenter on image classification, $N$-body dynamics prediction, and molecular property prediction.
We also examine our method on a dataset without symmetries as a control experiment.
Detailed implementation settings, including experimental setup and hyperparameters, are reported in Appendix~\ref{app:exp_details}. 
Further empirical results, including runtime analysis, ablations, studies on discovering discrete and partial symmetries, and on data from colorectal cancer detection and Lorentz symmetries in quantum field theory, are provided in Appendix~\ref{app:additional_exp}. 

\subsection{Image Classification}
\label{sec:rotated_mnist}

We first demonstrate the advantage of LieAugmenter for symmetry discovery and equivariant learning in the image domain using a CNN prediction network \citep{lecun1989backpropagation, krizhevsky2012imagenet}. 
We use the RotatedMNIST dataset \citep{deng2012mnist, larochelle2007empirical}, where each image is randomly rotated in-plane and the digit label is approximately invariant to in-plane rotations, i.e., $\rho_{\mathcal Y}(g) \equiv I$ for classification.
Here, we consider both in-distribution (ID) and out-of-distribution (OOD) settings. 
In the ID setting, both training and test images are rotated by angles $\theta \in [0, 360^\circ]$.
In the OOD setting, training images are rotated by $\theta \in [-90^\circ, 90^\circ)$ while test images are rotated by angles in the complementary set $\theta \in [90^\circ, 180^\circ)\cup[-180^\circ, -90^\circ)$. 
Note that from the symmetry discovery perspective, the OOD setting implies that the data has
a non-uniform distribution over the orbit, which is a more challenging scenario.

\begin{figure}[h]
  \centering
  \includegraphics[width=0.7\linewidth]{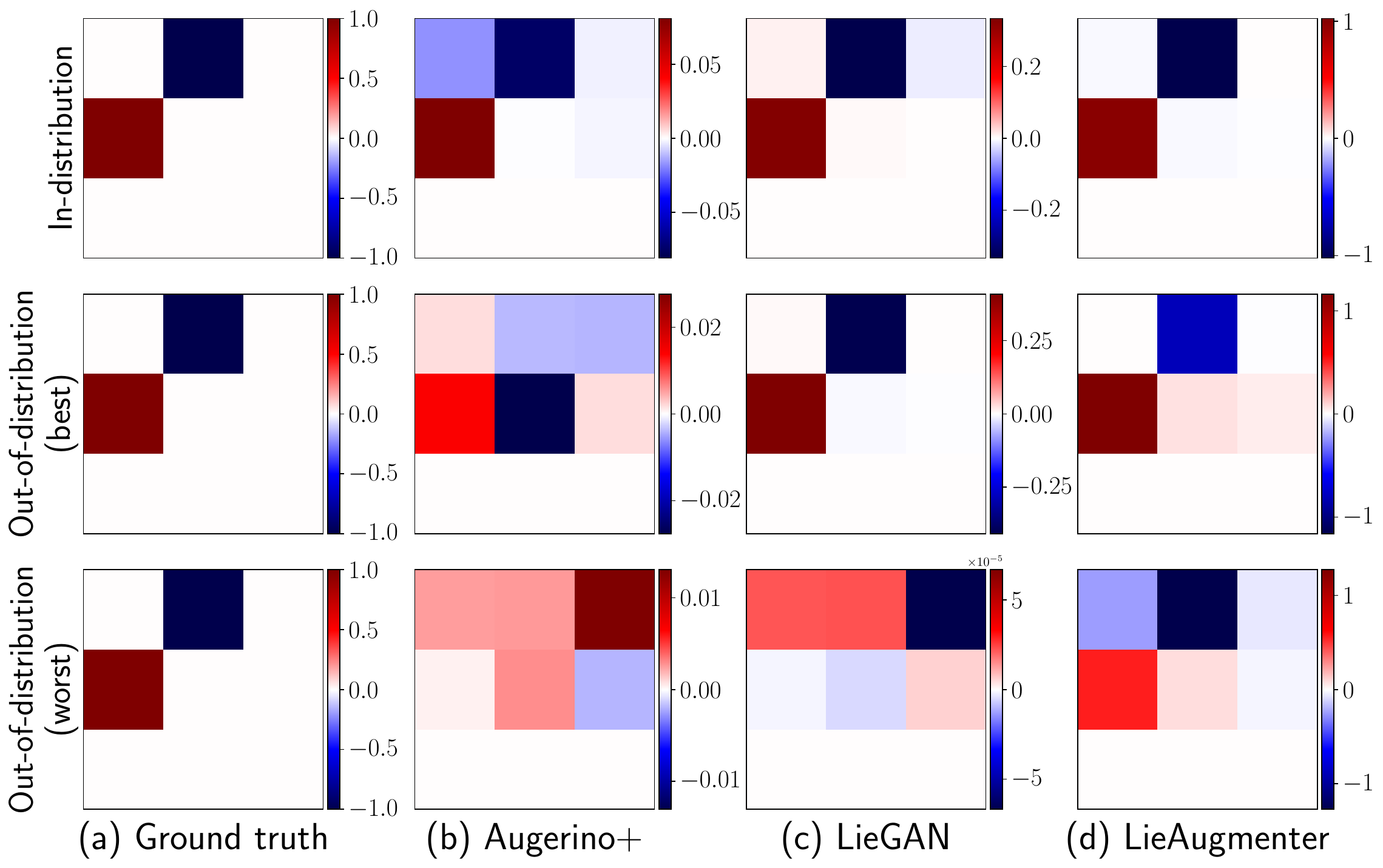}
  \caption{
  Lie algebra basis for the ground-truth $\mathrm{SO}(2)$ group and bases learned by baseline symmetry discovery methods and LieAugmenter on RotatedMNIST. Results are shown for the ID (\textit{top}) and OOD settings. For OOD, we report the best (\textit{middle}) and worst (\textit{bottom}) learned bases across three random seeds.
  }
  \label{fig:lie_algebras_rotmnist}
\end{figure}

\begin{figure}[h]
  \centering
  \includegraphics[width=0.6\linewidth]{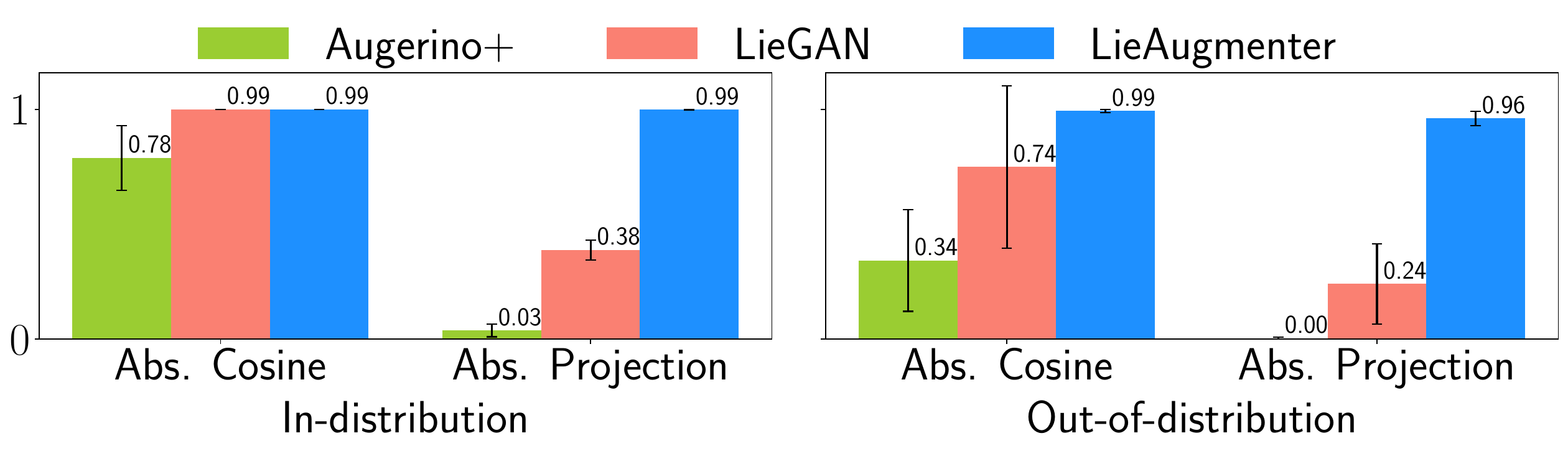}
  \caption{Absolute cosine similarity and absolute Frobenius projection for Lie algebra basis recovery on in-distribution and OOD RotatedMNIST ($\mathrm{mean}\pm\mathrm{std}$ over three random seeds).}
  \label{fig:lie_comparison_rotmnist}
\end{figure}

\paragraph{{Symmetry Discovery.}}
We compare LieAugmenter with LieGAN and Augerino+ \citep{yangGenerativeAdversarialSymmetry2023} for unknown symmetry discovery on RotatedMNIST.
Figure~\ref{fig:lie_algebras_rotmnist} shows visualizations of the Lie algebra bases learned by each method.
We see that LieAugmenter correctly recovers the Lie algebra basis of the ground-truth $\mathrm{SO}(2)$ group in both settings.
For a quantitative comparison, Figure~\ref{fig:lie_comparison_rotmnist} reports the absolute cosine similarity and the absolute Frobenius projection between the learned basis and the ground truth. 
In the ID setting, both LieGAN and LieAugmenter achieve near-perfect directional alignment. In the OOD setting, LieGAN shows degradation and higher variability across runs. 
In contrast, LieAugmenter remains well aligned and stable, indicating stronger robustness under the shifted distribution.

\paragraph{Predictive Performance.}
We evaluate the prediction performance by comparing LieAugmenter
with Augerino+ and the augmentation strategies applied to the base CNN prediction network, including (i) no augmentations, (ii) oracle augmentations using the ground truth $\mathrm{SO}(2)$ symmetry, and (iii) augmentations based on the symmetries discovered by LieGAN (denoted by LieGAN Aug).
We also include an equivariant architecture baseline, a Group Equivariant Convolutional Network (GCNN) \citep{cohenGroupEquivariantConvolutional2016}, which encodes invariance to the discrete $p4$ group through architectural constraints. 
Table~\ref{tab:results_rotmnist} reports test classification accuracy and the equivariance error.
We see that in the ID setting LieAugmenter outperforms the $p4$ equivariant architecture, oracle augmentation, and augmentations from discovered symmetries.
In the OOD setting, LieAugmenter matches oracle augmentation. 
Notably, LieAugmenter attains such strong accuracy and OOD generalization while discovering the symmetry jointly with the prediction network during end-to-end training, rather than relying on prior knowledge of the symmetry  (equivariant architectures or oracle augmentation) or a separate symmetry-discovery stage.

\begin{table*}[h]
\caption{RotatedMNIST results for in-distribution (ID) and out-of-distribution (OOD) evaluation. We report classification accuracy (higher is better) and equivariance error (lower is better) for an equivariant-architecture baseline (GCNN, $p4$) and for CNN models trained (i) without augmentation (\emph{Trivial}), (ii) with \emph{Oracle} rotation augmentation (ground-truth symmetry), (iii) with symmetry-discovery baselines (Augerino+ and LieGAN Aug.), and (iv) with LieAugmenter (\emph{Ours}). All values are reported as $\mathrm{mean}\pm\mathrm{std}$ over three random seeds.}
  \label{tab:results_rotmnist}
  \centering
  \resizebox{\textwidth}{!}{%
  \begin{tabular}{lcccccc}
    \toprule
    & \multicolumn{1}{c}{Equivariant architecture} & \multicolumn{1}{c}{No augmentation} & \multicolumn{1}{c}{Oracle augmentation} & \multicolumn{3}{c}{Symmetry discovery} \\
    \cmidrule(lr){2-2}\cmidrule(lr){3-3}\cmidrule(lr){4-4}\cmidrule(lr){5-7}
    & GCNN ($p4$) & Trivial & Rotation & Augerino+ & LieGAN Aug. & LieAugmenter \\
    \midrule
ID Acc. (\%) $(\uparrow)$ & $\text{96.78} \scriptstyle{\pm \text{0.36}}$ & $\text{96.17} \scriptstyle{\pm \text{0.34}}$ & $\underline{\text{99.06}} \scriptstyle{\pm \text{0.00}}$ & $\text{96.85} \scriptstyle{\pm \text{0.30}}$ & $\text{98.69} \scriptstyle{\pm \text{0.16}}$ & $\textbf{\text{99.08}} \scriptstyle{\pm \text{0.06}}$ \\
    ID Equiv. Error $(\downarrow)$ & $\text{5.31} \scriptstyle{\pm \text{1.16}}$ & $\text{5.52} \scriptstyle{\pm \text{0.23}}$ & $\underline{\text{4.01}} \scriptstyle{\pm \text{0.29}}$ & $\text{5.09} \scriptstyle{\pm \text{0.70}}$ & $\text{5.04} \scriptstyle{\pm \text{0.19}}$ & $\textbf{\text{3.10}} \scriptstyle{\pm \text{0.32}}$ \\
    \midrule
    OOD Acc. (\%) $(\uparrow)$ & $\text{55.60} \scriptstyle{\pm \text{1.45}}$ & $\text{55.24} \scriptstyle{\pm \text{1.14}}$ & $\textbf{\text{99.09}}\scriptstyle{\pm \text{0.08}}$ & $\text{57.68} \scriptstyle{\pm \text{1.45}}$ & $\underline{\text{91.59}} \scriptstyle{\pm \text{1.07}}$ & $\text{90.82} \scriptstyle{\pm \text{6.49}}$ \\
    OOD Equiv. Error $(\downarrow)$ & $\text{5.78} \scriptstyle{\pm \text{1.23}}$ & $\text{7.19} \scriptstyle{\pm \text{0.54}}$ & $\text{\underline{3.87}} \scriptstyle{\pm \text{0.35}}$ & $\text{3.97} \scriptstyle{\pm \text{0.51}}$ & $\text{5.82} \scriptstyle{\pm \text{0.47}}$ & $\textbf{\text{3.26}} \scriptstyle{\pm \text{0.62}}$ \\
    \bottomrule
  \end{tabular}
  }
\end{table*}

\subsection{$N$-Body Dynamics Prediction}
\label{sec:n_body}

\begin{figure}[h]
  \centering
  \includegraphics[width=0.8\linewidth]{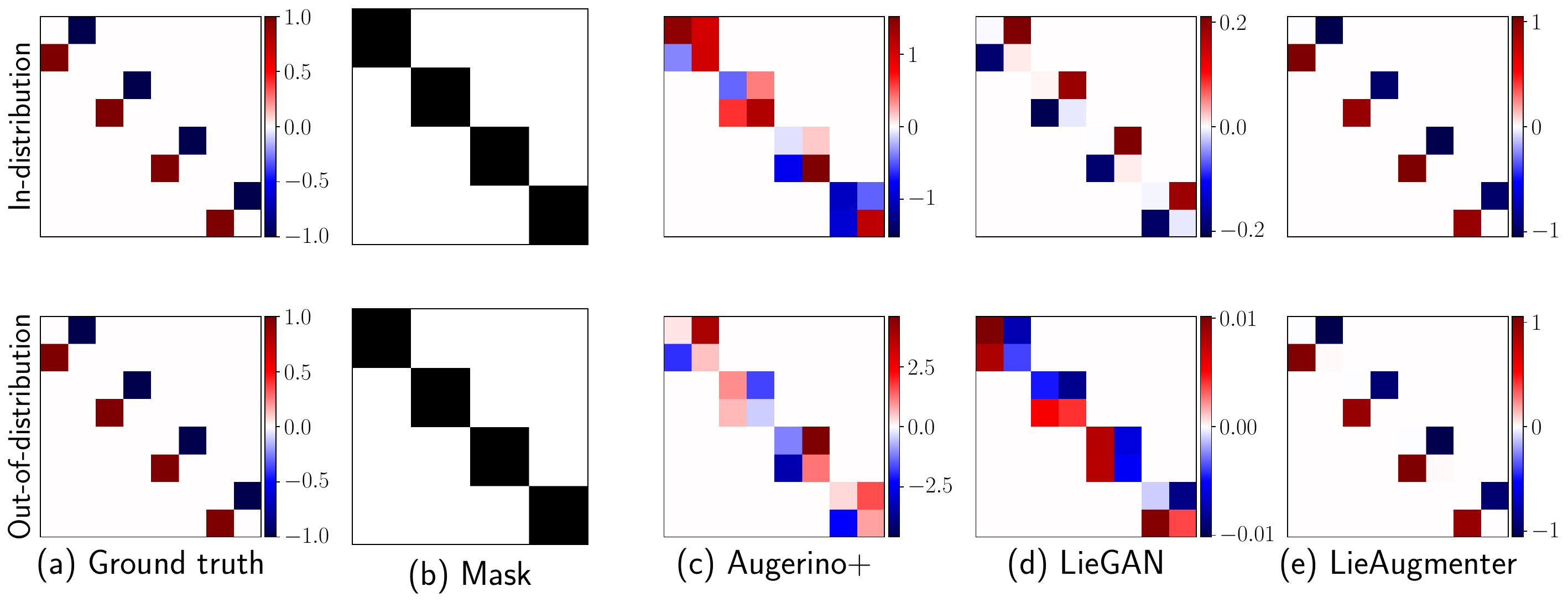}
  \caption{
  Ground-truth Lie algebra basis, mask of learnable entries, and bases learned by baseline symmetry discovery methods and LieAugmenter for 2-body dynamics prediction. Results are shown for the ID (\textit{top}) and the OOD setting (\textit{bottom}).
  }
  \label{fig:lie_algebras_nbody2}
\end{figure}

We further examine LieAugmenter on a simulated $N$-body trajectory dataset from \citet{greydanus2019hamiltonian}. 
Following the experimental protocol of \citet{yangGenerativeAdversarialSymmetry2023}, we generate the dynamical data and use an MLP as the prediction network. 
Specifically, we consider a setting with two bodies of identical mass rotating in nearly circular orbits, where the task is to predict their future coordinates based on past observations.
At each timestep, the input and output feature vectors contain $4N$ entries corresponding to the positions and momenta of all bodies $\{q_{i,x}, q_{i,y}, p_{i,x}, p_{i,y}\}_{i=1}^N$. 
In line with the experimental setting of \citet{yangGenerativeAdversarialSymmetry2023}, we restrict the symmetry search space to transformations that act separately on each body’s position and momentum (see Figure~\ref{fig:lie_algebras_nbody2} second column).
As in Section~\ref{sec:rotated_mnist}, we study both ID and OOD settings, where ID considers the default dataset while in OOD, the training data are constrained so that the first body’s position lies only in the top-left or bottom-right quadrants, and the test set is constructed from the complementary quadrants (top-right or bottom-left).
Here, we focus on the one-step prediction (one input timestep to one output timestep), and we provide additional results in Appendix~\ref{app:nbody}, including a more challenging three-step prediction, where we observe a similarly strong performance.

\paragraph{{Symmetry Discovery.}} 
In Figure~\ref{fig:lie_algebras_nbody2}, we visualize the learned symmetry structure by showing the resulting $2 \times 2$ block-diagonal mask constraint alongside the Lie algebra basis identified by each method.
Across both the ID and OOD settings, LieAugmenter correctly recovers the ground-truth Lie algebra basis, demonstrating that it can reliably infer the underlying equivariant transformations even under distribution shift. Additional results under less restrictive symmetry search spaces are provided in Appendix~\ref{app:nbody}.

\paragraph{Predictive Performance.}
We compare  our method with the analogous augmentation baselines as in Section~\ref{sec:rotated_mnist}. 
Table~\ref{tab:results_nbody2} reports mean squared error (MSE) and equivariance error.
In the ID setting, LieAugmenter achieves the best performance, surpassing oracle augmentation and augmentation based on discovered symmetries. 
In the OOD setting, the performance of the symmetry-discovery baselines drops markedly, while LieAugmenter still matches oracle performance. 
Notably, our method attains strong performance without prior knowledge of the symmetry, indicating that it can both recover symmetry and transform it into robust generalization.

\begin{table*}[h]
  \caption{$2$-body dynamics results for ID and OOD evaluations. We report MSE and equivariance error for MLP models trained (i) without augmentation (\emph{Trivial}), (ii) with \emph{Oracle} rotation augmentation (ground-truth symmetry), (iii) with symmetry-discovery baselines (Augerino+ and LieGAN Aug.), and (iv) with LieAugmenter (\emph{Ours}). All values are reported as $\mathrm{mean}\pm\mathrm{std}$ over three random seeds.}
  \label{tab:results_nbody2}
  \centering
\resizebox{\textwidth}{!}{%
  \addtolength{\tabcolsep}{-0.3em}
\begin{tabular}{lccccc} 
\toprule
& \multicolumn{1}{c}{No augmentation} & \multicolumn{1}{c}{Oracle augmentation} & \multicolumn{3}{c}{Symmetry discovery} \\
\cmidrule(lr){2-2}\cmidrule(lr){3-3}\cmidrule(lr){4-6}
 & Trivial & Rotation & Augerino+ & LieGAN Aug. & LieAugmenter \\
\midrule
ID MSE $(\downarrow)$ & $\text{5.16e-03} \scriptstyle{{\pm \text{2.79e-03}}}$ & $\underline{\text{2.13e-05}} \scriptstyle{{\pm \text{1.56e-05}}}$          & $\text{9.67e+03} \scriptstyle{{\pm \text{1.37e+04}}}$ & $\text{1.10e-03} \scriptstyle{{\pm \text{6.98e-04}}}$ & $\text{\textbf{1.30e-05}} \scriptstyle{{\pm \text{5.58e-06}}}$  \\ 
ID Equiv. Error $(\downarrow)$ & $\text{2.48e-02} \scriptstyle{{\pm \text{7.37e-03}}}$ & $\text{\textbf{3.03e-03}} \scriptstyle{{\pm \text{1.03e-03}}}$ & $\text{2.37e-01} \scriptstyle{{\pm \text{7.53e-02}}}$ & $\text{1.48e-02} \scriptstyle{{\pm \text{4.80e-03}}}$ & $\underline{\text{3.28e-03}} \scriptstyle{{\pm \text{9.36e-04}}}$           \\ 
\midrule
OOD MSE $(\downarrow)$          & $\text{1.86e-02} \scriptstyle{{\pm \text{3.74e-03}}}$ & $\text{\textbf{3.03e-05}} \scriptstyle{{\pm \text{8.81e-06}}}$ & $\text{1.48e+05} \scriptstyle{{\pm \text{2.09e+05}}}$ & $\text{8.94e-03} \scriptstyle{{\pm \text{9.31e-04}}}$ & $\underline{\text{3.73e-05}} \scriptstyle{{\pm \text{2.97e-05}}}$           \\ 
OOD Equiv. Error $(\downarrow)$ & $\text{3.33e-02} \scriptstyle{{\pm \text{2.56e-03}}}$ & $\text{\textbf{3.70e-03}} \scriptstyle{{\pm \text{1.14e-03}}}$ & $\text{1.95e-01} \scriptstyle{{\pm \text{9.23e-03}}}$ & $\text{2.73e-02} \scriptstyle{{\pm \text{1.10e-03}}}$ & $\underline{\text{5.96e-03}} \scriptstyle{{\pm \text{2.82e-03}}}$           \\
\bottomrule
\end{tabular}
}
\end{table*}

\subsection{Molecular Property Prediction}
\label{sec:qm9}

Next, we evaluate LieAugmenter on molecular property prediction using QM9 dataset \citep{blum2009970, rupp2012fast}, which consists of small inorganic molecules represented as sets of atoms with 3D coordinates and atom-type features (e.g., atomic numbers).
We focus on predicting the highest occupied molecular orbital energy (HOMO) and the lowest unoccupied molecular orbital energy (LUMO), two scalar targets that are invariant to global rigid motions of the atomic coordinates.
In other words, rotating and translating all atoms by the same element of $\mathrm{SE}(3)$ does not change the HOMO/LUMO values.
This setting is therefore well-suited for testing whether LieAugmenter can learn a \emph{higher-dimensional} continuous symmetry model, since the Lie algebra basis of $\mathrm{SE(3)}$ is six-dimensional (three infinitesimal rotations and three infinitesimal translations).
\footnote{QM9 models are also invariant to permutations of atom indices. This discrete symmetry is handled by the set architecture and is orthogonal to the continuous $\mathrm{SE}(3)$ symmetry studied here.}

Following the experimental protocol in prior symmetry discovery and data augmentation evaluations on QM9 \citep{bentonLearningInvariancesNeural2020, yangGenerativeAdversarialSymmetry2023}, 
we use LieConv \citep{finzi2020generalizing} with no equivariance (LieConv-Trivial) as the base model, such that any robustness to rigid motions arises from learned Lie-parameterized augmentations and the training objective rather than from architectural constraints. 
We compare the proposed LieAugmenter to alternative augmentation strategies, including (i) no augmentation, (ii) oracle $\mathrm{SE}(3)$ augmentations, and (iii) augmentations generated from symmetries discovered by Augerino and LieGAN. 

\begin{figure}[t]
  \centering
  \includegraphics[width=0.8\linewidth]{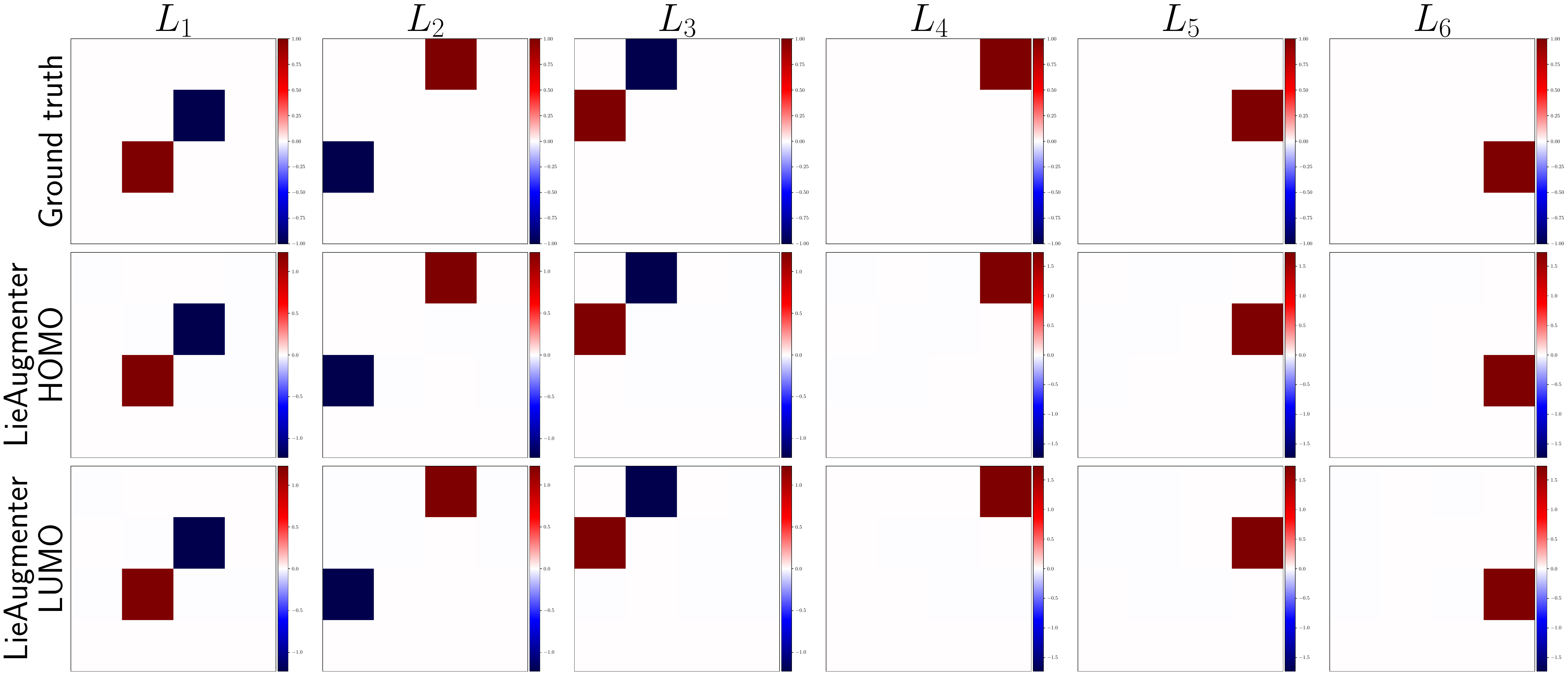}
  \caption{Comparison of canonical Lie algebra basis for $\mathrm{SE}(3)$ group (\textit{top}) and the generators learned by LieAugmenter for HOMO (\textit{middle}) and LUMO (\textit{bottom}) on QM9. 
  For interpretability, we match the learned generators to the closest canonical basis.
  }
  \label{fig:lie_algebras_qm9}
\end{figure}

\paragraph{{Symmetry Discovery.}}
To qualitatively assess symmetry recovery, we visualize the learned generators alongside a canonical Lie algebra basis of $\mathrm{SE(3)}$ in Figure~\ref{fig:lie_algebras_qm9}.
We see that the proposed LieAugmenter recovers generators that are consistent with the  $\mathrm{SE}(3)$ structure across both targets.
Note that bases of $\mathrm{SE(3)}$ are not unique: beyond permutations and sign flips, different linear combinations of generators can span the same Lie subalgebra. 
For this reason, we align learned generators to a canonical basis for visualization (for a more detailed description see Appendix~\ref{app:add_details_metrics}).

\begin{table}[t]
  \caption{MAE (in meV, lower is better) on QM9 dataset for HOMO and LUMO. \textit{Oracle} $\mathrm{SE}(3)$ augmentation uses the ground-truth rigid-motion symmetry. We denote this oracle setting with $^{\star}$. 
  Results marked with $^\dag$ are taken from \citet{bentonLearningInvariancesNeural2020}, and those marked with $^\ddag$ are from \citet{yangGenerativeAdversarialSymmetry2023}.}
  \label{tab:results_qm9}
  \centering
  \resizebox{0.6\linewidth}{!}{%
  \begin{tabular}{lccccc}
  \toprule
  & \multicolumn{1}{c}{No aug.} & \multicolumn{1}{c}{Oracle aug.} & \multicolumn{3}{c}{Symmetry discovery} \\
  \cmidrule(lr){2-2}\cmidrule(lr){3-3}\cmidrule(lr){4-6}
   & Trivial$^\dag$ & SE(3)$^\dag$ & Augerino$^\dag$ & LieGAN$^\ddag$ & LieAugmenter\\
  \midrule
  HOMO & 52.7 & $\text{36.5}^{\star}$ & \underline{38.3} & 43.5 & \textbf{37.1} \\
  LUMO & 43.5 & $\text{29.8}^{\star}$ & \underline{33.7} & 36.4 & \textbf{30.7} \\
  \bottomrule
  \end{tabular}
  }
\end{table}

\paragraph{Predictive Performance and Proximity to Oracle Augmentation.}
Table~\ref{tab:results_qm9} reports the mean absolute error (MAE) in meV for molecular property prediction.
LieAugmenter achieves performance close to oracle $\mathrm{SE}(3)$ augmentation, with an oracle gap of $0.6$ meV on HOMO and $0.9$ meV on LUMO, while outperforming augmentations from the competing symmetry discovery methods.
These results suggest that  LieAugmenter can scale beyond one-dimensional symmetries,
recover generators consistent with $\mathrm{SE}(3)$, and convert the discovered symmetry into improved performance and generalization on molecular regression tasks.

\subsection{Behavior in the Absence of Symmetry}
\label{sec:no_symmetry}

The experiments above show that LieAugmenter can recover the underlying symmetry and improve predictive performance across datasets and tasks. 
Here we study the complementary case: data with no symmetry.
Our goal is to test whether symmetry discovery methods learn misleading spurious transformations, or whether their outputs make it easy to diagnose that the dataset has no underlying symmetry.

We construct a synthetic regression problem that is learnable but has no nontrivial continuous equivariance as follows.
We first construct an anisotropic, mean-shifted input distribution to reduce spurious symmetries in the marginal distribution: $x \sim \mathcal{N}(\mu, \Sigma)$, where $ \mu = (0.2, -0.1, 0.3, -0.2, 0.15)$ and $\Sigma = \mathrm{diag}(1, 2, 4, 8, 16)$. 
We then generate labels using a randomly initialized and frozen MLP $f^\star : \mathbb{R}^5 \rightarrow \mathbb{R}$: $ y = f^\star(x) + \epsilon$, where $\epsilon \sim \mathcal{N}(0, \sigma^2)$ and $\sigma=0.01$. 
With probability one over the draw of $f^\star$, the resulting map remains learnable but has no nontrivial continuous symmetry. 

\begin{figure}[h]
    \centering
    \includegraphics[width=0.6\linewidth]{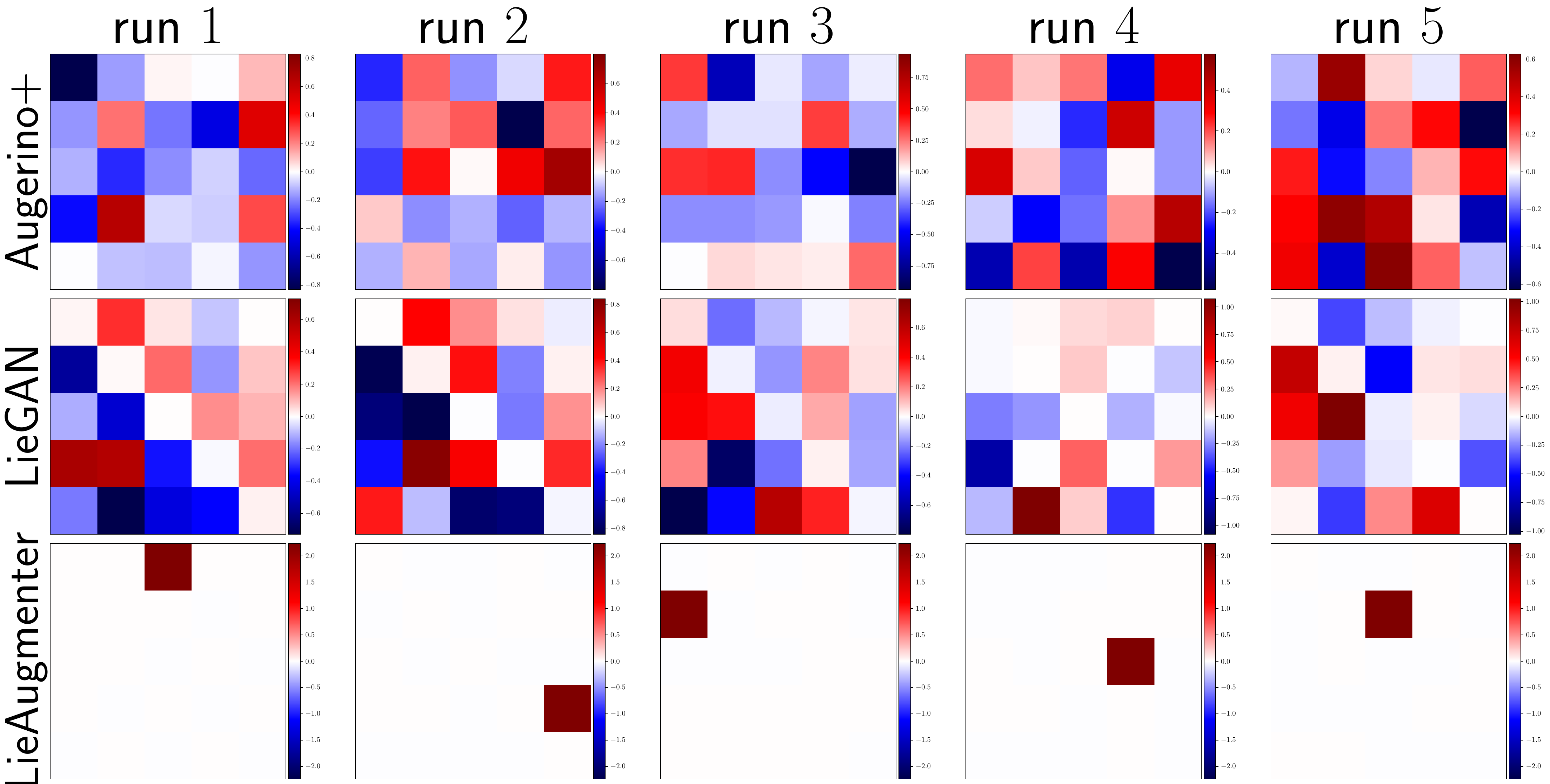}
    \caption{
    Learned Lie algebra bases across multiple runs on a synthetic dataset with no continuous symmetry. 
    Baselines (Augerino+, LieGAN) exhibit noisy representations. 
    LieAugmenter provides an interpretable signature for detecting the absence of symmetry.
    }
    \label{fig:lie_algebras_no_symmetry_runs}
\end{figure}

Figure~\ref{fig:lie_algebras_no_symmetry_runs} summarizes the learned Lie algebra bases across multiple runs.  
Augerino+ and LieGAN learn arbitrary dense and noisy representations across multiple runs, which makes the learned transformations hard to interpret. 
In contrast, LieAugmenter produces sparse bases whose norm concentrates on a single entry, and the active entry changes across runs. 
This run-to-run entry variability indicates that there is no consistent continuous symmetry to recover, and it provides an interpretable empirical signature for detecting the lack of symmetry (see further details in Appendix~\ref{app:detect_no_symmetry_identifiability}).
\section{Conclusions}

We introduce LieAugmenter, an end-to-end framework that jointly learns a prediction network and a Lie-parameterized augmentation generator for task-adaptive symmetry discovery.
Our method relies on a multi-task objective consisting of supervised prediction, approximate equivariance, and regularization to avoid degeneracies and promote an interpretable generator basis.
From a theoretical standpoint, we show identifiability conditions for the discovered symmetry, demonstrate that our method yields approximately symmetry-preserving networks with universal approximation properties, and connect approximate equivariance to robustness under transformation shifts.
In addition, we demonstrate empirical advantages of LieAugmenter in symmetry discovery and predictive performance over competing methods on image classification, $N$-body dynamics prediction, and molecular property prediction, as well as providing an interpretable signature for detecting absence of symmetries.

\paragraph{Limitations and Future Work.}
A limitation of LieAugmenter is that augmentations are sampled via the matrix exponential, which restricts the reachable transformations to the identity component of the Lie subgroup generated by the learned generators. 
In future work, we plan to extend our framework to disconnected symmetries and non-compact Lie groups. 
We also plan to broaden our task-dependent soft-equivariant models to more general symmetry notions, e.g., nonlinear symmetries or structures beyond group actions.
\section*{Acknowledgments}
We thank Valerie Engelmayer, Daniel Herbst, and Andreas Bergmeister for proofreading an earlier version of this manuscript. 
We also thank Jianke Yang for clarifying aspects of the experimental setup used in  \cite{yangGenerativeAdversarialSymmetry2023}.
This work was supported by an Alexander von Humboldt Professorship and a PhD fellowship from the Munich Center for Machine Learning.

\bibliography{paper}
\bibliographystyle{plainnat}

\newpage
\appendix

\begin{center}
    \Large \textbf{Appendix}
\end{center}

The appendix is organized as follows:

\begin{itemize}
    \item Appendix~\ref{app:Additional_Background} includes notation used throughout the paper, explains our definition of the dimensionality of group generators for both vector and image data, and characterizes the class of symmetry groups that our method can recover.

    \item Appendix~\ref{app:Additional_RelatedWork} provides a detailed discussion of prior work related to our approach.
    We cover equivariant machine learning,  data augmentation and learned augmentation, symmetry discovery, and Lie group–based approaches to symmetry discovery.
    
    \item Appendix~\ref{app:Theoretical_Analysis_of_LieAugmenter} includes the proofs of the theoretical properties of LieAugmenter presented in Section~\ref{sec:theoretical_results}, including the Monte Carlo formulation of the equivariance objective, identifiability of the recovered symmetries, universal approximation properties, and the connection between approximate equivariance and robustness under transformation shifts.
    In addition, we provide additional theoretical analysis on a Fourier operator view of partial augmentation, the role of the additional regularization terms as a canonicalization principle, 
    the detection of the absence of symmetry, and a convergence analysis. 
    
    \item Appendix~\ref{app:exp_details} provides the experimental details of our empirical results in Section~\ref{sec:experiments}. 
    We describe additional details of the Lie group parameterization, the hyperparameter settings used across experiments, and the inference procedure. We also define the metrics used to evaluate symmetry discovery and equivariance error, and we summarize the experimental setups for image classification, $N$-body dynamics prediction, molecular property prediction, and for the experiment analyzing behavior in the absence of symmetry.

    \item Appendix~\ref{app:additional_exp} presents additional experimental results for LieAugmenter. We explore an alternative simplified inference strategy and show that the method remains robust, demonstrate empirical convergence, and provide an ablation study and hyperparameter analysis showing stability across a wide range of settings. We also evaluate sample efficiency.
    We also include further results for the $N$-body dynamics experiments, focusing on symmetry discovery with a less restrictive mask and without a mask. We additionally show strong performance on the more challenging scenario of considering and predicting additional timesteps. 
    We provide a runtime analysis, along with further investigation of the behavior in the absence of symmetry.
    We also demonstrate that our approach can discover discrete symmetries, including discrete rotations and partial permutations. Finally, we showcase applications across multiple domains, including Lorentz symmetries in quantum field theory amplitude prediction and rotation symmetries in human colorectal cancer images.
\end{itemize}
\section{Extended Background}\label{app:Additional_Background}

\paragraph{Notation.}
For $v\in\mathbb{R}^d$, $\norm{v}_1$, $\norm{v}_2$, and $\norm{v}_\infty$ denote the $\ell_1$, $\ell_2$, and
$\ell_\infty$ norms. 
For a matrix $A\in\mathbb{R}^{p\times q}$,
let $\norm{A}_F$ be the Frobenius norm and $\langle A,B\rangle_F = \mathrm{tr}(A^\top B)$ the Frobenius inner product.
When using an entry-wise $\ell_1$ penalty, we write $\norm{A}_{1} = \sum_{i,j} \abs{A(i,j)}$.
For a continuous function $f:\mathcal{X}\to\mathbb{R}^d$, we define the  infinity norm over a topological space $\mathcal{X}$
by $\norm{f}_\infty = \sup_{x\in\mathcal{X}} \abs{f(x)}$.
If $\norm{f-g}_\infty<\varepsilon$, we say that $g$ uniformly approximates $f$ up to error $\varepsilon$.
For a probability distribution $Q$, $\mathrm{supp}(Q)$ denotes its topological support.

\subsection{Generator Dimension for Vector Data and Images}\label{app:Generator_Dimension}

Our symmetry model samples Lie algebra elements of the form $A(w)=\sum_{i=1}^C w_iL_i$ and maps them to group elements via $g=\exp(A(w))$. 
The dimensionality of the generator matrices $L_i$ is determined by the space in which the transformation is most naturally parameterized.
For image data, a common choice is to parameterize transformations in a low-dimensional coordinate space (e.g., rotation, scaling, shear) and then apply the corresponding induced action to the high-dimensional pixel representation \citep{gonzalez2009digital, torralba2024foundations, jaderberg2015spatial, mumuni2022data}.
We use $d$ to denote this \emph{transformation parameterization dimension} and distinguish it from $n$, the ambient dimension of the input representation $x\in\mathbb{R}^n$.
In this setting, the generators satisfy $L_i \in \mathbb{R}^{d \times d}$, while their effect on inputs is realized through an induced representation acting on $\mathbb{R}^n$. 
This separation between a low-dimensional Lie algebra parameterization and an induced action on high-dimensional inputs follows common practice in Lie group reparameterization and symmetry discovery for images \citep{falorsi2019reparameterizing, bentonLearningInvariancesNeural2020, yangGenerativeAdversarialSymmetry2023, moskalev2022liegg, ko2024learning}. 

\paragraph{Vector-Valued Data: Generators in the Ambient Representation ($d=n$).}
For vector-valued inputs $x\in\mathbb{R}^n$, we represent the symmetry action directly in the ambient feature space. 
In this case, we take $d=n$ and learn generators $L_i\in\mathbb{R}^{n\times n}$. The action on inputs is realized as the standard matrix action on $\mathbb{R}^n$, yielding augmentations of the form
\begin{equation}
x \mapsto \rho_{\mathcal X}(g)x \equiv g x,\;
g=\exp\left[\sum_{i=1}^C w_i L_i\right]\in\mathbb{R}^{n\times n}.    
\end{equation}
This formulation allows the model to learn general linear symmetries acting directly on the feature representation.

\paragraph{Images: Generators in Coordinate Space (Homogeneous $d=3$).}
For images, most continuous symmetries act on \emph{pixel coordinates} rather than as arbitrary linear maps on vectorized pixel intensities. 
We write an image as $x\in \mathbb{R}^{H \times W \times C}$, and for notational convenience as a vector in $\mathbb{R}^n$ with $n=HWC$. 
Transformations are parameterized in a low-dimensional coordinate space.
We parameterize coordinate transformations in homogeneous coordinates with $d=3$.
A pixel coordinate $p=(u,v)^\top\in\mathbb{R}^2$ is embedded as $\widetilde{p} = (u, v, 1)^\top \in\mathbb{R}^3$, and the sampled group element $g\in\mathbb{R}^{3\times 3}$ acts linearly via $\tilde p' = g \tilde p$.
The corresponding coordinate map is denoted $T_g:\mathbb{R}^2\to\mathbb{R}^2$, and the induced action on image values is given by the pullback $(\rho_{\mathcal X}(g)x)(p) = x(T_g^{-1}(p))$.
In practice, this action is implemented using differentiable grid warping and interpolation.\footnote{For a fixed interpolation scheme, warping is linear in the pixel intensities, so $\rho_{\mathcal X}(g)$ can be viewed as a linear operator on $\mathbb{R}^n$. Due to discretization and resampling effects, this operator need not be exactly invertible for all $g$ in practice. We therefore interpret $\rho_{\mathcal X}$ as an idealized representation of the underlying continuous transformation and its induced action on discrete image data.} 
Learning $L_i\in\mathbb{R}^{3\times 3}$ corresponds to learning infinitesimal
generators of the chosen coordinate transformation family, and $\rho_{\mathcal X}(g)$ is realized by warping the image grid according to $T_g$.

\subsection{Recoverable Symmetry Groups}\label{app:recoverable_sym_group}

Here, we discuss  which symmetry groups can be represented by our Lie-exponential sampler parameterization, which cases are not covered, and how the sampler can be extended to broaden coverage beyond these limitations.

\paragraph{Reachable Set under Single-Exponential Sampling.}
Let $Q_{L,\gamma}$ denote the distribution over group elements induced by the Lie algebra sampler
Eq.~\eqref{eq:group_sampling}.
Our sampler draws a Lie algebra element $A(w)=\sum_{i=1}^C w_i L_i$ and maps it to a group element via a single matrix exponential $g=\exp(A(w))$. 
The induced distribution over transformations therefore satisfies $\mathrm{supp}(Q_{L,\gamma}) \subseteq \exp(\mathrm{span}\{L_1,\dots,L_C\})$.
Since $t\mapsto \exp(tA)$ is a continuous path with $\exp(0)=\mathrm{Id}$,  every sampled $g$ lies in the \emph{identity component} of the Lie subgroup generated by the learned generators.
Therefore, if the ground-truth symmetry group is disconnected, sampling a single exponential cannot reach components not containing the identity (e.g., reflection components).
It is also important to distinguish the linear span $\mathrm{span}\{L_1,\dots,L_C\} $  from the Lie algebra generated  by $\{L_i\}$ under commutators (their Lie closure).
Even if a small set of generators generates a full Lie algebra via commutators, single-exponential sampling only explores $\exp(\mathrm{span}\{L_1,\dots,L_C\})$, i.e., exponentials of linear combinations.
Thus, full recovery of a connected target group by a single exponential requires that $\mathrm{span}\{L_1,\dots,L_C\} $ spans the target Lie algebra (up to a change of basis) and that the exponential map is surjective on
the corresponding connected group.

\paragraph{Examples of Groups that Are Recoverable by a Single Exponential (under Correct Generators).}

The conditions for reachable set are satisfied in many practically relevant continuous symmetry families.
The following families are standard settings in which $\exp$ is surjective (or even a global diffeomorphism), so the single-exponential sampler can in principle cover the entire connected group if $\mathrm{span}\{L_1,\dots,L_C\} $ spans the appropriate Lie algebra: (i) translation $\mathbb{R}^n$ and other abelian additive groups, 
(ii) positive scaling $\mathbb{R}_{>0}$ and products $\mathbb{R}_{>0}^k$ (e.g., global or channel-wise intensity scalings), 
(iii) tori $\mathbb{T}^k \cong \mathrm{SO}(2)^k$ (e.g., phase or angle symmetries), 
(iv) connected and compact matrix groups, including $\mathrm{SO}(n)$, $\mathrm{U}(n)$, and $\mathrm{SU}(n)$ (for which $\exp$ is surjective) \citep{kirillov2008introduction}, 
(v) rigid motions groups $\mathrm{SE}(n) = \mathrm{SO}(n) \ltimes \mathbb{R}^n$, which is commonly parameterized via exponential coordinates, and 
(vi) connected, simply connected nilpotent matrix groups (e.g., unipotent upper-triangular groups and the Heisenberg group) admit a global diffeomorphism.
These cases cover many continuous augmentation families used in practice, including rotations, translations, rigid motions, and certain continuous photometric transformations \citep{gross1996role, gilmore2006lie, thomas2018tensor, rao1998learning, finzi2020generalizing, falorsi2019reparameterizing, sola2018micro, kondor2018generalization, sohl2010unsupervised, cohen2014learning,  dehmamy2021automatic, miao2007learning, shorten2019survey, taylor2018improving, kim2021point, alomar2023data}.

\paragraph{What Is Not Recoverable by a Single Exponential.}
There are two main ways single-exponential sampling can fail to recover a desired symmetry group. 
First, for disconnected groups, since the ground-truth symmetry group has multiple connected components, then $g=\exp(A(w))$ cannot leave the identity component. 
Therefore, the sampler cannot represent symmetries such as reflections, flips (and other discrete components), orthogonal group $\mathrm{O}(n)$ (two components, $\det=\pm 1$), where the identity component is $\mathrm{SO}(n)$, and full Euclidean group  $\mathrm{E}(n)=\mathrm{O}(n)\ltimes\mathbb{R}^n$, whose identity component is $\mathrm{SE}(n)$.
Second, even when the symmetry group is connected, the exponential map need not be surjective in general (notably for many non-compact Lie groups), so sampling a \emph{single} exponential may not cover all elements of the identity component \citep{kirillov2008introduction, moskowitz2003exponential}. 
A standard connected counterexample is $\mathrm{SL}(2,\mathbb{R})$, whose exponential image is a strict subset of the group.
A further representational limitation is that, even when the target group is exponential, if $C$ is too small or optimization converges to generators whose span $\mathrm{span}\{L_1,\dots,L_C\} $  is a strict subspace of the target Lie algebra, then the single-exponential sampler can only recover a proper subgroup.

While our main experiments use a single-exponential sampler $g=\exp(\sum_i w_iL_i)$, there are standard extensions that broaden the representable symmetry families. 
First, one can sample a finite product of exponentials:
\begin{equation}\label{eq:multi_exp_sampler}
    g = \prod_{r=1}^{R}\exp\left[\sum_{i=1}^{C} w^{(r)}_i L_i\right],
\; w^{(r)} \sim P(\gamma)
\end{equation}
which remains differentiable and stays within the identity component \citep{wustner2003connected, kirillov2008introduction}.
Second, disconnected symmetries can be handled by a discrete mixture over connected components using the decomposition
$G=\bigsqcup_{r\in\mathcal{R}} rG_0$, where $G_0$ is the identity component, $\mathcal{R}$ is a discrete set of
representatives for the connected components,  and $G/G_0$ is discrete \citep{kirillov2008introduction}. 
We leave a systematic study of these extensions to future work, since they introduce additional computational overhead  and discrete optimization complexity (learning over components) beyond the scope of this paper.
\section{Extended Related Work}
\label{app:Additional_RelatedWork}

The ubiquitous nature of symmetries in nature have led them to continuously be a predominant part of machine learning approaches. In particular, the three main ways in which symmetry is considered are: (i) the enforcement of known symmetries when training a model, (ii) the discovery of unknown symmetries from data and tasks, and (iii) the promotion of symmetries during training for a model that breaks pre-specified symmetries \citep{otto2023unified}.
In this extended related work section, we present different examples of such approaches, with a special focus on the ones that are most related to our proposed method in that they discover unknown symmetries and softly promote them during training.

\subsection{Equivariant Machine Learning}

Many works have explored the design of equivariant models for groups known a priori, such as permutations in sets \citep{zaheerDeepSets2017}, translations \citep{krizhevsky2012imagenet}, roto-translations \citep{fuchs2020se}, or local gauge transformations \citep{cohen2019gauge}. 
Related to our proposed parameterization of symmetries is the use of Lie groups for the definition of networks equivariant to known continuous symmetries.
For example, \citet{macdonald2022enabling} proposes a framework for the definition of architectures invariant to any pre-specified finite-dimensional Lie group, which includes previously overlooked warp transformations. 
The scalability of this approach is however limited by the high memory consumption of their convolutional layers.

Nevertheless, while harnessing symmetries as inductive biases to define equivariant architectures leads to many benefits, their highly constrained design for a priori specified symmetries is a challenging endeavor and limits their flexibility and applicability \citep{goel2025any}.
In particular, the associated architectural design requires the challenging derivation and implementation of group-specific equivariant layers, which precludes their direct application to other settings and domains with potentially different symmetries.
This limitation has motivated a line of work that seeks to automatize the design of architectures equivariant to pre-specified symmetry groups, with methods such as EMLP \citep{finzi2021practical} and GCNN \citep{cohenGroupEquivariantConvolutional2016}. Similarly, ASEN \citep{goel2025any} is designed as an approach that can be simultaneously equivariant to different subgroups of a pre-defined base group by modulating auxiliary input features.

In addition, another limitation of equivariant architectures arises in their application to real-world data, which often deviates from strict mathematical symmetries due to, for instance, noisy or biased measurements and symmetry-breaking effects \citep{manolache2025learning, wang2022approximately}. In such cases, strictly enforcing equivariance can lead to a degradation of the training dynamics and predictive performance of the models. As a result, approximately equivariant networks emerge as a more fitting alternative for such scenarios, where they are still biased towards preserving the symmetry but are not strictly restricted to do so. Some examples include the constrained optimization approach ACE \citep{manolache2025learning}, the relaxed weight-sharing and weight-tying strategy from \citet{wang2022approximately}, or the graph coarsening method for GNNs described in \citet{huang2023approximately}.
Most related to our method is REMUL \citep{elhagRelaxedEquivarianceMultitask2024}, which frames learning equivariance as a weighted multi-task problem for unconstrained models. 

Besides the introduction of architectural constraints, there are also various model-agnostic methods for realizing equivariant models to a priori known symmetries.

Frame averaging \citep{puny2022frame, duval2023faenet, lin2024equivariance} seeks to achieve equivariance by averaging over a small input-dependent subset of the full group, which is known as a frame. This approach has been widely adopted, e.g., for material modeling \citep{duval2023faenet} or antibody generation \citep{martinkus2023abdiffuser}, but its applicability is restricted by its prohibitive computational cost for many other domains.

Canonicalization \citep{kaba2023equivariance, ma2024canonicalization}, on the other hand, implements equivariance by defining a map from all equivalent inputs under the symmetry group to the same canonical form prior to their processing by the prediction network. In this way, any non-invariant base network can be made to produce invariant predictions in a simpler and more efficient manner.
This approach, however, has some limitations, such as potentially resulting in architectures discontinuous with respect to the input \citep{dym2024equivariant, lin2025adaptive}.
Related to our work in its consideration of Lie groups is LieLAC \citep{shumaylov2025lie}, which is designed to harness the action of the infinitesimal generators of the symmetry group, avoiding the need for full knowledge of the group structure.

\subsection{Data Augmentation and Learned Augmentation}

An integral part of our proposal consists of defining a data augmentation process that, rather than relying on predefined transformations, as is predominantly common in the literature \citep{shorten2019survey, mumuni2022data, gerken2022equivariance, quiroga2020revisiting, xu2023comprehensive}, has learnable parameters that are trained jointly with the prediction model. 

Learnable augmentations have already been explored in the context, for instance, of anomaly detection for time series \citep{qiu2021neural}, or self-supervised GNNs \citep{kefato2021jointly}, showing that they are capable of providing additional flexibility and enhanced performance. 
However, in these previous works, the learnable augmentations are represented by black-box models, such as Multi-Layer Perceptrons (MLPs), limiting the interpretability of the generated augmentations. 
In contrast, our approach parameterizes the augmentations using Lie group theory, thus enabling simplified interpretability. 

A more interpretable learnable augmentation approach is InstaAug \citep{miao2022learning} which learns an invariance module simultaneously with the end-to-end training of a downstream model. Differently from previous learnable augmentation approaches, that invariance module generates instance-specific augmentations by capturing local invariances.
However, this approach relies on a parameterization over a distribution of pre-specified transformations, and so it also does not learn to generate augmentations for a priori fully unknown symmetries. 

\subsection{Symmetry Discovery and Lie Group-Based Approaches}

In addition to LieGAN \citep{yangGenerativeAdversarialSymmetry2023} and Augerino \citep{bentonLearningInvariancesNeural2020}, which are the two models most related to our proposal, many other symmetry discovery methods with different restrictions on the symmetry search space and limitations have been proposed in recent years.
The majority of these approaches share the usage of Lie groups for parameterizing the symmetries, which allows for wide applicability to many relevant tasks and also for simplified interpretability. 
In this section, we analyze some of the most relevant such approaches, which we categorize according to their underlying methodology, and provide a comparison of their main characteristics in Table~\ref{tab:comparison_symmetry_discovery}.

\begin{table}[t]
    \centering
    \caption{Comparison of symmetry discovery approaches and our proposed LieAugmenter method based on their most relevant characteristics. While there are methods that are able to discover a wider class of symmetries, LieAugmenter provides widespread applicability and adaptability to a common and useful set of symmetries, while simultaneously producing approximately equivariant predictions.
    We group the methods according to their underlying approach and specify for each the type of symmetry generators, whether the discovered symmetry is explicitly represented, whether the approach produces approximately equivariant predictions jointly with the symmetry discovery, whether the mode can learn task-dependent symmetries, and whether it has been shown to be applicable to discrete and continuous symmetries.
    }
    \label{tab:comparison_symmetry_discovery}
    \resizebox{1\textwidth}{!}{%
    \begin{tabular}{llccccccc}
        \toprule
         & & Symmetry generators & Explicit & Equiv. predictions & Task-dependent & Discrete & Continuous \\
         \midrule
         \multirow{5}{*}[0em]{Generative} & LieFlow \citep{park2025discovering} & Pre-specified & $\times$ & $\times$ & $\times$ & $\checkmark$ & $\checkmark$ \\
         & SGM \citep{allingham2024generative} & Pre-specified & $\times$ & $\times$ & $\times$ & $\checkmark$ & $\checkmark$ \\
         & SymmetryGAN \citep{desaiSymmetryGANSymmetryDiscovery2022} & Linear & $\times$ & $\times$ & $\times$ & $\checkmark$ & $\checkmark$ \\
         & LieGAN \citep{yangGenerativeAdversarialSymmetry2023} & Affine & $\checkmark$ & $\times$ & $\times$ & $\checkmark$ & $\checkmark$ \\
         & LaLiGAN \citep{yang2023latent} & Non-linear & $\times$ & $\times$ & $\times$ & $\checkmark$ & $\checkmark$ \\
         \midrule
         \multirow{3}{*}[0em]{Probing} & LieGG \citep{moskalev2022liegg} & Affine & $\checkmark$ & $\times$ & $\times$ & $\times$ & $\checkmark$ \\
         & \citet{shaw2024symmetry} & Vector fields & $\times$ & $\times$ & $\times$ & $\checkmark$ & $\checkmark$ \\
         & LieNLSD \citep{hu2025explicit} & Non-linear & $\checkmark$ & $\times$ & $\times$ & $\times$ & $\checkmark$ \\
         \midrule
         Neural ODEs & \citet{ko2024learning} & Vector fields & $\times$ & $\times$ & $\checkmark$ & $\times$ & $\checkmark$ \\
         \midrule
         \multirow{2}{*}[0em]{Architectural} & L-conv \citep{dehmamy2021automatic} & Vector fields & $\times$ & $\checkmark$ & $\checkmark$ & $\times$ & $\checkmark$ \\
         & WSCNN \citep{van2024learning} & Regular representations & $\times$ & $\checkmark$ & $\checkmark$ & $\checkmark$ & $\times$ \\
         \midrule
         \multirow{2}{*}[0em]{Augmentations} & Augerino \citep{bentonLearningInvariancesNeural2020} & Pre-specified & $\times$ & $\checkmark$ & $\checkmark$ & $\times$ & $\checkmark$ \\
         &LieAugmenter (\textit{Ours}) & Affine & $\checkmark$ & $\checkmark$ & $\checkmark$ & $\checkmark$ & $\checkmark$ \\
         \bottomrule
    \end{tabular}
    }
\end{table}

\subsubsection{Generative Methods}

Many symmetry discovery approaches coincide in their modeling of the symmetry discovery process as a \textit{generative method} that seeks to identify the transformations that preserve the data distribution. As previously explained, this type of models place strong and not always realistic assumptions on the uniformity of the action of the group on the data, cannot detect task-specific symmetries, and also require the definition and training of a subsequent prediction network to make use of the discovered symmetries. 

As an example, LieFlow \citep{park2025discovering} defines symmetry discovery as a flow matching problem from a pre-specified supergroup to the subgroup that matches the true underlying symmetries of the data. This approach is applicable to both continuous and discrete symmetries, however its scalability to real-world datasets is still unexplored, and it displays a ``last-minute mode convergence'' behavior that limits its ability to discover complex symmetries. 

Symmetry-aware Generative Model (SGM) \citep{allingham2024generative} is another approach that defines a generative model to capture approximate symmetries by learning the extent to which a broad pre-specified set of possible symmetries is present in the data. In particular, this method models a distribution over the data distribution by decomposing it into a distribution over prototypes and a distribution over parameters that control the transformations applied to each prototype. This two-step process depends, however, on the identification of a correct and unique prototype per sample, which might not always be possible for real-world data.

SymmetryGAN \citep{desaiSymmetryGANSymmetryDiscovery2022} learns symmetry transformations on the dataset distribution using a method based on generative adversarial networks (GANs). In particular, this method defines symmetries in terms of inertial reference densities of the data, in such a way that it requires the generator network to preserve not only the PDF of the input, but also the PDF of the defined inertial density. An important limitation of this approach, however, is that it can only discover one group element per training round and requires additional terms and techniques to extend its applicability to finding subgroups or fully identifying the group.

LaLiGAN \citep{yang2023latent} is an extension of the approach described in LieGAN to non-linear symmetries. This method learns to map the data to a latent space where the symmetry becomes linear through the combination of an autoencoder model and a GAN network trained on the latent embeddings. 
Just like LieGAN \citep{yangGenerativeAdversarialSymmetry2023}, this approach shares with our method the inclusion of regularization terms in the loss function to encourage desirable properties for the discovered symmetry.
In any case, while this approach is able to learn a wider class of symmetries, it is still restricted to the assumption of compact Lie groups and various restrictive conditions on the orbit space of the group actions. Moreover, the discovered symmetries are explicitly represented only in terms of their linear action on the latent space, but not directly for their effect on the input space. 

Therefore, given that our approach already considers a large class of useful and common symmetries and that beyond non-linear symmetries there is still a larger family of symmetries that remains undiscoverable by this type of generative approaches, we believe that the question of greater importance lies in addressing their other common limitations that we previously highlighted.
In particular, we view the necessity of training an additional network to make use of the symmetry, the assumptions on the data distribution, and the lack of adaptability to the task as more critical restrictions for the usefulness and applicability of these symmetry discovery techniques.
As a result, those are the factors that we addressed with our proposed method and we leave the exploration of potential extensions to larger classes of symmetries to future work.

\subsubsection{Probing Trained Networks}

In contrast to approaches that discover symmetries in an unsupervised manner directly from the data, we also find a line of work that studies the problem of identifying symmetries from previously trained unconstrained prediction models \citep{krippendorf2020detecting, moskalev2022liegg, hu2025symmetry}.
These methods share the limitation that they assume that the prediction network needs to learn to make (approximately) equivariant predictions in order to perform well on the considered task, which might not be necessarily the case in the overparameterized regime and under limited data.

As an example, LieGG \citep{moskalev2022liegg} finds a Lie algebra basis of the data symmetries learned by a neural network from the polarization matrix of the discriminator function of the dataset, thus providing straightforward and explicit interpretability of the learned symmetry. 

Similarly, \citet{shaw2024symmetry} explores another method for the detection of symmetries from previously defined machine learning functions for the considered dataset. Despite its applicability to continuous symmetries beyond affine transformations, it lacks explicit interpretability into the learned symmetries and, thus, requires their subsequent extraction and analysis. 

Recently, some works have also attempted to discover a larger class of symmetries, with special focus on non-linear ones. 
For instance, LieNLSD \citep{hu2025explicit} is designed to determine the number of infinitesimal generators and their explicit representation from a pre-specified function library. In particular, this approach consists of two separate and decoupled stages: training a neural network and then discovering symmetries from it. In addition, its scalability and lack of applicability to discrete symmetries are important limitations. 

\subsubsection{Neural ODEs}

In order to relax the restrictions over the class of symmetries that can be discovered (e.g., pre-specified supergroups, affine transformations, etc.), \citet{ko2024learning} introduce an approach based on Neural Ordinary Differential Equations (Neural ODEs) that models one-parameter Lie groups.
However, this approach requires to carefully select numerical differentiation and interpolation methods to ensure stability of the training and it comes with an associated high computational cost due to the integration with ODEs. Moreover, it can only be applied to domains with proper validity scores, which are selected based on characteristics of the data and target task. This method also does not directly produce predictions, needing an additional network to apply the discovered symmetries.

\subsubsection{Learnable Architectural Constraints}

There are also some works that attempt to discover symmetry through \textit{learnable architectural constraints}. The resulting approaches learn to generate equivariant predictions that are potentially more exact than those obtained by data augmentation-based methods, but this comes at the cost of additional computational complexity, less flexible architectures, and a restricted class of discoverable symmetries.

For instance, L-conv \citep{dehmamy2021automatic} proposes a building block for group equivariant networks that is based on using localized kernels constructed from a learnable Lie algebra basis.
However, this method suffers from high computational complexity due to the computation of exponential maps in every layer of their architecture, and is also not well-suited for dealing with approximate and inexact symmetries.

Similarly, Weight Sharing Convolutional Neural Networks (WSCNNs) \citep{van2024learning} learn a weight-sharing scheme from data in the form of a collection of learnable doubly stochastic matrices, which can implement regular group convolutions for strongly symmetric datasets. The associated computational requirements, however, prevent their application to large groups or high-resolution data, and the need for task-specific regularization also limits their applicability.

In addition, other methods have been proposed with the goal of streamlining the discovery of highly specific families of symmetries. 
As an example, \citet{karjol2025learning} proposes G-Ortho-Nets as an approach for the discovery of orthogonal groups, which preserve a specific quadratic form, and their joint incorporation into equivariant models. While restricted to such a narrow class of symmetries, this method has the advantage of simultaneously learning a prediction network during training in contrast to the previously presented two-step approaches. 
Similarly, Noether's razor \citep{ouderaaNoethersRazorLearning2024} parameterizes symmetries as learnable quadratic conserved quantities and discovers them from training data through approximate Bayesian model selection. This approach requires some post-processing based on the singular value decomposition of the learned generators to identify the discovered symmetries, but also has the advantage of jointly learning a prediction model.

\subsubsection{Latent Space Disentanglement}

Another line of work explores the related problem of learning a latent space where the embeddings are equivariant to the action of a pre-specified group.

Equivariance by Contrast (EbC) \citep{schmidt2025equivariance}, for instance, applies contrastive learning to train an encoder-only method on a dataset of paired samples $(y, g \cdot y)$, showing that the resulting embeddings present high-fidelity equivariance. 

Similarly, there is also a set of related works that correspond to unsupervised representation learning of latent spaces that disentangle the action of a potentially unknown group on the original input space to facilitate the subsequent application of simpler equivariant models that operate on that pretrained latent space \citep{koyama2023neural, mitchel2024neural}. These works, however, are not designed to produce interpretable representations of the possibly unknown group or to directly generate approximately equivariant predictions, and, thus, strongly differ from the problems that our method aims to address.
\section{Theoretical Properties of LieAugmenter}\label{app:Theoretical_Analysis_of_LieAugmenter}

We present the theoretical results and proofs for LieAugmenter.
To support these proofs, we also provide several additional propositions and lemmas, which are numbered separately for clarity.
In addition, we provide an extended theoretical analysis on the ability of when and why the augmenter learns the correct symmetry, including bounds and considerations on the number of augmentations required.
We further analyze the role of our regularizers, interpreting them as a form of canonicalization that promotes an interpretable generator.
Finally, we present a theoretical justification for detecting the absence of symmetry when no nontrivial symmetry is present in the data.

\paragraph{Population Objectives.}
The (unknown) continuous symmetry is a connected matrix Lie group $G$ with representations $\rho_{\mathcal{X}}:G\to \mathrm{GL}(n)$ and $\rho_{\mathcal{Y}}:G\to \mathrm{GL}(m)$ acting on $\mathcal{X}\subseteq\mathbb{R}^n$ and $\mathcal{Y}\subseteq\mathbb{R}^m$.
A function $f:\mathcal{X}\to\mathcal{Y}$ is called \emph{$G$-equivariant} if $f(\rho_{\mathcal{X}}(g)x) = \rho_{\mathcal{Y}}(g) f(x),  \;\; \forall g\in G, \forall x\in\mathcal{X}$. 
LieAugmenter learns generators $L=\{L_i\in\mathbb{R}^{n\times n}\}_{i=1}^C$ and a coefficient distribution $P(\gamma)$ on $w\in\mathbb{R}^C$, sampling $g = \exp(\sum_{i=1}^C w_i L_i),\; w\sim P(\gamma)$, as in Eq.~\eqref{eq:group_sampling}. 
Let $Q_{L,\gamma}$ denote the distribution over group elements induced by the Lie algebra sampler
Eq.~\eqref{eq:group_sampling}.
Define the learned Lie algebra subspace and its generated connected subgroup: $\widehat{\mathfrak{g}}(L):=\mathrm{span}\{L_1,\dots,L_C\}, \; \widehat{G}(L):=\text{the connected subgroup generated by }\widehat{\mathfrak{g}}(L)$. 
Let $(X,Y)\sim P$.
Define $\mathcal{L}_{\mathrm{emp}}^{\mathrm{pop}}(\Psi): = \mathbb{E}[\norm{Y-\Psi(X)}_2^2]$, $\mathcal{L}_{\mathrm{equiv}}^{\mathrm{pop}}(\Psi,L,\gamma):= \mathbb{E}[\mathbb{E}_{g\sim Q_{L,\gamma}}
[\norm{\Psi(\rho_{\mathcal{X}}(g)X)-\rho_{\mathcal{Y}}(g)Y}_1]]$, and $\mathcal{L}_{\mathrm{main}}^{\mathrm{pop}}:=\alpha \mathcal{L}_{\mathrm{emp}}^{\mathrm{pop}}
+\beta \mathcal{L}_{\mathrm{equiv}}^{\mathrm{pop}}$. 
We study the population quantities and connect them to the
finite-$K$ empirical objective in Eq.~\eqref{eq:main_loss} and  Eq.~\eqref{eq:total_loss}.

\subsection{Proof of Lemma~\ref{lemma:monte_carlo}}

LieAugmenter constructs $K$ augmentations per sample by drawing $g\sim Q_{L,\gamma}$. Therefore, $\mathcal{L}_{\mathrm{equiv}}$ in Eq.~\eqref{eq:Equivariance loss} is a Monte Carlo estimate of $\mathcal{L}_{\mathrm{equiv}}^{\mathrm{pop}}$.
We formalize the unbiasedness and explain why increasing $K$ stabilizes the symmetry-learning signal in the next lemma. 

\paragraph{Lemma~\ref{lemma:monte_carlo}.}
\textit{
Fix $(x,y)\in\mathcal{X}\times\mathcal{Y}$ and parameters $(\Psi,L,\gamma)$.
Let $g_1,\dots,g_K \overset{\mathrm{i.i.d.}}{\sim} Q_{L,\gamma}$ and define 
\begin{equation}
    \widehat{\ell}_{\mathrm{equiv}}(x,y) := \frac{1}{K}\sum_{j=1}^{K}
\norm{\Psi(\rho_{\mathcal{X}}(g_j)x)-\rho_{\mathcal{Y}}(g_j)y}_1.
\end{equation}
Assuming $\mathbb{E}_{g\sim Q_{L,\gamma}}
[\norm{\Psi(\rho_{\mathcal{X}}(g)x)-\rho_{\mathcal{Y}}(g)y}_1]<\infty$, we have $
    \mathbb{E}[\widehat{\ell}_{\mathrm{equiv}}(x,y)] = \mathbb{E}_{g\sim Q_{L,\gamma}} [\norm{\Psi(\rho_{\mathcal{X}}(g)x)-\rho_{\mathcal{Y}}(g)y}_1]$.
Moreover, if $
    \sigma^2(x,y):= \mathrm{Var}_{g\sim Q_{L,\gamma}} (\norm{\Psi(\rho_{\mathcal{X}}(g)x)-\rho_{\mathcal{Y}}(g)y}_1) < \infty
$, then $\mathrm{Var}(\widehat{\ell}_{\mathrm{equiv}}(x,y))=\frac{\sigma^2(x,y)}{K}$.
}
\begin{proof}
    Let $Z_j := \norm{\Psi(\rho_{\mathcal{X}}(g_j)x)-\rho_{\mathcal{Y}}(g_j)y}_1$.
    By i.i.d.\ sampling and linearity of expectation, we have 
    \begin{equation}
        \mathbb{E}[\widehat{\ell}_{\mathrm{equiv}}(x,y)]
=\frac{1}{K}\sum_{j=1}^K \mathbb{E}[Z_j]
=\mathbb{E}[Z_1]
=\mathbb{E}_{g\sim Q_{L,\gamma}}
[\norm{\Psi(\rho_{\mathcal{X}}(g)x)-\rho_{\mathcal{Y}}(g)y}_1],
    \end{equation}
which gives unbiasedness. 
For the variance, independence gives 
\begin{equation}
    \mathrm{Var}(\widehat{\ell}_{\mathrm{equiv}}(x,y))
=\mathrm{Var}(\frac{1}{K}\sum_{j=1}^K Z_j)
=\frac{1}{K^2}\sum_{j=1}^K \mathrm{Var}(Z_j)
=\frac{1}{K^2}\cdot K \sigma^2(x,y)
=\frac{\sigma^2(x,y)}{K}.
\end{equation}
\end{proof}

\begin{remark}
     Let $\{Z_j := \norm{\Psi(\rho_{\mathcal{X}}(g_j)x)-\rho_{\mathcal{Y}}(g_j)y}_1\}_{j=1}^K$ be independent random variables such that $0 \leq Z_j \leq B$   are almost surely, using Hoeffding's inequality yields $\mathbb{P}(|\widehat{\ell}_{\mathrm{equiv}}(x,y)-\mathbb{E}[Z]|\ge \varepsilon)\le
2\exp(-2K\varepsilon^2/B^2)$ for every $\varepsilon>0$.
\end{remark}

\subsubsection{A Fourier-Operator View of Partial Augmentation}\label{app:fourier_operator_K}

Our equivariance loss draws $K$ sampled group elements per example, which can be viewed as a form of partial data augmentation. 
That is, instead of averaging over the entire group, we average over a random subset of size $K$ at each training step. This perspective is closely related to recent Fourier analysis results from \citet{tahmasebi2025data}. 
We briefly summarize the main ideas of their work below and refer to \citet{tahmasebi2025data} for full details.

Let $G$ be a compact group and let $\rho:G\to U(r)$ be a unitary representation acting on an $r$-dimensional subspace. 
Given a distribution $\mu$ on $G$, we define the averaged representation by $D_\mu = \mathbb{E}_{g\sim \mu}[\rho(g)] \in \mathbb{R}^{r\times r}$.
Given i.i.d. samples $g_1,\dots,g_K\sim \mu$, we then define the empirical average as 
$ D_K= \frac{1}{K}\sum_{j=1}^K \rho(g_j)$. 
By the relation between the population operator and empirical operator, we have the following lemma for the concentration of the empirical augmentation operator.
\begin{lemma}
    Let $g_1,\dots,g_K\sim \mu$ be i.i.d. and let $\rho:G\to U(r)$ be unitary. Then, for any $\delta\in(0,1)$, with probability at least $1-\delta$, we have
    $
        \norm{
        D_K - D_\mu
        }_{\mathrm{op}}
    \le
    C \sqrt{\frac{\log(r/\delta)}{K}}
    $
    for some $C>0$. 
\end{lemma}

Now, specifically, we consider $\mu$ as the Haar measure on $G$ and define the invariant subspace by 
$V^G := \{v\in\mathbb{R}^r : \rho(g)v=v \ \forall g\in G\}$ with dimension $r_{\mathrm{inv}}$.
Then, $D_\mu$ is the orthogonal projector onto $V^G$.
Therefore, in an orthonormal basis aligned with $V^G \oplus (V^G)^\perp$, $D_\mu = \mathrm{diag}(I_{r_{\mathrm{inv}}},0)$. 
Let $\widetilde{D}_K$ be the restriction of $D_K$ to the non-invariant subspace in this basis. 
We have $\norm{\widetilde{D}_K}_{\mathrm{op}} = \norm{D_K - D_\mu}_{\mathrm{op}}\leq  C \sqrt{\frac{\log(r/\delta)}{K}} $ 
with probability at least $1-\delta$ for some $C>0$. 
Hence, partial augmentation with $K$ random group elements suppresses non-invariant Fourier modes at a rate $\norm{\widetilde{D}_K}_{\mathrm{op}}^2 = (\log r / K)$. 
In settings where the relevant representation space has dimension $r=O(d^k)$ (e.g., spherical harmonics
up to degree $k$ on $\mathbb{S}^{d-1}$), we have $\log r = O(k\log d)$, and hence $\norm{\widetilde{D}_K}_{\mathrm{op}}^2 = (\frac{k\log d}{K})$ with high probability. 
Thus, the number of augmentations needed to approximate full
group-averaging grows only logarithmically with the ambient dimension and polynomially with the effective Fourier degree.
The connection to LieAugmenter is as follows. 
For fixed augmentation generator parameters $(L,\gamma)$, LieAugmenter samples $K$ transformation augmentations per example from $Q_{L,\gamma}$ and optimizes a Monte Carlo approximation of a group-averaged equivariance objective.
When the augmentation distribution is (approximately) supported on a compact subgroup and the prediction network effectively operates in an $r$-dimensional representation subspace, the resulting empirical augmentation operator concentrates at rate $\tilde O(1/\sqrt{K})$ (up to $\sqrt{\log r}$ factors) following by the results from \citet{tahmasebi2025data}, which further provides a spectral justification for small $K$ in practice.
Empirically, we find that a small number of samples per example (e.g., $K=10$) already yields strong performance (see Section~\ref{sec:experiments} and Appendix~\ref{app:additional_exp}). Appendix~\ref{app:additional_exp} further provides a sensitivity analysis over $K$, showing that our method is stable across a range of augmentations.

\subsection{Proof of Theorem~\ref{thm:identifiability}}\label{app:proof_identifiability}
We now turn to the symmetry discovery mechanism. 
Intuitively, if a sampled transformation does not preserve the task labels, then enforcing label-consistency under that transformation creates a systematic discrepancy and increases the population equivariance loss $\mathcal{L}_{\mathrm{equiv}}^{\mathrm{pop}}$. 
The supervised term fixes the predictor to the ground-truth target, and the equivariance term then selects transformations that act as symmetries of that target. 
This leads to an identifiability statement for the learned connected subgroup $\widehat{G}(L)$, up to the usual non-uniqueness of a basis for the learned Lie algebra subspace.

Consider without loss of generality the following conditions. 
We assume  $Y=f^\star(X)$ almost surely.
In addition, suppose $\mathcal{L}_{\mathrm{equiv}}^{\mathrm{pop}}(f^\star,L, \gamma)=0$.
We assume the supervised population objective $\mathcal{L}_{\mathrm{emp}}^{\mathrm{pop}}(\Psi)$ has a unique minimizer $f^\star$ over the predictor class.
We assume $\mathrm{supp}(P(\gamma))$ contains an open neighborhood of $0\in\mathbb{R}^C$ so $\mathrm{supp}(Q_{L,\gamma})$ contains a neighborhood of the identity in $\widehat{G}(L)$ generated by $\widehat{\mathfrak{g}}(L)=\mathrm{span}\{L_1,\dots,L_C\}$.
When stating conditions for exact recovery, we assume the connected symmetry group of $f^\star$ (under the given representations) is exactly $G$.

\begin{lemma}\label{lem:support_symmetry}
    Assume $Y=f^\star(X)$ almost surely.
    Suppose $\Psi=f^\star$ and $\mathcal{L}_{\mathrm{equiv}}^{\mathrm{pop}}(f^\star,L,\gamma)=0$.
    Then for  every $g\in \mathrm{supp}(Q_{L,\gamma})$,
    \begin{equation}
        f^\star(\rho_{\mathcal{X}}(g)x)=\rho_{\mathcal{Y}}(g)f^\star(x)
\quad\text{for $P_X$-almost every }x\in\mathcal{X}.
    \end{equation}
\end{lemma}
\begin{proof}
    For fixed $g$, define $\phi(g) :=\mathbb{E}_{X}[\norm{f^\star(\rho_{\mathcal{X}}(g)X)-\rho_{\mathcal{Y}}(g)f^\star(X)}_1]$.
    The integrand is nonnegative, so $\mathcal{L}_{\mathrm{equiv}}^{\mathrm{pop}}(f^\star,L,\gamma)=0$ implies $\phi(g)=0$ for $Q_{L,\gamma}$-almost every $g$. 
    Now take any $g_0\in\mathrm{supp}(Q_{L,\gamma})$.
    If $\phi(g_0)>0$, then under the standard continuity of the group actions and the norm, $\phi$ is lower semicontinuous, so there exists a neighborhood $U$ of $g_0$ such that $\phi(g)>0$ for all $g\in U$. Since $g_0$ lies in the support of $Q_{L,\gamma}$, we have $Q_{L,\gamma}(U)>0$, which contradicts $\phi(g)=0$ for $Q_{L,\gamma}$-almost every $g$. Therefore $\phi(g)=0$ for all $g\in\mathrm{supp}(Q_{L,\gamma})$, which is equivalent to the stated equivariance identity holding for $P_X$-almost every $x$.
\end{proof}

\begin{lemma}\label{lem:open_subgroup}
    Let be a connected topological group and let $S\subseteq H$ be a subgroup.
    If $S$ contains a nonempty open neighborhood of the identity in $H$, then $S=H$.
\end{lemma}

\begin{proof}
    If $U\subseteq S$ is an open neighborhood of the identity, then for any $s\in S$ the translate $sU$ is open and contained in $S$, so $S$ is an open subgroup of $H$. A connected topological group has no proper nontrivial open subgroups, because the (open) cosets would partition $H$ into disjoint open sets. Hence $S=H$.
\end{proof}

\paragraph{Theorem~\ref{thm:identifiability}.}
\textit{
Assume  $Y=f^\star(X)$ almost surely and that $f^\star$ is the unique minimizer of $\mathcal{L}_{\mathrm{emp}}^{\mathrm{pop}}(\Psi)$ over the predictor class.
Assume $\mathcal{L}_{\mathrm{equiv}}^{\mathrm{pop}}(f^\star,L)=0$ and  that $P(\gamma)$ has support containing a neighborhood of $0 \in \mathbb{R}^C$, so that $\mathrm{supp}(Q_{L,\gamma})$ contains a neighborhood of the identity element in $\widehat{G}(L)$.
Then, for $\alpha>0$, $\beta>0$: 
\begin{enumerate}
    \item  $\Psi^\star=f^\star$ almost surely.
    \item Every element of the learned connected subgroup $\widehat{G}(L^\star)$ is a symmetry of $f^\star$, i.e., $\widehat{G}(L^\star)\subseteq G$.
\end{enumerate}
If in addition the connected symmetry group of $f^\star$ is exactly $G$ and $\dim(\widehat{G}(L^\star))=\dim(G)$, then $\widehat{G}(L^\star)=G$ and $\widehat{\mathfrak{g}}(L^\star)=\mathfrak{g}$ (the true Lie algebra).
The generator $\{L_i^\star\}$ is identifiable only up to invertible changes of basis spanning the same subspace.
}
\begin{proof}
     We proceed in two steps.
    \begin{enumerate}
    \item The supervised population objective $\mathcal{L}_{\mathrm{emp}}^{\mathrm{pop}}(\Psi)$ is uniquely minimized by $f^\star$.
    Since $\alpha>0$ and $\beta>0$, any global minimizer of $\mathcal{L}_{\mathrm{main}}^{\mathrm{pop}}$ must also minimize the supervised term, which forces $\Psi^\star(x)=f^\star(x)$ for $P_X$-almost every $x$.

    \item With $\Psi^\star=f^\star$ (in the $P_X$-almost-everywhere sense), consider the choice $L=0$. 
    This yields only the identity transformation and therefore $\mathcal{L}_{\mathrm{equiv}}^{\mathrm{pop}}(f^\star,0,\gamma)=0$. 
    Hence any global minimizer must satisfy $\mathcal{L}_{\mathrm{equiv}}^{\mathrm{pop}}(f^\star,L^\star,\gamma^\star)=0$.
    Lemma~\ref{lem:support_symmetry} then implies that every $g\in\mathrm{supp}(Q_{L^\star,\gamma^\star})$ acts as a symmetry of $f^\star$ on the data distribution.
    Moreover, $\mathrm{supp}(Q_{L^\star,\gamma^\star})$ contains an open neighborhood of the identity inside $\widehat{G}(L^\star)$. 
    Let $S := \{g\in \widehat{G}(L^\star): f^\star(\rho_{\mathcal{X}}(g)x)=\rho_{\mathcal{Y}}(g)f^\star(x) \text{ for $P_X$-almost every }x\}$.
    The set $S$ is a subgroup of $\widehat{G}(L^\star)$, and it contains a neighborhood of the identity.
    Lemma~\ref{lem:open_subgroup} gives $S=\widehat{G}(L^\star)$, so every element of $\widehat{G}(L^\star)$ is a symmetry of $f^\star$, which is exactly $\widehat{G}(L^\star)\subseteq G$.
    If in addition the connected symmetry group of $f^\star$ is exactly $G$ and $\dim(\widehat{G}(L^\star))=\dim(G)$, then the inclusion of connected Lie groups with equal dimension implies equality, so $\widehat{G}(L^\star)=G$ and $\widehat{\mathfrak{g}}(L^\star)=\mathfrak{g}$. The remaining ambiguity is the standard one: any invertible linear recombination of the generators produces a different basis for the same subspace $\widehat{\mathfrak{g}}(L^\star)$ and generates the same connected subgroup.
    \end{enumerate}
\end{proof}

\begin{lemma}
Let $B\in\mathrm{GL}(C)$ and define new generators $L'_i:=\sum_{j=1}^C B_{ji}L_j$. 
For any $w\in\mathbb{R}^C$ define $w':=B^{-1}w$.
Then $\sum_i w_iL_i=\sum_i w'_iL'_i$ and hence
$\exp(\sum_i w_iL_i)=\exp(\sum_i w'_iL'_i)$.
Therefore, if the distribution of $w'$ is the pushforward of the distribution of $w$ under
$B^{-1}$, the induced distribution $Q_{L,\gamma}$ over group elements is unchanged.
\end{lemma}
\begin{proof}
By construction, $\sum_i w'_iL'_i=\sum_i w'_i\sum_j B_{ji}L_j=\sum_j (\sum_i B_{ji}w'_i)L_j
=\sum_j w_j L_j$ since $w=B w'$. Applying the matrix exponential, we have $\exp(\sum_i w_iL_i)=\exp(\sum_i w'_iL'_i)$.
\end{proof}

Theorem~\ref{thm:identifiability} provides  a population-level explanation of why LieAugmenter performs symmetry discovery.
The supervised term recovers the target $f^\star$, and the equivariance term then filters the learned transformations so that the connected subgroup generated by $\{L_i^\star\}$ is consistent with the true symmetry structure of the task.

\subsubsection{Regularizers induce canonicalization and regularized identifiability}

Our regularizers in Eq.~\eqref{eq:total_loss} are designed to break this degeneracy and promote a non-trivial, interpretable basis.
We formalize this statement as follows. 

\begin{proposition}\label{prop:reg_canonicalization}
    Let $\mathcal{L}_{\mathrm{main}}^{\mathrm{pop}}(\Psi,L,\gamma) = \alpha \mathcal{L}_{\mathrm{emp}}^{\mathrm{pop}}(\Psi) +\beta \mathcal{L}_{\mathrm{equiv}}^{\mathrm{pop}}(\Psi,L,\gamma)$ with $\alpha,\beta>0$, and define the canonicalization functional
    \begin{equation}
        \mathcal{C}(L,\gamma) :=\lambda \ell_{\mathrm{areg}}^{\mathrm{pop}}(L,\gamma)
        +\eta \ell_{\mathrm{bcreg}}(L)
        +\nu \ell_{\mathrm{bsreg}}(L), \; \text{where }\lambda,\eta,\nu\ge 0.
    \end{equation}
    Assume  $Y=f^\star(X)$ almost surely, and that $f^\star$ is the unique minimizer of $\mathcal{L}_{\mathrm{emp}}^{\mathrm{pop}}$ over the predictor class. 
    Suppose $\mathcal{L}_{\mathrm{equiv}}^{\mathrm{pop}}(f^\star,L)=0$ without loss of generality.
    Assume additionally a gauge fixing condition preventing rescaling degeneracies between $(L,\gamma)$ (e.g., $\|L_i\|_F=1$ for all $i$).
    Let $\mathcal{S}: = \underset{\Psi,L,\gamma}{\arg\min} \; (\Psi,L,\gamma)$  denote the set of global minimizers of the main population objective, and define the set of \emph{symmetry-consistent symmetry parameters} $\mathcal{S}_G
    := \{(L,\gamma):\; (f^\star,L,\gamma)\in \mathcal{S}\}$.
    For $\tau>0$, consider the \emph{canonicalized} (regularized) population objective $\mathcal{L}_{\tau}^{\mathrm{pop}}(\Psi,L,\gamma):= \mathcal{L}_{\mathrm{main}}^{\mathrm{pop}}(\Psi,L,\gamma)
    +\tau\mathcal{C}(L,\gamma)$ and let $(\Psi_\tau,L_\tau,\gamma_\tau)$ be any global minimizer of $\mathcal{L}_\tau^{\mathrm{pop}}$.
    Then, 
    \begin{enumerate}
        \item Let $C^\star:=\underset{(L,\gamma)\in\mathcal{S}_G}{\inf}\mathcal{C}(L,\gamma)$. Then, $\mathcal{L}_{\mathrm{main}}^{\mathrm{pop}}(\Psi_\tau,L_\tau,\gamma_\tau)\le \tau C^\star$ and $\underset{\tau\to 0}{\lim} \;  \mathcal{L}_{\mathrm{main}}^{\mathrm{pop}}(\Psi_\tau,L_\tau,\gamma_\tau)=0$.
        \item Any accumulation point $(\bar\Psi,\bar L,\bar\gamma)$ of $\{(\Psi_\tau,L_\tau,\gamma_\tau)\}_{\tau\rightarrow 0}$ belongs to $\mathcal{S}$.
        Thus, under the assumptions of Theorem~\ref{thm:identifiability} for the main objective, such limit points satisfy $\bar\Psi=f^\star$ almost surely and the learned connected subgroup satisfies $\widehat{G}(\bar L)\subseteq G^\star$.
        \item Every accumulation point $(\bar\Psi,\bar L,\bar\gamma)\in\mathcal{S}$ additionally minimizes the canonicalization functional over $\mathcal{S}_G$, i.e., 
        \begin{equation} 
            (\bar L,\bar\gamma)\in  \underset{(L,\gamma)\in \mathcal{S}_G}{\arg\min} \; \mathcal{C}(L,\gamma).
        \end{equation}
        If this minimizer is unique up to a prescribed equivalence (e.g., signed permutation of generators), then $(\bar L,\bar\gamma)$ is identifiable up to that equivalence.
    \end{enumerate}
\end{proposition}
\begin{proof}
    We proceed in three steps.
    \begin{enumerate}
        \item Let $(L^\dagger,\gamma^\dagger)\in\mathcal{S}_G$ be such that $\mathcal{C}(L^\dagger,\gamma^\dagger)\le C^\star+\epsilon$ for some $\epsilon>0$.
        Since $(f^\star,L^\dagger,\gamma^\dagger)\in \mathcal{S}$, we have $\mathcal{L}_{\mathrm{main}}^{\mathrm{pop}}(f^\star,L^\dagger,\gamma^\dagger)=0$, hence 
        $\underset{\Psi,L,\gamma}{\inf}\mathcal{L}_\tau^{\mathrm{pop}}(\Psi,L,\gamma)
       \le \mathcal{L}_\tau^{\mathrm{pop}}(f^\star,L^\dagger,\gamma^\dagger) = \tau\mathcal{C}(L^\dagger,\gamma^\dagger) \le \tau(C^\star+\epsilon)$.
        By optimality of $(\Psi_\tau,L_\tau,\gamma_\tau)$ and nonnegativity of $\mathcal{L}_{\mathrm{main}}^{\mathrm{pop}}$, we have 
        $\mathcal{L}_{\mathrm{main}}^{\mathrm{pop}}(\Psi_\tau,L_\tau,\gamma_\tau)
        \le
        \mathcal{L}_\tau^{\mathrm{pop}}(\Psi_\tau,L_\tau,\gamma_\tau)
        \le
        \tau(C^\star+\epsilon)$, and letting $\epsilon\to 0$ yields item (1). 
        \item Take any sequence $\tau_k\rightarrow 0$ along which $(\Psi_{\tau_k},L_{\tau_k},\gamma_{\tau_k})\to (\bar\Psi,\bar L,\bar\gamma)$, where the existence of accumulation points follows from compactness and coercivity implied by the gauge fixing. 
        Item (1) gives $\mathcal{L}_{\mathrm{main}}^{\mathrm{pop}}(\Psi_{\tau_k},L_{\tau_k},\gamma_{\tau_k})\to 0$.
        By continuity of $\mathcal{L}_{\mathrm{main}}^{\mathrm{pop}}$,
        $\mathcal{L}_{\mathrm{main}}^{\mathrm{pop}}(\bar\Psi,\bar L,\bar\gamma)=0$, hence
        $(\bar\Psi,\bar L,\bar\gamma)\in\mathcal{S}$.
        The stated symmetry-consistent conclusions then follow
        by Theorem~\ref{thm:identifiability}.
        \item Rewrite optimality as $\mathcal{L}_{\mathrm{main}}^{\mathrm{pop}}(\Psi_\tau,L_\tau,\gamma_\tau)
        +\tau\mathcal{C}(L_\tau,\gamma_\tau)
        \le
        \tau\mathcal{C}(L^\dagger,\gamma^\dagger)$.
        Divide by $\tau$ and use nonnegativity of $\mathcal{L}_{\mathrm{main}}^{\mathrm{pop}}$ to obtain
        $\mathcal{C}(L_\tau,\gamma_\tau)\le \mathcal{C}(L^\dagger,\gamma^\dagger)$ for all
        $(L^\dagger,\gamma^\dagger)\in\mathcal{S}_G$.
        Letting $\tau\rightarrow 0$ along $\tau_k$ and taking limits yields $\mathcal{C}(\bar L,\bar\gamma)\le \mathcal{C}(L^\dagger,\gamma^\dagger)$ for all $(L^\dagger,\gamma^\dagger)\in\mathcal{S}_G$, proving the minimization claim. 
        Uniqueness yields identifiability up to the specified equivalence.
    \end{enumerate}
\end{proof}
The main population objective typically admits a \emph{family} of symmetry-consistent solutions in $(L,\gamma)$ (e.g., corresponding to different subgroups of $G$ and different generator bases), so the symmetry parameters are not uniquely determined by
$\mathcal{L}_{\mathrm{main}}^{\mathrm{pop}}$ alone. Proposition~\ref{prop:reg_canonicalization}
formalizes the role of the regularizers as a \emph{canonicalization principle}: as the
regularization strength $\tau$ decreases, global minimizers of the canonicalized objective approach main-optimal solutions, while simultaneously selecting a canonical representative within the symmetry-consistent solution set by minimizing $\mathcal{C}(L,\gamma)$. 
Under a uniqueness condition for this canonical representative (after gauge fixing), the learned generators become identifiable up to the remaining unavoidable equivalences (e.g., signed permutation).

\subsubsection{Detecting the Absence of Symmetry}\label{app:detect_no_symmetry_identifiability}

LieAugmenter is designed to discover task symmetries by minimizing an equivariance deviation under learned augmentation. 
However, in some applications, the target mapping admits no non-trivial continuous symmetry.
In this case, a desirable behavior of symmetry discovery methods is that the learned augmentation collapses to the identity, thereby avoiding spurious symmetry discovery.
To the best of our knowledge, this no-symmetry regime has not been explicitly investigated in existing Lie-group-based symmetry discovery approaches.  In contrast, we explicitly consider this setting and assess whether the learned augmentations remain non-trivial only when supported by the task.

Note that in practical implementations, we include non-triviality regularization (e.g., $\ell_{\mathrm{areg}}$ in Eq.~\eqref{eq:total_loss}), which discourages identity transformations.
We therefore analyze two complementary cases: (i) a setting in which identity collapse is permitted, allowing the augmenter to revert to the identity when no meaningful continuous symmetry is present, and (ii) a setting in which non-triviality is enforced, in which case the augmenter is driven to learn the simplest admissible transformation that still satisfies the regularized objective.

We assume without loss of generality noiseless supervision $Y=f^\star(X)$ almost surely.
For $g\in G$, the population equivariance violation of the ground-truth map is defined as
\begin{equation}
\label{eq:delta_g_def}
    \Delta(g) =
    \mathbb{E}\left[
    \ell_{\mathrm{task}}\left(
        f^\star(\rho_{\mathcal{X}}(g)X),\ \rho_{\mathcal{Y}}(g)f^\star(X)
    \right)\right]
    \ge0, 
\end{equation}
where $\ell_{\mathrm{task}}$ is the supervised loss.  
We say that $f^\star$ has no non-trivial connected symmetry (locally around the identity) if
\begin{equation}\label{eq:no_symmetry_condition}
    \Delta(g)=0 \iff g=\mathrm{Id}
\end{equation}
for all $g$ in a neighborhood of the identity. 

\begin{lemma}\label{lemma:no_symmetry_identity}
    Let $\alpha,\beta>0$ and we assume $Y=f^\star(X)$ almost surely.
    We further assume $f^\star$ is the unique minimizer of $\mathcal{L}_{\mathrm{emp}}^{\mathrm{pop}}(\Psi)=\mathbb{E}[\ell_{\mathrm{task}}(\Psi(X),Y)]$ over the prediction network class, and assume the no-symmetry condition Eq.~\eqref{eq:no_symmetry_condition}.
    Suppose the sampling family $P(\gamma)$ can realize (or arbitrarily approximate) a point mass at $w=0$ (hence $g=\mathrm{Id}$).
    Consider the population main objective (without regularization),  $\mathcal{L}_{\mathrm{main}}^{\mathrm{pop}}=\alpha \mathcal{L}_{\mathrm{emp}}^{\mathrm{pop}}+\beta \mathcal{L}_{\mathrm{equiv}}^{\mathrm{pop}}$.
    Then any global minimizer $(\Psi^\star,L^\star,\gamma^\star)$ satisfies (i) $\Psi^\star=f^\star$ almost surely and (ii) $Q_{L^\star,\gamma^\star}$ is supported on the identity, i.e., $g=\mathrm{Id}$ almost surely for $g\sim Q_{L^\star,\gamma^\star}$.
\end{lemma}
\begin{proof}
    By definition,
    \begin{equation}
        \mathcal{L}_{\mathrm{main}}^{\mathrm{pop}}(\Psi,L,\gamma)
=
\alpha\mathbb{E}[\ell_{\mathrm{task}}(\Psi(X),Y)]
+
\beta\mathbb{E}_{g\sim Q_{L,\gamma}}
\mathbb{E}\left[
\ell_{\mathrm{task}}\!\left(
\Psi(\rho_{\mathcal{X}}(g)X) \rho_{\mathcal{Y}}(g)Y
\right)
\right].
    \end{equation}

Since $Y=f^\star(X)$ almost surely and $\alpha>0$, any global minimizer must minimize $\mathcal{L}_{\mathrm{emp}}^{\mathrm{pop}}$. By uniqueness, $\Psi^\star=f^\star$ almost surely.
Now fix $\Psi=f^\star$. The equivariance term becomes
\begin{equation}
    \mathcal{L}_{\mathrm{equiv}}^{\mathrm{pop}}(f^\star,L,\gamma)
=
\mathbb{E}_{g\sim Q_{L,\gamma}}[\Delta(g)],
\end{equation}
where $\Delta(g)\ge 0$ is defined in Eq.~\eqref{eq:delta_g_def}.
 Hence, $\mathcal{L}_{\mathrm{equiv}}^{\mathrm{pop}}(f^\star,L,\gamma)=0$
if and only if $\Delta(g)=0$ holds $Q_{L,\gamma}$-almost surely.
Under Eq.~\eqref{eq:no_symmetry_condition}, $\Delta(g)=0$ implies $g=e$ (in a neighborhood of the identity), and since
the sampling can realize $w=0$ (hence $g=\mathrm{Id}$), the minimal attainable value of the equivariance term is $0$ and it is achieved only when $Q_{L,\gamma}$ is supported on $\mathrm{Id}$.
Because $\beta>0$, any global minimizer must attain this minimum, so $g=\mathrm{Id}$ almost surely under $Q_{L^\star,\gamma^\star}$.
\end{proof}

Lemma~\ref{lemma:no_symmetry_identity} provides a  ``no symmetry'' statement that when identity collapse is permitted, the optimal learned augmentations collapse to the identity (vanishing Lie algebra element).

In the full LieAugmenter objective in Eq.~\eqref{eq:total_loss}, the non-triviality regularizer $\ell_{\mathrm{areg}}$ discourages near-identity augmentations.
When the task has no symmetry, this creates a trade-off: the equivariance loss prefers identity, while $\ell_{\mathrm{areg}}$ prefers non-trivial transformations.
In this setting, the learned augmentation generators canonicalize the solution on the least complex non-identity transformation that balances these competing terms, i.e., the simplest transformation consistent with the regularized objective.
In our method, we model enforced non-triviality by constraining the expected Lie algebra energy to be non-zero.
Let $A(w)=\sum_{i=1}^C w_i L_i$ and we use the Frobenius-normalization of generators as $L^\prime_i = \frac{\sqrt{D}}{||L_i||_F} L_i$.
We consider the following constrained problem
\begin{equation}\label{eq:energy_proxy_problem}
    \min_{L,\gamma} \;
\mathcal{L}_{\mathrm{equiv}}^{\mathrm{pop}}(f^\star,L,\gamma)
+
\nu\sum_{i=1}^C \|L_i\|_1
\end{equation}
such that $\mathbb{E}_{w\sim P(\gamma)}[\|A(w)\|_F^2]=\sigma^2$ for  some fixed $\sigma>0$ and $\nu>0$, where $\|\cdot\|_1$ is the entrywise $\ell_1$ norm.
The phenomenon of random single-entry generators is most pronounced when the task provides no strong directional preference.
We capture this via a local isotropy condition around identity: there exist $\kappa>0$ and $r_0>0$ such that, for all $A$ with $\|A\|_F\le r_0$,
\begin{equation}\label{eq:local_isotropy}
    \Delta(\exp(A)) = \kappa \|A\|_F^2.
\end{equation}
Equivalently, the second-order sensitivity of the equivariance violation at identity is proportional to the identity operator in the chosen basis.

\begin{lemma}\label{lemma:l1_ge_frob}
    For any matrix $M$, $\|M\|_1 \ge \|M\|_F$, with equality if and only if $M$ has at most one non-zero entry.
\end{lemma}

\begin{proposition}\label{prop:no_symmetry_single_entry}
    Assume the no-symmetry condition Eq.~\eqref{eq:no_symmetry_condition} and the local isotropy condition Eq.~\eqref{eq:local_isotropy}.
Assume the generator parametrization enforces $\|L_i\|_F=1$ for any active generator, while unused generators can be set to $0$.
Assume $\sigma>0$ is small enough that $\|A(w)\|_F\le r_0$ almost surely under $w\sim P(\gamma)$ at optimality.
Then there exists an optimal solution $(L^\star,\gamma^\star)$ of Eq.~\eqref{eq:energy_proxy_problem} with:
(i) there is an index $i^\star$ such that $w_{i^\star}$ is non-degenerate and $w_j=0$ almost surely for all $j\neq i^\star$, 
(ii) $L_{i^\star}$ has exactly one non-zero entry (up to sign), i.e., $L_{i^\star}=\pm E_{pq}$ for some canonical basis matrix $E_{pq}$, and consequently,
(iii) $A(w)=w_{i^\star}L_{i^\star}$ injects randomness into a single matrix entry.
\end{proposition}

\begin{proof}
    Under Eq.~\eqref{eq:local_isotropy} and the assumption $\|A(w)\|_F\le r_0$ almost surely,
    \begin{equation}
        \mathcal{L}_{\mathrm{equiv}}^{\mathrm{pop}}(f^\star,L,\gamma)
=
\mathbb{E}_{w\sim P(\gamma)}[\Delta(\exp(A(w)))]
=
\kappa\,\mathbb{E}_{w\sim P(\gamma)}[\|A(w)\|_F^2].
    \end{equation}
By the constraint in Eq.~\eqref{eq:energy_proxy_problem}, this equals $\kappa\sigma^2$ for all feasible $(L,\gamma)$.
Therefore, the optimization reduces to minimizing the regularizer $\nu\sum_i \|L_i\|_1$ over feasible solutions.

Let $M$ be the number of active generators (i.e., those used with non-zero coefficient variance).
For each active generator, $\|L_i\|_F=1$ and Lemma~\ref{lemma:l1_ge_frob} gives $\|L_i\|_1\ge 1$.
Therefore, $\sum_{i=1}^C \|L_i\|_1 \ge M$.
This lower bound is achieved by choosing exactly one active generator $L_{i^\star}$ and making it one-sparse, i.e., $L_{i^\star}=\pm E_{pq}$, which satisfies $\|L_{i^\star}\|_F=1$ and $\|L_{i^\star}\|_1=1$.
Feasibility of the energy constraint is achieved by setting all coefficients except $w_{i^\star}$ to zero almost surely and choosing $P(\gamma^\star)$ such that $\mathbb{E}[w_{i^\star}^2]=\sigma^2$,  then
$\mathbb{E}\|A(w)\|_F^2=\mathbb{E}[w_{i^\star}^2]\|L_{i^\star}\|_F^2=\sigma^2$.
Thus there exists a feasible solution attaining the minimal regularizer value, and it is optimal.
\end{proof}

Proposition~\ref{prop:no_symmetry_single_entry} characterizes an optimal structure, i.e., single-entry concentration, but it also implies a large set of equivalent optima.

\begin{corollary}\label{cor:seed_dependence}
Under the assumptions of Proposition~\ref{prop:no_symmetry_single_entry}, any choice of $(p,q)$ and sign $\pm$ yields an optimal one-sparse generator $L_{i^\star}=\pm E_{pq}$ (with an appropriate $P(\gamma)$ satisfying $\mathbb{E}[w_{i^\star}^2]=\sigma^2$).
Hence the set of global optima contains at least $2d^2$ distinct single-entry solutions, and the location $(p,q)$ is not identifiable.
\end{corollary}
\begin{proof}
All canonical basis matrices satisfy $\|E_{pq}\|_F=1$ and $\|E_{pq}\|_1=1$.
By Proposition~\ref{prop:no_symmetry_single_entry}, the objective value is $\kappa\sigma^2+\nu$ for any such choice, and does not depend on $(p,q)$ or sign.
\end{proof}

Corollary~\ref{cor:seed_dependence} explains the empirical observation in Section~\ref{sec:no_symmetry}. 
In no-symmetry tasks with enforced non-triviality, the learned generator norm concentrates on a single entry whose location changes across random seeds. 
Different runs can converge to different (equally optimal) single-entry solutions.
In contrast, when a true continuous symmetry exists, the learned generators exhibit stable, structured patterns across seeds, reflecting identifiable directions preferred by the equivariance objective.

\subsection{Proof of Proposition~\ref{prop:ua_main}}
We now examine whether adding the equivariance term restricts the expressive power of the predictor class. 
We work under the following conditions.
We assume the input domain $\mathcal{X}\subset\mathbb{R}^n$ is compact and closed under the input action, i.e., $\rho_{\mathcal{X}}(g)\mathcal{X}\subseteq \mathcal{X}$ for all $g$ in the support of $Q_{L,\gamma}$.
We assume the target function $f^\star:\mathcal{X}\to\mathcal{Y}$ is continuous and $G$-equivariant $ \forall g\in\mathrm{supp}(Q_{L,\gamma})$.
For approximation, we assume the hypothesis class used for $\Psi$ is dense in $C(\mathcal{X},\mathcal{Y})$ under the uniform norm.
For example, standard multilayer perceptrons $\Psi$ satisfy universal approximation results on compact domains under standard activation assumptions \citep{hornik1989multilayer, cybenko1989approximation, LESHNO1993861}. 
When relating approximation error to self-consistency, we additionally assume the output action is uniformly bounded over the support of the learned augmentation distribution,
\begin{equation}
    M_Y(L,\gamma):=\sup_{g\in \mathrm{supp}(Q_{L,\gamma})}\norm{\rho_{\mathcal{Y}}(g)}_{\mathrm{op}}<\infty.
\end{equation}
Finally, we assume symmetry-consistent sampling $\mathrm{supp}(Q_{L,\gamma})\subseteq G$.

We first state that uniform approximation of an equivariant target immediately controls the population \emph{label-consistency} equivariance loss.

\begin{lemma}\label{lem:approx_implies_label}
    Assume $\mathcal{X}$ is compact and  $\rho_{\mathcal{X}}(g)\mathcal{X}\subseteq \mathcal{X}$ for all
    $g\in \mathrm{supp}(Q_{L,\gamma})$ and all $x\in \mathcal{X}$.
    Suppose the target function $f^\star:\mathcal{X}\to\mathcal{Y}$ is continuous and equivariant $ \forall g\in\mathrm{supp}(Q_{L,\gamma})$. 
    Assume a model $Y=f^\star(X)$ almost surely. If $\Psi$ satisfies $\sup_{x\in\mathcal{X}}\abs{\Psi(x)-f^\star(x)} < \varepsilon$, then 
    \begin{equation}
        \mathcal{L}_{\mathrm{equiv}}^{\mathrm{pop}}(\Psi,L,\gamma)< \varepsilon.
    \end{equation}
\end{lemma}
\begin{proof}
    Fix $g\in \mathrm{supp}(Q_{L,\gamma})\subseteq G$  and $x\in\mathcal{X}$. 
    Using realizability and equivariance of $f^\star$,
    \begin{align*}
\norm{\Psi(\rho_{\mathcal{X}}(g)x)-\rho_{\mathcal{Y}}(g)f^\star(x)}_1
&\le \norm{\Psi(\rho_{\mathcal{X}}(g)x)-f^\star(\rho_{\mathcal{X}}(g)x)}_1
+\norm{f^\star(\rho_{\mathcal{X}}(g)x)-\rho_{\mathcal{Y}}(g)f^\star(x)}_1\\
&= \norm{\Psi(\rho_{\mathcal{X}}(g)x)-f^\star(\rho_{\mathcal{X}}(g)x)}_1
< \varepsilon,
\end{align*}
where the last inequality uses $\rho_{\mathcal{X}}(g)x\in\mathcal{X}$.
Taking expectations over $(X,Y)$ and $g\sim Q_{L,\gamma}$ yields the claim.
\end{proof}

Next, we show that the same approximation guarantee also controls a population \emph{self-consistency} discrepancy, provided the output action is uniformly bounded on the sampled transformations.

\begin{lemma}\label{lem:approx_implies_equiv}
    Assume $\mathcal{X}$ is compact and  $\rho_{\mathcal{X}}(g)\mathcal{X}\subseteq \mathcal{X}$ for all
    $g\in \mathrm{supp}(Q_{L,\gamma})$ and all $x\in \mathcal{X}$.
    Suppose the target function $f^\star:\mathcal{X}\to\mathcal{Y}$ is continuous and equivariant $ \forall g\in\mathrm{supp}(Q_{L,\gamma})$. 
    Suppose the output action is uniformly bounded over sampled transforms: $M_Y(L,\gamma):=\underset{g\in \mathrm{supp}(Q_{L,\gamma})}{\sup}
    \norm{\rho_{\mathcal{Y}}(g)}_{\mathrm{op}}<\infty$.
    Let $\Psi$ satisfy $\sup_{x\in\mathcal{X}}\abs{\Psi(x)-f^\star(x)} < \varepsilon$.
    Define the population self-consistency discrepancy 
    \begin{equation}
\label{eq:self_consistency_def_app}
\mathcal{E}_{\mathrm{self}}^{\mathrm{pop}}(\Psi,L,\gamma)
:=\mathbb{E}[\mathbb{E}_{g\sim Q_{L,\gamma}}
[\norm{\Psi(\rho_{\mathcal{X}}(g)X)-\rho_{\mathcal{Y}}(g)\Psi(X)}_1]].
\end{equation}
Then, $\mathcal{E}_{\mathrm{self}}^{\mathrm{pop}}(\Psi,L,\gamma)
\le (1+M_Y(L,\gamma))\varepsilon$.
\end{lemma}
\begin{proof}
    Fix $g\in \mathrm{supp}(Q_{L,\gamma})\subseteq G$ and $x\in\mathcal{X}$.
    Add and subtract the equivariant target: 
    \begin{align*}
    \norm{\Psi(\rho_{\mathcal{X}}(g)x)-\rho_{\mathcal{Y}}(g)\Psi(x)}_1
    &\le
    \norm{\Psi(\rho_{\mathcal{X}}(g)x)-f^\star(\rho_{\mathcal{X}}(g)x)}_1 \\
    &\quad
    +\norm{f^\star(\rho_{\mathcal{X}}(g)x)-\rho_{\mathcal{Y}}(g)f^\star(x)}_1\\
    &\quad
    +\norm{\rho_{\mathcal{Y}}(g)\big(f^\star(x)-\Psi(x)\big)}_1.
    \end{align*}
    The first term is at most $\varepsilon$ since $\rho_{\mathcal{X}}(g)x\in\mathcal{X}$.
    The middle term is $0$ by equivariance of $f^\star$. 
    The last term is bounded by $\norm{\rho_{\mathcal{Y}}(g)(f^\star(x)-\Psi(x))}_1
\le \norm{\rho_{\mathcal{Y}}(g)}_{\mathrm{op}}\,\norm{f^\star(x)-\Psi(x)}_1
\le M_Y(L,\gamma)\varepsilon$. 
Taking expectations over $X$ and $g$ yields the result.
\end{proof}

We now combine the universal approximation property with the results above to obtain a direct implication: if the ground-truth target is equivariant,  then symmetry-agnostic predictors can approximate it arbitrarily well, and therefore can achieve arbitrarily small label-consistency equivariance loss under symmetry-consistent transformations.

\paragraph{Proposition~\ref{prop:ua_main}.}
\textit{
Assume $\mathcal{X}$ is compact and  $\rho_{\mathcal{X}}(g)\mathcal{X}\subseteq \mathcal{X}$ for all
    $g\in \mathrm{supp}(Q_{L,\gamma})$ and all $x\in \mathcal{X}$.
Assume a model $Y=f^\star(X)$ where $f^\star$ is
continuous and equivariant $ \forall g\in\mathrm{supp}(Q_{L,\gamma})$. 
Then for any $\varepsilon>0$ there exists a (symmetry-agnostic) predictor $\Psi$ in the hypothesis class such that 
\begin{equation}
    \sup_{x\in\mathcal{X}}\abs{\Psi(x)-f^\star(x)} < \varepsilon \text{ and }\mathcal{L}_{\mathrm{equiv}}^{\mathrm{pop}}(\Psi,L,\gamma)< \varepsilon.
\end{equation}
If further $M_Y(L,\gamma):=\underset{g\in \mathrm{supp}(Q_{L,\gamma})}{\sup}
    \norm{\rho_{\mathcal{Y}}(g)}_{\mathrm{op}}<\infty$, then $\mathcal{E}_{\mathrm{self}}^{\mathrm{pop}}(\Psi,L,\gamma)\le (1+M_Y(L,\gamma))\varepsilon$.
}
\begin{proof}
    By universal approximation theorem \citep{hornik1989multilayer, cybenko1989approximation, LESHNO1993861},  for any $\varepsilon>0$ there exists a network $\Psi$ in the hypothesis class such that $\sup_{x\in\mathcal{X}}\abs{\Psi(x)-f^\star(x)}< \varepsilon$.
    The remaining claims follow from Lemmas~\ref{lem:approx_implies_label}
and~\ref{lem:approx_implies_equiv}.
\end{proof}

We show that if the target is equivariant, then symmetry-agnostic predictors can approximate it arbitrarily well, and consequently can achieve arbitrarily small \emph{label-consistency} equivariance loss under symmetry-consistent transformations.

\begin{remark}
It is a standard technique to assume that $\mathcal{X}\subset\mathbb{R}^n$ is compact and $\rho_{\mathcal{X}}(g)\mathcal{X}\subseteq\mathcal{X}$, since it is used to i)  cite uniform approximation on compact domains and (ii) make sure that transformed inputs remain within the domain on which $\Psi$ is approximating. 
In the proof above, this assumption is used only to guarantee that $\rho_{\mathcal{X}}(g)x\in\mathcal{X}$ whenever $x\in\mathcal{X}$ and $g$ is a transformation against which the equivariance deviation is evaluated.

It is important to note that LieAugmenter does not require invariance for \emph{all} $g\in G$ in practice.
It only promotes consistency for transformations sampled via Eq.~\eqref{eq:group_sampling}, i.e., $g\sim Q_{L,\gamma}$.
Accordingly, the closure condition can be weakened to $\rho_{\mathcal{X}}(g)\mathcal{X}\subseteq\mathcal{X}$ for all $g\in \mathrm{supp}(Q_{L,\gamma})$, or, equivalently, to invariance over a compact subset $G_0\subseteq G$ on which $Q_{L,\gamma}$ concentrates (e.g., induced by restricting the coefficient samples $w\sim P(\gamma)$ to a bounded range).
\end{remark}

\subsection{Proof of Proposition~\ref{prop:ood}}\label{app:proof_ood}

We now connect the approximate equivariance to out-of-distribution  (OOD) generalization under transformation shift.
 Here the test distribution is obtained by applying a  group element to both inputs and labels. 
 Under such shifts, the key quantity that governs
robustness is a self-consistency discrepancy of the predictor, 
$\norm{\Psi(\rho_{\mathcal{X}}(g)X)-\rho_{\mathcal{Y}}(g)\Psi(X)}_1$, which measures whether applying
the transformation before or after prediction leads to consistent outputs.
Since LieAugmenter directly optimizes the \emph{label-consistency} term in
$\mathcal{L}_{\mathrm{equiv}}^{\mathrm{pop}}$, we first relate these two discrepancies.

Let $(X,Y)\sim P$ be the training distribution. For $g\in G$, define the shifted distribution
$P^{(g)}$ by $(X',Y')=(\rho_{\mathcal{X}}(g)X,\rho_{\mathcal{Y}}(g)Y)$.
Let $\ell:\mathcal{Y}\times\mathcal{Y}\to\mathbb{R}_+$ be the task loss and define the risk
$\mathcal{R}_P(\Psi):=\mathbb{E}_{(X,Y)\sim P}[\ell(\Psi(X),Y)]$.

\begin{lemma}\label{lem:self_consistency_control}
Suppose $M_Y(L,\gamma):=\underset{g\in \mathrm{supp}(Q_{L,\gamma})}{\sup}
    \norm{\rho_{\mathcal{Y}}(g)}_{\mathrm{op}}<\infty$.
    Then the self-consistency discrepancy  in Eq.~\eqref{eq:self_consistency_def_app} satisfies
    \begin{equation}
    \label{eq:self_consistency_control}
    \mathcal{E}_{\mathrm{self}}^{\mathrm{pop}}(\Psi,L,\gamma)
    \le
    \mathcal{L}_{\mathrm{equiv}}^{\mathrm{pop}}(\Psi,L,\gamma)
    + M_Y(L,\gamma)\,\mathbb{E}[\norm{Y-\Psi(X)}_1].
    \end{equation}
\end{lemma}
\begin{proof}
    For any $g$ and any $(X,Y)$, by the triangle inequality, we have
    \begin{align*}
    \norm{\Psi(\rho_{\mathcal{X}}(g)X)-\rho_{\mathcal{Y}}(g)\Psi(X)}_1
    &\le
    \norm{\Psi(\rho_{\mathcal{X}}(g)X)-\rho_{\mathcal{Y}}(g)Y}_1
    +\norm{\rho_{\mathcal{Y}}(g)(Y-\Psi(X))}_1\\
    &\le
    \norm{\Psi(\rho_{\mathcal{X}}(g)X)-\rho_{\mathcal{Y}}(g)Y}_1
    +\norm{\rho_{\mathcal{Y}}(g)}_{\mathrm{op}}\norm{Y-\Psi(X)}_1.
    \end{align*}
    Taking expectations over $(X,Y)\sim P$ and $g\sim Q_{L,\gamma}$, and using $\norm{\rho_{\mathcal{Y}}(g)}_{\mathrm{op}}\le M_Y(L,\gamma)$ on the support of $Q_{L,\gamma}$, we have $\mathcal{E}_{\mathrm{self}}^{\mathrm{pop}}(\Psi,L,\gamma)
    \le
    \mathcal{L}_{\mathrm{equiv}}^{\mathrm{pop}}(\Psi,L,\gamma)
    + M_Y(L,\gamma)\,\mathbb{E}[\norm{Y-\Psi(X)}_1]$.
\end{proof}

\begin{remark}
    If $\mathcal{Y}\subseteq\mathbb{R}^m$, the norm equivalence and Cauchy-Schwarz inequality imply $\mathbb{E}\norm{Y-\Psi(X)}_1 \le \sqrt{m}(\mathbb{E}\norm{Y-\Psi(X)}_2^2)^{1/2}
= \sqrt{m}(\mathcal{L}_{\mathrm{emp}}^{\mathrm{pop}}(\Psi))^{1/2}$. 
This shows that small supervised population loss controls the $\ell_1$ prediction error term in Lemma~\ref{lem:self_consistency_control}.
\end{remark}

We now state a risk bound under the transformation shift.
The assumptions on the task loss $\ell$ are standard: we require invariance under the label action and Lipschitz continuity in the prediction argument.

\paragraph{Proposition~\ref{prop:ood}.}
\textit{
Assume that the task loss $\ell$ satisfies: (i) $\ell(\rho_{\mathcal{Y}}(g)a,\rho_{\mathcal{Y}}(g)b)=\ell(a,b) \; \forall  a,b\in\mathcal{Y}, g\in G$, and
    (ii) there exists $L_\ell>0$ s.t. $|\ell(u,y)-\ell(v,y)|\le L_\ell\norm{u-v}_1\; \forall \; u,v,y \in \mathcal{Y}$.
    Then for any network $\Psi$ and any $g\in G$, 
    \begin{equation}
        \mathcal{R}_{P^{(g)}}(\Psi)
\le
\mathcal{R}_{P}(\Psi)
+
L_\ell\,\mathbb{E}[\norm{\Psi(\rho_{\mathcal{X}}(g)X)-\rho_{\mathcal{Y}}(g)\Psi(X)}_1].
    \end{equation}
    In addition, if $\Psi$ is exactly $G$-equivariant, then $\mathcal{R}_{P^{(g)}}(\Psi)=\mathcal{R}_P(\Psi)$.
}
\begin{proof}
    By the definition of $P^{(g)}$, we have $\mathcal{R}_{P^{(g)}}(\Psi)
=\mathbb{E}_{(X,Y)\sim P}[\ell(\Psi(\rho_{\mathcal{X}}(g)X),\rho_{\mathcal{Y}}(g)Y)]$.
Add and subtract $\rho_{\mathcal{Y}}(g)\Psi(X)$ in the first argument and apply Lipschitzness, we obtain
\begin{equation}
    \ell(\Psi(\rho_{\mathcal{X}}(g)X),\rho_{\mathcal{Y}}(g)Y)
\le \ell(\rho_{\mathcal{Y}}(g)\Psi(X),\rho_{\mathcal{Y}}(g)Y)
+L_\ell\norm{\Psi(\rho_{\mathcal{X}}(g)X)-\rho_{\mathcal{Y}}(g)\Psi(X)}_1.
\end{equation}
By label-action invariance, we have
$\ell(\rho_{\mathcal{Y}}(g)\Psi(X),\rho_{\mathcal{Y}}(g)Y)=\ell(\Psi(X),Y)$.
Taking expectations over $(X,Y)\sim P$ gives $\mathcal{R}_{P^{(g)}}(\Psi)
\le
\mathcal{R}_{P}(\Psi)
+
L_\ell\,\mathbb{E}[\norm{\Psi(\rho_{\mathcal{X}}(g)X)-\rho_{\mathcal{Y}}(g)\Psi(X)}_1]$.
\end{proof}

Lemma~\ref{lem:self_consistency_control} and Proposition~\ref{prop:ood} link the training objectives to robustness under transformation shift.
Proposition~\ref{prop:ood} shows that the excess risk under $P^{(g)}$ is controlled by a self-consistency term at $g$, and Lemma~\ref{lem:self_consistency_control} bounds the corresponding population self-consistency discrepancy using the label-consistency equivariance loss and the supervised prediction error. In addition, averaging Proposition~\ref{prop:ood} over $g\sim Q_{L,\gamma}$ yields 
\begin{equation}
    \mathbb{E}_{g\sim Q_{L,\gamma}}[\mathcal{R}_{P^{(g)}}(\Psi)]
\le
\mathcal{R}_{P}(\Psi)
+
L_\ell\,\mathcal{E}_{\mathrm{self}}^{\mathrm{pop}}(\Psi,L,\gamma),
\end{equation} 
and substituting Eq.~\eqref{eq:self_consistency_control} provides an explicit bound in terms of the two quantities optimized by LieAugmenter.

\subsection{Convergence Analysis}\label{app:convergence_analysis}

We now provide a standard convergence guarantee for stochastic gradient methods \citep{kingma2014adam} applied to the reparameterized objective.
Since our objective contains non-smooth components (e.g.,  $\norm{\cdot}_1$ and absolute values in the regularizers), we state the result for a differentiable surrogate $\mathcal{L}^{(\tau)}_{\mathrm{total}}$ obtained by replacing $\abs{z}$ with a smooth approximation such as $\abs{z}_\tau:=\sqrt{z^2+\tau^2}$ and $\norm{v}_1$ with $\norm{v}_{1,\tau}:=\sum_k \abs{v_k}_\tau$ for $\tau>0$.
This yields a differentiable objective with the same minimizers in the limit $\tau\to 0$ under mild conditions.

Let $\theta:=(\theta_\Psi,L,\gamma)$ denote all trainable parameters. 
For a fixed dataset
$\mathcal{D}=\{(x_i,y_i)\}_{i=1}^N$, the expected (smoothed) training objective is defined by 
\begin{equation}\label{eq:F_theta_def}
    F_\tau(\theta)
=
\mathbb{E}_{i\sim \mathrm{Unif}([N])}
\mathbb{E}_{\varepsilon}[
\ell_\tau(\theta;i,\varepsilon)
],
\end{equation}
where $\varepsilon$ is base noise for the reparameterization, $w=r(\varepsilon;\gamma)$, and
$g(\varepsilon;L,\gamma)=\exp(\sum_{c=1}^C w_c L_c)$.
The per-sample loss $\ell_\tau$ corresponds to one datapoint and one sampled group element (or one
sampled set of $K$ elements.

\begin{lemma}\label{lem:unbiased_grad_complete}
Assume $\ell_\tau(\theta;i,\varepsilon)$ is differentiable in $\theta$ and that
$\nabla_\theta$ can be interchanged with $\mathbb{E}_\varepsilon$ in Eq.~\eqref{eq:F_theta_def}.
Let $\widehat{g}(\theta)$ denote the gradient estimator obtained by sampling a minibatch
$\mathcal{B}$ of size $B$ and $K$ independent noises $\varepsilon_{i,1},\dots,\varepsilon_{i,K}$ per
$i\in\mathcal{B}$
\begin{equation}
    \widehat{g}(\theta)
=
\frac{1}{B}\sum_{i\in\mathcal{B}}\frac{1}{K}\sum_{k=1}^K
\nabla_\theta \ell_\tau(\theta;i,\varepsilon_{i,k}).
\end{equation}
Then $\mathbb{E}[\widehat{g}(\theta)]=\nabla F_\tau(\theta)$, where the expectation is over the
minibatch sampling and reparameterization noise.
\end{lemma}
\begin{proof}
    By linearity of expectation and the interchange assumption, $\nabla F_\tau(\theta)
=
\mathbb{E}_{i}\mathbb{E}_{\varepsilon}[\nabla_\theta \ell_\tau(\theta;i,\varepsilon)]$.
Since $\mathcal{B}$ is sampled uniformly and $\varepsilon_{i,k}$ are i.i.d.,
$\mathbb{E}[\widehat{g}(\theta)]=\mathbb{E}_{i}\mathbb{E}_{\varepsilon}[\nabla_\theta \ell_\tau(\theta;i,\varepsilon)]
=\nabla F_\tau(\theta)$.
\end{proof}

\begin{lemma}\label{lem:var_BK_complete}
Assume there exists $\sigma^2<\infty$ such that for all $\theta$,
\begin{equation}
    \mathbb{E}_{i,\varepsilon}[\|\nabla_\theta \ell_\tau(\theta;i,\varepsilon)-\nabla F_\tau(\theta)\|_2^2]
\le \sigma^2.
\end{equation}
Then the estimator $\widehat{g}(\theta)$ in Lemma~\ref{lem:unbiased_grad_complete} satisfies $\mathbb{E}[\|\widehat{g}(\theta)-\nabla F_\tau(\theta)\|_2^2]
\le \frac{\sigma^2}{BK}$.
\end{lemma}

\begin{proof}
$\widehat{g}(\theta)$ is the average of $BK$ independent, mean-zero noise terms
$\nabla_\theta \ell_\tau(\theta;i,\varepsilon)-\nabla F_\tau(\theta)$ (conditionally on $\theta$).
Thus, the variance of an average of independent terms scales inversely with the number of terms.
\end{proof}

\begin{proposition}\label{thm:sgd_convergence_complete}
    Assume $F_\tau$ is lower bounded by $F_{\inf}$ and has $L$-Lipschitz gradient:
$\|\nabla F_\tau(u)-\nabla F_\tau(v)\|_2\le L\|u-v\|_2$ for all $u,v$.
Run SGD with constant step size $\eta\le 1/L$: $\theta_{t+1}=\theta_t-\eta\,\widehat{g}(\theta_t)$, 
where $\widehat{g}(\theta_t)$ is the unbiased estimator in Lemma~\ref{lem:unbiased_grad_complete}
with variance controlled as in Lemma~\ref{lem:var_BK_complete}.
Then for any $T\ge 1$,
\begin{equation}
    \frac{1}{T}\sum_{t=0}^{T-1}\mathbb{E}\big[\|\nabla F_\tau(\theta_t)\|_2^2\big] \le \frac{2\,(F_\tau(\theta_0)-F_{\inf})}{\eta\,T}
+ \eta\frac{L\sigma^2}{BK}.
\end{equation}
Specifically, choosing $\eta=\min\{1/L,\sqrt{2BK(F_\tau(\theta_0)-F_{\inf})/(L\sigma^2T)}\}$ gives $\min_{0\le t<T}\mathbb{E}[\|\nabla F_\tau(\theta_t)\|_2^2] =O\left(\sqrt{\frac{1}{BK\,T}}\right)$.
\end{proposition}

\begin{proof}
    By $L$-smoothness, for any $t$, we have $F_\tau(\theta_{t+1}) \le F_\tau(\theta_t) +
\langle \nabla F_\tau(\theta_t),\theta_{t+1}-\theta_t\rangle
+\frac{L}{2}\|\theta_{t+1}-\theta_t\|_2^2$. 
By substituting $\theta_{t+1}-\theta_t=-\eta\widehat{g}(\theta_t)$, we have $F_\tau(\theta_{t+1})
\le
F_\tau(\theta_t)
-\eta\langle \nabla F_\tau(\theta_t),\widehat{g}(\theta_t)\rangle
+\frac{L\eta^2}{2}\|\widehat{g}(\theta_t)\|_2^2$.
Now, we take the expectation conditional on $\theta_t$ and use unbiasedness
$\mathbb{E}[\widehat{g}(\theta_t)\mid \theta_t]=\nabla F_\tau(\theta_t)$ to have  $\mathbb{E}[F_\tau(\theta_{t+1})\mid \theta_t]
\le
F_\tau(\theta_t)
-\eta\|\nabla F_\tau(\theta_t)\|_2^2
+\frac{L\eta^2}{2}\mathbb{E}[\|\widehat{g}(\theta_t)\|_2^2\mid \theta_t]$.
By decomposition
$\mathbb{E}\|\widehat{g}\|_2^2
=
\|\nabla F_\tau(\theta_t)\|_2^2+\mathbb{E}\|\widehat{g}-\nabla F_\tau(\theta_t)\|_2^2$
and applying Lemma~\ref{lem:var_BK_complete}, we have 
$\mathbb{E}[F_\tau(\theta_{t+1})\mid \theta_t]
\le
F_\tau(\theta_t)
-\eta(1-\frac{L\eta}{2})\|\nabla F_\tau(\theta_t)\|_2^2
+\frac{L\eta^2}{2}\cdot\frac{\sigma^2}{BK}$.
Since $\eta\le 1/L$, we have $1-\frac{L\eta}{2}\ge \frac{1}{2}$. 
By taking the total expectation and summing
over $t=0,\dots,T-1$, it gives 
$\mathbb{E}[F_\tau(\theta_T)]
\le
F_\tau(\theta_0)
-\frac{\eta}{2}\sum_{t=0}^{T-1}\mathbb{E}\|\nabla F_\tau(\theta_t)\|_2^2
+\frac{L\eta^2T}{2}\cdot\frac{\sigma^2}{BK}$.
\end{proof}

Proposition~\ref{thm:sgd_convergence_complete} shows that the stochastic optimization of the (smoothed) LieAugmenter objective converges to an approximate stationary point.
The gradient variance decreases as $1/(BK)$ (Lemma~\ref{lem:var_BK_complete}), so using more
augmentations per sample reduces optimization noise in the same way as increasing the minibatch
size, at the expected computational cost per iteration.
Empirically, we corroborate this convergence behavior in Appendix~\ref{app:empirical_convergence}.
\section{Experimental Details}\label{app:exp_details}

In this section, we provide additional experimental details and setup supporting the empirical results in Section~\ref{sec:experiments}.

We utilized different GPUs for each of our experiments in order to adapt to the demands that their increasing complexity imposed. Specifically, the synthetic dataset without continuous symmetries was run on an NVIDIA GeForce GTX 1060 Mobile GPU, the RotatedMNIST and $N$-body dynamics experiments were run on an NVIDIA A100, and  the QM9 experiment was run on an NVIDIA H100.

All experiments were implemented using PyTorch and executed using CUDA version $\mathrm{12.2}$. 

\subsection{Sampling Strategy and Lie Algebra Basis Initialization}

While the parameter $\gamma$ for the sampling distribution $P(\gamma)$ could in principle be learned, in our experiments we adopt a simplified setting with fixed $\gamma$ and uniform sampling $P(\gamma) = \mathcal{U}[-\gamma, \gamma]$, aiming to avoid any unwanted bias towards sampling a specific subset of group elements.
Even under this simplification, we observe that our method already yields strong empirical performance (see Section~\ref{sec:experiments} and Appendix~\ref{app:additional_exp}). 
In addition, we conduct a sensitivity analysis of $\gamma$ in Appendix~\ref{app:ablation}, demonstrating that our method remains largely robust to changes in the specified value.
Finally, note that our framework can seamlessly support alternative sampling distributions, which is useful in settings such as discrete symmetry discovery (Appendix~\ref{app:discrete_symmetries}).
Extending the framework to a learnable $\gamma$ is a natural next step, and we leave it to future work.

We initialize all the parameters of the Lie algebra basis to a small constant value (1e-2).
This provides an uninformative prior over possible structures and reduces sensitivity to potentially favorable, or adversarial, random initializations.
During training, we additionally enforce a fixed Frobenius norm by re-scaling each basis element as
\begin{equation}
    L^\prime_i = \frac{\sqrt{D}}{||L_i||_F} L_i.
\end{equation}
This normalization keeps the basis at a constant scale, improving optimization stability.
The scale factor depends on the basis dimensionality and thus encourages more consistent representations across bases of different dimensions.

\subsection{Hyperparameter Configuration}

To enable a fair comparison with the considered baselines, we run all competing methods using the same hyperparameter settings reported in their papers or provided in the corresponding public implementations. 
We also follow the training schedule from those sources, including  the same number of epochs and batch size.
As a result, the reported performance of LieAugmenter may be slightly conservative and could potentially be improved with additional tuning of the training schedule.
We also do not perform an exhaustive, fine-grained search over the hyperparameters that are specific to our approach.
More generally, hyperparameter tuning remains an important and widely used technique across many fields, and a more systematic search could further improve performance. In this work, however, our primary goal is not to maximize results through extensive tuning, but to introduce LieAugmenter and demonstrate its effectiveness within a reasonable range of hyperparameter values that aligns with the existing experimental protocols.

Table~\ref{tab:hyperparameters} summarizes the number of epochs, batch size, and hyperparameter values for LieAugmenter used across all experiments.  
In addition, we used the Adam optimizer with a learning rate of 1e-3 in all experiments, keeping the rest of its default hyperparameter values.

\begin{table}[h]
    \centering
    \caption{Number of epochs, batch size, and hyperparameter values for LieAugmenter in each of the experiments. We use $^{\star}$ to denote that the value of $\eta$ is irrelevant for certain experiments due to the Lie algebra basis having $C=1$, which means that $\ell_{\mathrm{bcreg}}=0$.}
    \label{tab:hyperparameters}
    \resizebox{0.8\linewidth}{!}{%
    \begin{tabular}{lcccccccccc}
    \toprule
    & Epochs & Batch size & $\alpha$ & $\beta$ & $\lambda$ & $\eta$ & $\nu$ & $\gamma$ & $K$ & $C$ \\
    \midrule
    Image Classification & 15 & 64 & 1.0 & 7.0 & 1e-1 & $\text{0}^\star$ & 1e-2 & 3.0 & 10 & 1 \\
    $N$-Body Dynamics & 100 & 64 & 1.0 & 10.0 & 1.0 & $\text{0}^\star$ & 1e-3 & 2.0 & 10 & 1 \\
    Molecular Property Prediction & 500 & 125 & 1.0 & 3.0 & 1e-1 & 1.0 & 1e-3 & 3.0 & 3 & 6 \\
    Absence of Symmetry & 25 & 64 & 1.0 & 1.0 & 1e-1 & $\text{0}^\star$ & 1e-1 & 5.0 & 10 & 1 \\
    \bottomrule
    \end{tabular}
    }
\end{table}

\subsection{Inference}

At inference time, we compute final predictions by
\begin{equation}
    \hat{y}_i = \frac{1}{K+1} \Bigl( \hat{y}_{i,1} + \sum_{j=1}^{K} \rho_\mathcal{Y} (g^{-1}_{i,j}) \hat{y}_{i,j+1} \Bigr) ,
    \label{eq:inference}
\end{equation}
where $\rho_\mathcal{Y} (g^{-1}_{i,j})$  denotes the induced action on the output space: for equivariant tasks it corresponds to applying  $g^{-1}_{i,j}$ to the prediction, while for invariant tasks it reduces to the identity.
This inverse transformation is well-defined because the labels admit a corresponding group action, and because the equivariance loss in our training objective encourages the model to produce outputs that are consistent with the learned transformations.

We also explore a simplified alternative inference strategy in Appendix~\ref{app:Exploring_Alternative_Inference_Strategy}, which allows for faster but less robust predictions.

\subsection{Evaluation Metrics for Symmetry Discovery and Equivariance}
\label{app:add_details_metrics}

\noindent \textbf{Symmetry Discovery Metrics.}
To quantitatively evaluate the symmetry discovery results for LieAugmenter and the baseline methods, we compare the learned Lie algebra basis to a canonical ground-truth basis using two complementary metrics.
First, we report the absolute cosine similarity, which is a scale-invariant measure of the directional alignment of the basis to the ground truth.
Second, we report the absolute Frobenius projection of the learned basis $\hat{L}$ to the ground truth $L^\star$, given by 
$\frac{\langle \hat{L}, L^\star \rangle_F}{\lVert L^\star \rVert^2_F}$, which captures the scale alignment of the bases. 

Note that direction is typically more critical for correctness in symmetry discovery, since it determines whether the learned generators span the right tangent directions of the group action. 
However, the scale of the learned bases can have a significant effect on the diversity of augmentations that can be sampled using the described reparameterization technique with a fixed distribution.
Therefore, these two metrics combine to provide a global insight into both the correctness of the learned Lie algebra bases and their potential direct usefulness for downstream tasks.

For visualization, we match each learned generator to the closest canonical basis element and optionally flip its sign so that the matched pair has a positive Frobenius inner product. This improves interpretability when comparing against the ground truth. Importantly, permuting generators or flipping their signs does not change the induced augmentation distribution under the symmetric coefficient sampling used in our method (see Section~\ref{sec:background} and Appendix~\ref{app:Additional_Background}).

\noindent \textbf{Equivariance Error.}
We assess equivariance by measuring the consistency of model predictions under transformations sampled from the ground-truth symmetry group. Specifically, we compute
\begin{equation}
    \dfrac{1}{N} \sum_{i=1}^N \Big\lVert \dfrac{1}{K} \sum_{j=1}^K \rho_\mathcal{Y} (g_j) (\Psi(x_i)) - \dfrac{1}{K} \sum_{j=1}^K \Psi(\rho_\mathcal{X} (g_j)(x_i)) \Big\rVert_1, 
\end{equation}
which contrasts transform-then-predict with predict-then-transform, averaged over $K$ sampled group elements and $N$ data points.
To provide a more detailed characterization, we additionally report an alternative equivariance-error metric in Appendix~\ref{app:equiv_error2}.

\subsection{Image Classification}

In this experiment, we used the prediction network as a CNN with four convolutional filters and two interleaved max-pooling layers. 
Following \citet{cohenGroupEquivariantConvolutional2016},
, we also consider a group-equivariant CNN (GCNN) that is invariant to the $p4$ group, i.e., the set of planar translations composed with rotations by multiples of 90 degrees about any grid-aligned center.
Its underlying implementation is defined analogously to our base CNN model in terms of the number and type of layers, in order to facilitate the comparison. 
We note, however, that this invariance is enforced architecturally (hard-coded) and is restricted to a small discrete subgroup of the underlying continuous rotation group, which can limit performance when the true variability is continuous.

We construct a RotatedMNIST dataset by applying random rotations to MNIST images \citep{deng2012mnist}.
We maintain the default train-test splits of MNIST in the resulting dataset, but also create a validation set with $10\%$ of the training set samples. Thus, the final dataset consists of $54,000$ training samples, $6,000$ validation samples, and $10,000$ test samples.

\begin{minipage}{0.5\linewidth}
Regarding the results for this experiment, it is worth emphasizing an additional challenge in the out-of-distribution (OOD) setting beyond reduced data diversity.
Specifically, restricting the rotation range can increase the effective overlap between certain classes (e.g., digits 6 and 9) making them harder to distinguish.
This can lead to a somewhat counterintuitive outcome: a model that learns a more faithful symmetry (e.g., LieAugmenter) may still incur higher test-time confusion between these classes than a model that relies on an imperfect or coarser symmetry assumption (e.g., LieGAN-based augmentations), simply because the latter may not fully align samples in a way that amplifies the ambiguity.
Moreover, in our end-to-end setting, symmetry discovery and prediction are learned jointly. 
When the relevant symmetry is difficult to recover, the model may benefit from longer training to fully realize the gains from the learned augmentations (even if this is still typically less costly than sequentially training separate discovery and prediction stages used by some baselines). These factors help explain part of the performance gap between the in-distribution (ID) and OOD results reported in Table~\ref{tab:results_rotmnist}.
The effect is particularly evident in the class-wise error patterns, which we visualize via confusion matrices in Figure~\ref{fig:confusion_rotmnist}.
\end{minipage}
\hspace{3 mm}
\begin{minipage}{0.45\linewidth}
\centering
\includegraphics[width=\linewidth]{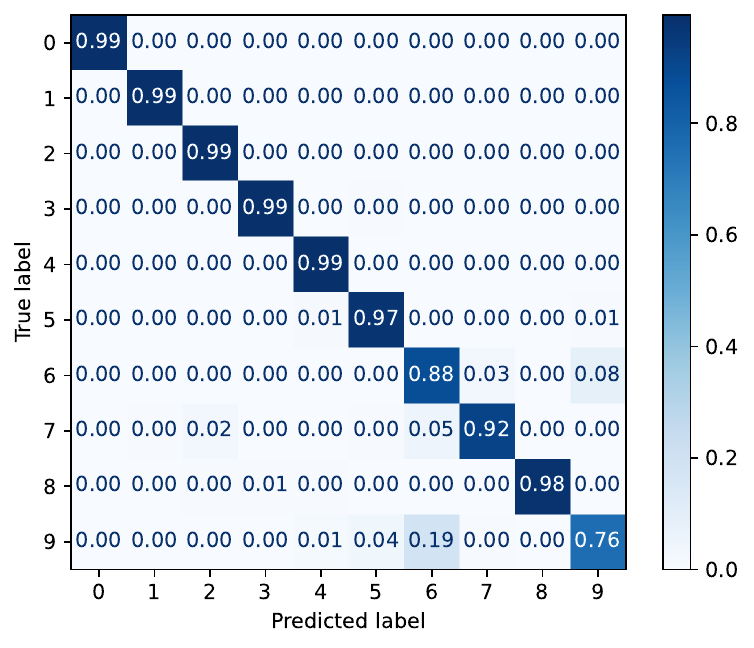}
\captionof{figure}{Confusion matrix for the test predictions of LieAugmenter in the OOD setting of the RotatedMNIST dataset.}
\label{fig:confusion_rotmnist}
\end{minipage}

\subsection{$N$-Body Dynamics Prediction}

For this experiment, we used a multilayer perceptron (MLP) as the prediction network for LieAugmenter and all augmentation-based baselines. The MLP comprises four fully connected layers with ReLU activations.
We follow the dataset construction and splits of \citet{yangGenerativeAdversarialSymmetry2023}, using 
$14,652$ training samples and 
$14,220$ test samples.
We also provide additional experiments on the $N$-body dynamic prediction in Appendix~\ref{app:nbody}, which covers a more challenging setup for 3-timestep prediction as well as the no mask variant. 

\subsection{Molecular Property Prediction}

In this experiment, we again use the standard QM9 train/validation/test split \citep{blum2009970, rupp2012fast}, consisting of $100,000$ training samples, $17,748$ validation samples, and $13,083$ test samples. 

We note that the results reported by \citet{bentonLearningInvariancesNeural2020} and \citet{yangGenerativeAdversarialSymmetry2023} were obtained under a slightly different training protocol.
For example, using a batch size of $75$ and employing a learning rate scheduling strategy.
Despite these differences, we expect the reported numbers to remain broadly comparable.

\subsection{Behavior in the Absence of Symmetry}

For this final synthetic experiment, we generate $50,000$ training samples, $10,000$ validation samples, and $10,000$ test samples.
The frozen MLP $f^\star$ consists of two hidden layers of width 64 with tanh activations, and we randomly sample its weights i.i.d. from a continuous distribution $\mathcal{N}(0, \sigma)$ with $\sigma = 0.01$.
\subsection{Scope and Future Work Beyond Linear Actions and the Exponential Map}

Our current implementation assumes that the group acts linearly on inputs and outputs via a representation. This assumption is in line with  Lie-based symmetry discovery and equivariant learning methods.
It makes the learned symmetries easy to interpret and compare. Importantly, it is a modeling choice rather than a fundamental restriction of the framework.

One natural extension is to handle projective group actions.
For example, $\mathrm{SL}(3,\mathbb{R})$ acting on the plane, $\mathrm{SL}(2,\mathbb{C})$ acting on the Riemann sphere.
In such cases, an embedding layer could lift inputs to homogeneous coordinates, where the group action becomes linear. LieAugmenter could then be applied in the lifted space, with an additional projection step to map outputs back to the original domain. This would support projective actions with relatively minor pre- and post-processing, while keeping our core symmetry-learning mechanism unchanged. More general non-linear actions without an explicit linearization step would likely require further architectural design and analysis, which we leave for future work.

Another direction is to consider alternative exponential maps that might mitigate some limitations of the current design. The Riemannian exponential is a natural candidate, and on certain manifolds it can have favorable global properties (e.g., surjectivity in some settings). However, these properties are not guaranteed in general, and the Riemannian exponential does not necessarily preserve the algebraic structure we rely on. For matrix Lie groups, when the metric is bi-invariant, the Riemannian exponential agrees with the Lie (matrix) exponential under the standard identification at the identity (see Appendix C in \citet{lezcano2019cheap}). More broadly, closed-form Riemannian exponentials are only available for particular groups and choices of invariant metrics \citep{zacur2014left}. This makes it less straightforward to use for symmetry discovery in our setting, since it can depend on geometric assumptions about the group that may not be known a priori.

That being said, exploring alternative exponential maps would require a careful treatment of both their algebraic and computational implications. In this work, our focus is on proposing task-adaptive symmetry discovery, i.e., learning symmetries end-to-end in a way that is directly driven by the downstream objective, under the standard and interpretable setting of linear group actions with the Lie (matrix) exponential. Extending this task-adaptive formulation to broader classes of group actions and alternative exponential constructions is an interesting direction to explore in future work.
\section{Additional Experiments}\label{app:additional_exp}

In this section, we present additional experimental results that further demonstrate the effectiveness of our method across a broader range of settings, spanning different datasets, symmetry structures, and model classes.

Specifically, we study an alternative inference procedure and find that performance remains robust. We provide evidence of empirical convergence, along with an ablation study and a hyperparameter sensitivity analysis indicating stable behavior across a wide range of configurations. We also evaluate sample efficiency.

We further extend the $N$-body experiments by considering symmetry discovery without a mask, as well as different masked variants, and we additionally report results for three timestep prediction. We include a runtime analysis and examine model behavior in settings where no meaningful symmetry is present.

Finally, we demonstrate that the proposed framework can also recover discrete symmetries, including discrete rotations and partial permutations, and we showcase applications across multiple domains, including Lorentz symmetries in quantum field theory amplitude prediction and human colorectal cancer data.

\subsection{Exploring Alternative Inference Strategy}\label{app:Exploring_Alternative_Inference_Strategy}

We study a faster inference variant in which no augmentations are generated at test time. In this setting, we feed only the original inputs through the prediction network and output its direct prediction. This option can be preferable in applications where inference latency is a primary constraint.

To evaluate the impact of this choice, we rerun the RotatedMNIST image classification and apply the no-test-augmentation strategy to LieAugmenter and the fixed augmentation strategies with oracle and LieGAN symmetries.
As shown in Tables~\ref{tab:results_rotmnist_app} and \ref{tab:time_rotmnist_alt_inf}, accuracy decreases slightly compared to the aggregation-based inference used in the main results (Table~\ref{tab:results_rotmnist}), while the overall test-time cost is correspondingly reduced (Table~\ref{tab:time_rotmnist}).

\begin{table*}[h]
\caption{RotatedMNIST results for in-distribution (ID) and out-of-distribution (OOD) evaluation following the alternative inference strategy. We report classification accuracy (higher is better) and equivariance error (lower is better) for CNN models trained (i) with \emph{Oracle} rotation augmentation (ground-truth symmetry), (ii) with LieGAN augmentations, and (iii) with LieAugmenter (\emph{Ours}). All values are reported as $\mathrm{mean}\pm\mathrm{std}$ over three random seeds.}
  \label{tab:results_rotmnist_app}
  \centering
  \resizebox{0.7\textwidth}{!}{%
    \begin{tabular}{lccc} 
    \toprule
    & Oracle augmentation            & \multicolumn{2}{c}{Symmetry discovery}                           \\ 
    \cmidrule(lr){2-2}\cmidrule(l){3-4}
    & Rotation                       & LieGAN Aug.                    & LieAugmenter (Ours)             \\ 
    \midrule
    ID Acc. (\%) $(\uparrow)$       & $98.45 \scriptstyle{\pm 0.11}$ & $97.84 \scriptstyle{\pm 0.13}$ & $98.55 \scriptstyle{\pm 0.08}$  \\
    ID Equiv. Error $(\downarrow)$  & $3.91 \scriptstyle{\pm 0.23}$  & $4.92 \scriptstyle{\pm 0.74}$  & $3.72 \scriptstyle{\pm 0.15}$   \\ 
    \midrule
    OOD Acc. (\%) $(\uparrow)$      & $98.35 \scriptstyle{\pm 0.11}$ & $84.97 \scriptstyle{\pm 0.43}$ & $93.40 \scriptstyle{\pm 2.46}$  \\
    OOD Equiv. Error $(\downarrow)$ & $4.03 \scriptstyle{\pm 0.17}$  & $6.80 \scriptstyle{\pm 0.85}$  & $3.40 \scriptstyle{\pm 0.47}$   \\
    \bottomrule
    \end{tabular}
    }
\end{table*}

\subsection{Empirical Convergence}
\label{app:empirical_convergence}

\begin{minipage}{0.45\textwidth}
In order to complement the previous analysis of the convergence of our reparameterized objective presented in Appendix~\ref{app:convergence_analysis}, we carry out an empirical exploration of convergence in our experiments, seeking to provide greater insight and evidence of the applicability of our proposed approach in practice. 
We report the train and validation loss curves as well as the corresponding separate evolution of the main loss terms of the train loss (i.e., objective/empirical loss and equivariance loss) during training for RotatedMNIST (Figure~\ref{fig:rotmnist_losses}) and QM9 (Figure~\ref{fig:qm9_losses}). We see that all losses quickly decrease and start converging despite the complexity of the associated optimization problem.
\end{minipage}
\hfill
\begin{minipage}{0.5\textwidth}
  \includegraphics[width=\linewidth]{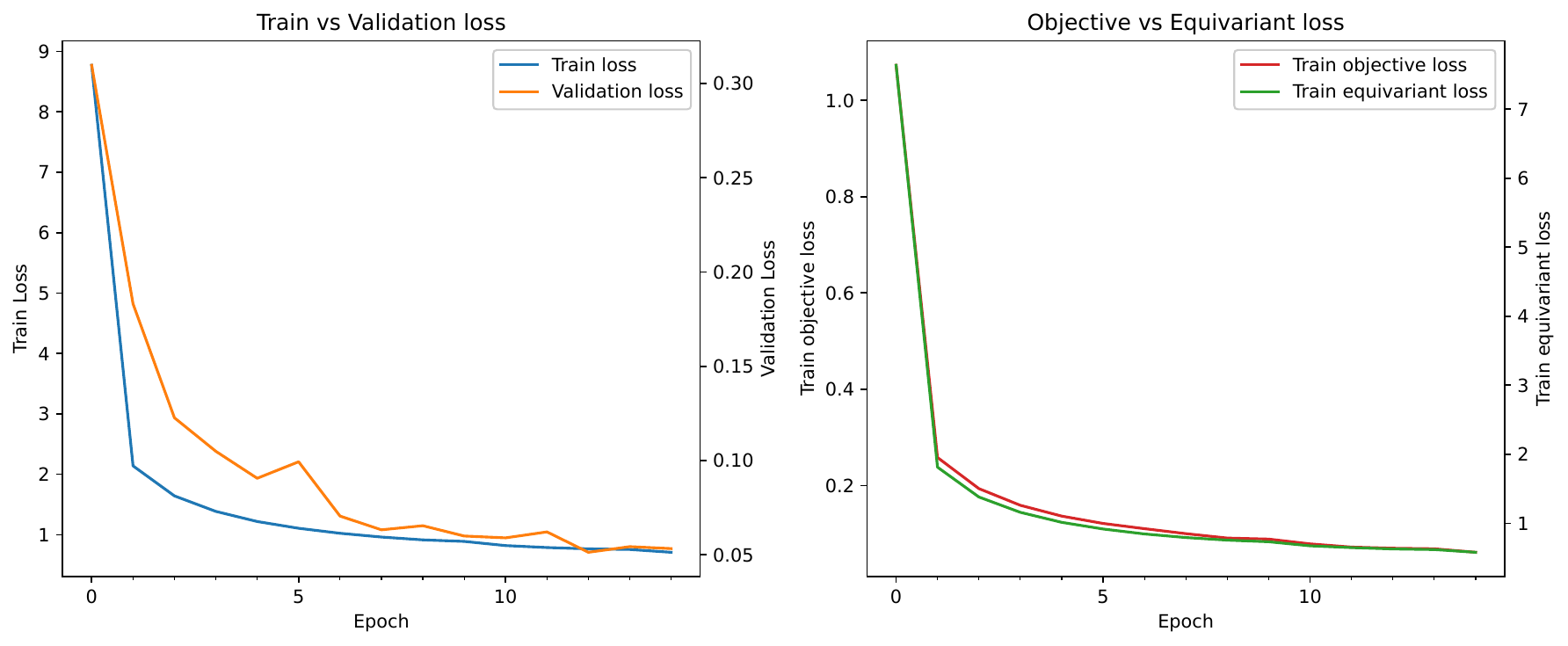}
  \captionof{figure}{Loss curves of LieAugmenter for RotatedMNIST dataset. (Left) Train and validation losses. (Right) Objective and equivariance losses.}
  \label{fig:rotmnist_losses}
  
  \includegraphics[width=\linewidth]{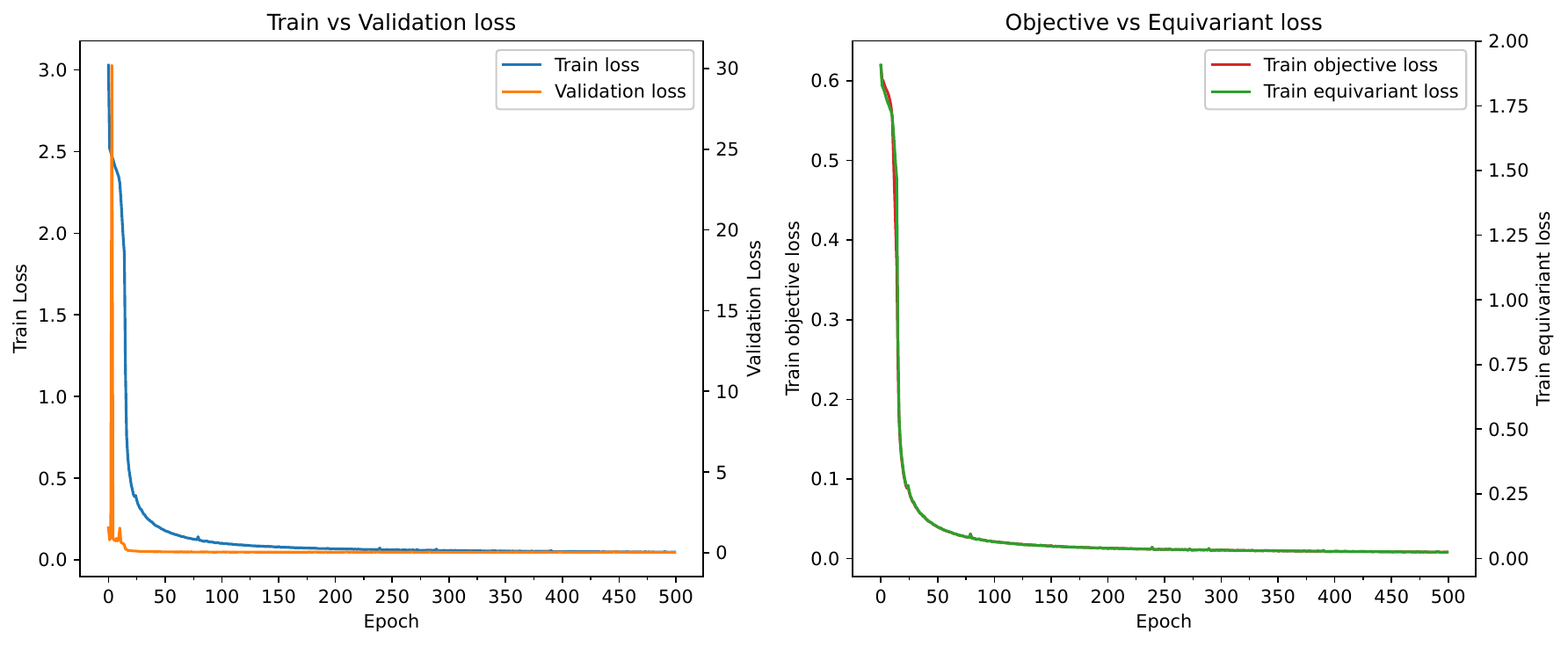}
  \captionof{figure}{Loss curves for LUMO task of the QM9 dataset. (Left) Train and validation losses. (Right) Objective and equivariance losses.}
  \label{fig:qm9_losses}
\end{minipage}

\subsection{Ablation Study and Hyperparameter Analysis}\label{app:ablation}

We conduct a set of ablations over LieAugmenter’s loss weights and key hyperparameters to assess how each component affects performance. For this analysis, we focus on RotatedMNIST, since it has the lowest runtime among our symmetry-based benchmarks and therefore supports a more extensive sweep.

Note that $\alpha$ is not a strictly necessary weight coefficient, given that setting $\alpha = \beta$ would be approximately equivalent to merging the empirical prediction loss $\mathcal{L}_{emp}$ into the equivariance loss $L_{equiv}$ by changing the averaging over augmentations to also include the original sample. Nevertheless, we consider a separate weight to enhance the flexibility of our method and provide a more rigorous analysis of its impact on the performance.

\begin{figure}[h]
  \centering
  \includegraphics[width=0.48\textwidth]{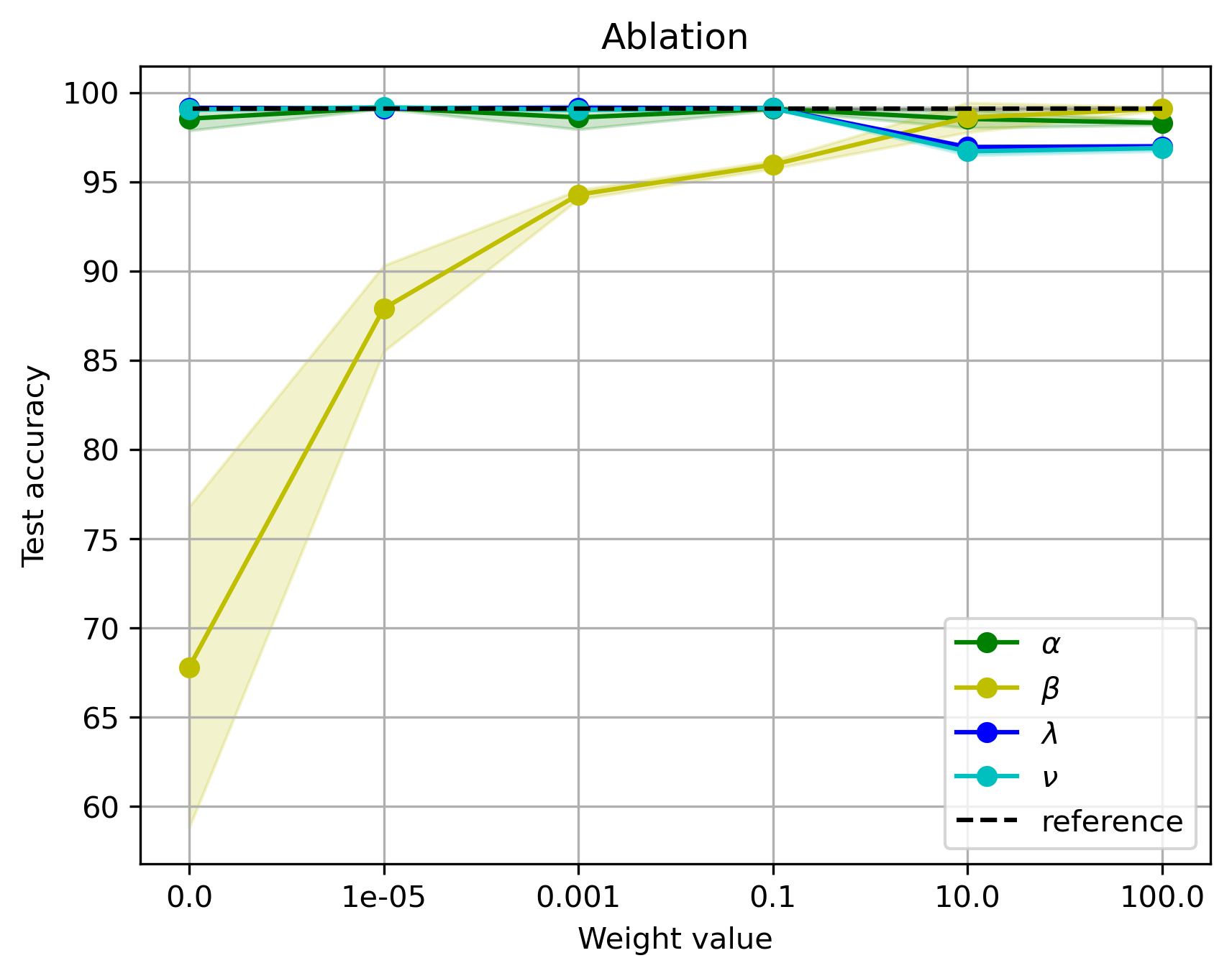}
  \includegraphics[width=0.48\textwidth]{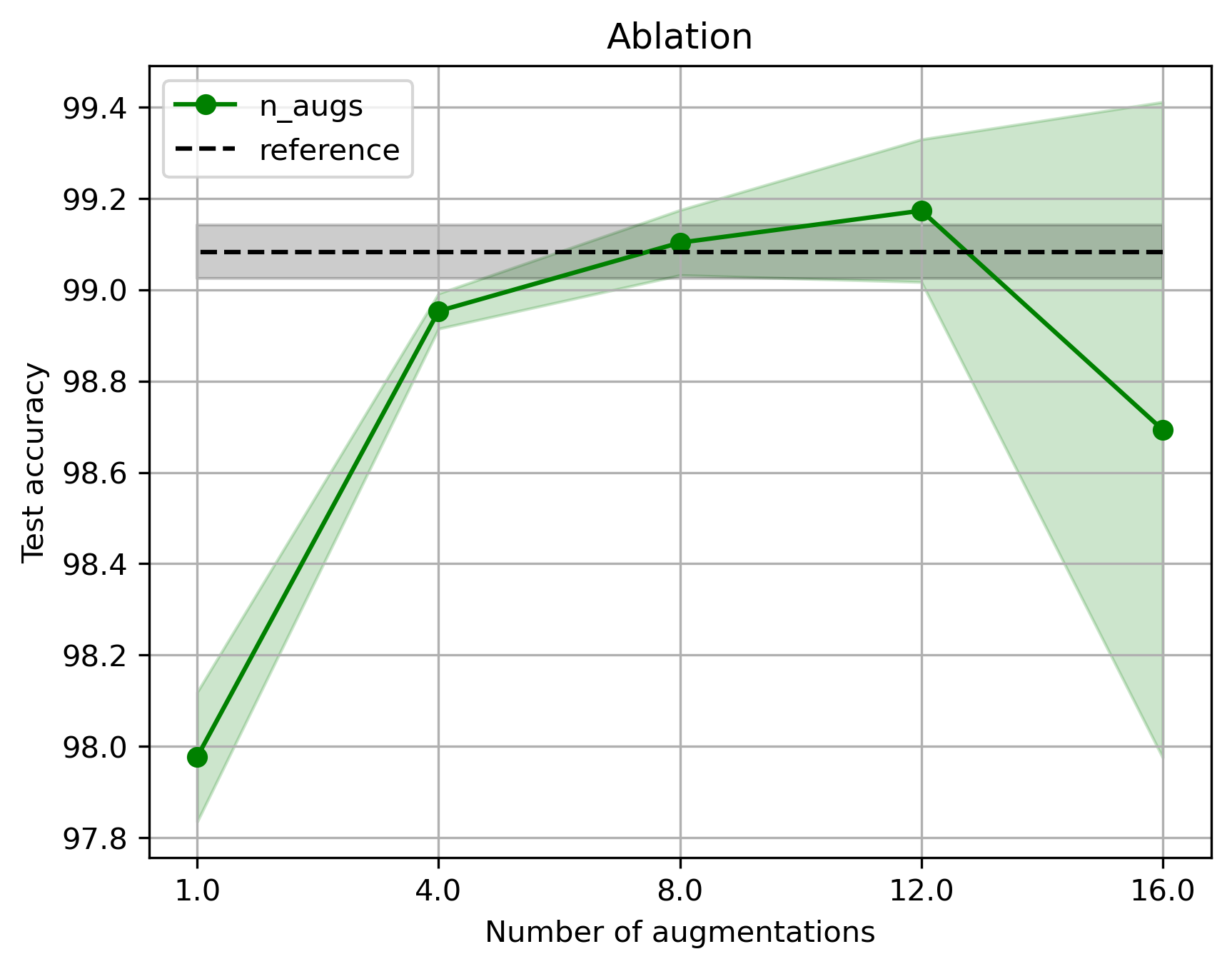}
  \caption{Ablation results for the accuracy metric in the RotatedMNIST dataset for varying values for the weight terms of the loss function (left) and the number of augmentations per sample (right). The reference results correspond to the hyperparameters considered in the main experiments. All results are reported as the mean and standard deviation over three runs with different random seeds.}
  \label{fig:ablation_metric}
\end{figure}

From Figure~\ref{fig:ablation_metric} (left), we observe that the performance of our model is quite robust to changes in the values of the weights for the different terms of the loss function except for $\beta$, i.e., the weight for the $\mathcal{L}_{equiv}$ term. 
In particular, decreasing the value of $\beta$ leads to a degraded performance, as expected, since the LieAugmenter will not receive a strong enough learning signal to properly learn the correct symmetry.
In addition, in Figure~\ref{fig:ablation_metric} (right), we see that the number of sampled augmentations per input sample can also considerably affect the performance of the model, although not so drastically. 
In particular, increasing the number of augmentations generally leads to an improved performance, as we could expect, but after a certain point, the performance starts decreasing and becomes more variable across runs. 
A plausible explanation is that when the number of augmented instances becomes large relative to the batch size, optimization becomes dominated by augmented samples, effectively down-weighting the original data and making training less stable.

\begin{figure}[h]
  \centering
  \includegraphics[width=0.48\textwidth]{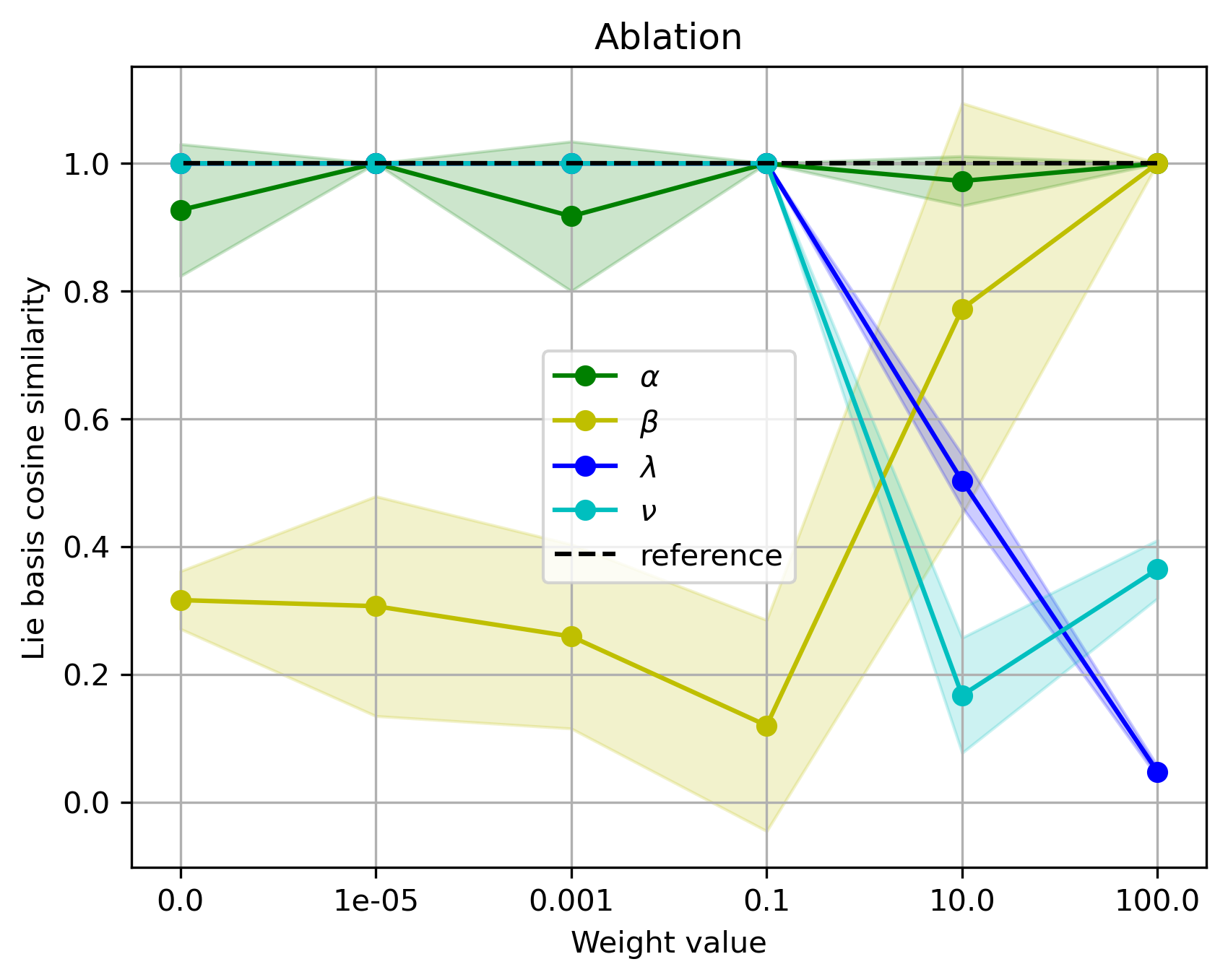}
  \includegraphics[width=0.48\textwidth]{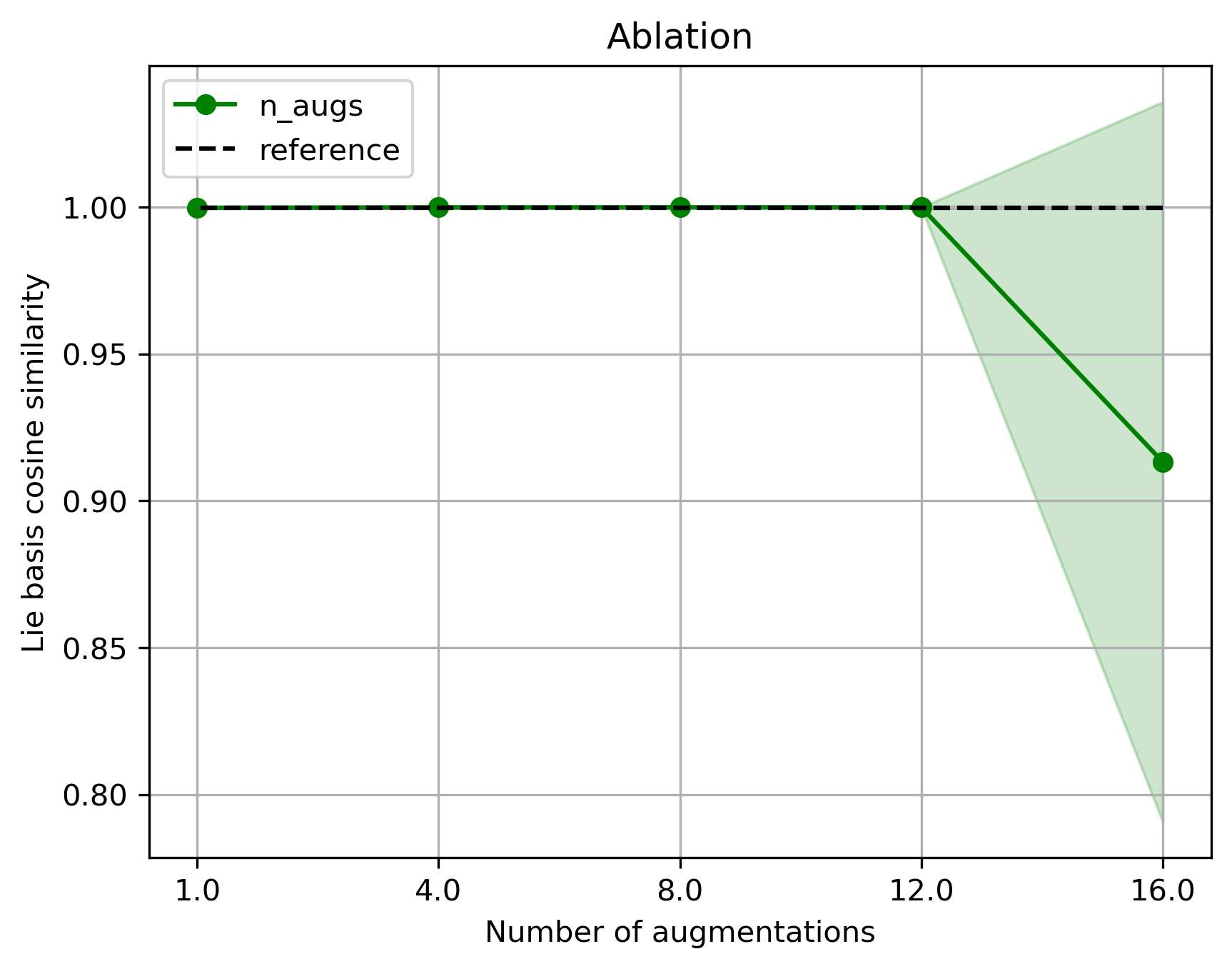}
  \caption{Ablation results for the cosine similarity of the learned Lie algebra basis to the ground truth in the RotatedMNIST dataset for varying values for the weight terms of the loss function (left) and the number of augmentations per sample (right). The reference results correspond to the hyperparameters considered in the main experiments. All results are reported as the mean and standard deviation over three runs with different random seeds.}
  \label{fig:ablation_lie}
\end{figure}

In Figure~\ref{fig:ablation_lie}, we can find the corresponding results for the ablation analysis focusing on the cosine similarity of the learned Lie algebra basis to the ground truth instead of the performance metric.
For the weight terms of the loss function, we can see that modifying $\alpha$ (the weight of $\mathcal{L}_{obj}$) has little effect on the symmetry discovery process. Modifying $\lambda$ and $\nu$ also has little impact below a threshold of $0.1$, but when those terms are increased too much, they lead to a degraded symmetry discovery performance, as the difference of the augmentations from the original data ($\lambda$) or the sparsity of the learned basis ($\nu$) become too prominent in the loss.
Furthermore, $\beta$ also has a relevant impact on the symmetry discovery process, where we observe that higher values of that weight help the performance of the model.
Lastly, the number of augmentations per sample has little impact on the symmetry discovery process for values below $16$, but at that point the performance degrades, possibly for the same reason that we hypothesized for the performance metric.

\begin{minipage}{0.57\textwidth}
Next, we also explore the effect of the chosen range $\gamma$ for the uniform sampling of coefficients for the linear combination of elements of the Lie algebra basis, which is then mapped to a group element. This is a hyperparameter that we did not extensively tune in our experiments, and its values were chosen mainly seeking to encourage the sampling of diverse elements of the group while not being excessively large to avoid distortions in the uniformity of the sampling (e.g., for rotation groups the sampled elements would ``wrap around'' for larger ranges, thus distorting the effective sampling density of the group elements).
\\

In order to verify our intuition and solidify our explanation in that regard, we run an additional ablation experiment for the $\gamma$ hyperparameter using the RotatedMNIST dataset, the results of which are shown in Table~\ref{tab:ablation_gamma}. In these results, we would start observing the ``wrap around'' effect starting at $\gamma=5.0$ if the model were to learn the ground truth symmetry, which may explain the decrease in performance and cosine similarity for that and the larger values of this hyperparameter. In the case of $\gamma=13.0$, we hypothesize that the slight increase in the performance and similarity of the basis is due to the fact that learning the ground truth Lie algebra basis would lead to rotations approximately in the range $[-2\pi, 2\pi]$, thus circumventing the issue.
\end{minipage}
\hfill
\hspace{2 mm}
\begin{minipage}{0.4\textwidth}
  \captionof{table}{Results for the ablation of $\gamma$. We report the mean and standard deviation for both metrics across three runs using different random seeds. The accuracies are expressed as percentages.} 
  \label{tab:ablation_gamma}
  \centering
  \resizebox{0.9\linewidth}{!}{%
    \begin{tabular}{lcc}
    \toprule
    $\gamma$ & Test accuracy & Cosine similarity basis \\
    \midrule
    1.0 & $97.73 \scriptstyle{\pm 0.17}$ & $0.9999 \scriptstyle{\pm 0.0001}$ \\
    2.0 & $98.04 \scriptstyle{\pm 0.22}$ & $1.0000 \scriptstyle{\pm 0.0000}$ \\
    3.0 & $99.08 \scriptstyle{\pm 0.06}$ & $0.9997 \scriptstyle{\pm 0.0002}$ \\
    5.0 & $95.97 \scriptstyle{\pm 3.24}$ & $0.7860 \scriptstyle{\pm 0.3026}$ \\
    8.0 & $93.11 \scriptstyle{\pm 3.79}$ & $0.5706 \scriptstyle{\pm 0.3571}$ \\
    9.0 & $94.93 \scriptstyle{\pm 4.94}$ & $0.6783 \scriptstyle{\pm 0.4549}$ \\
    10.0 & $92.56 \scriptstyle{\pm 4.19}$ & $0.6273 \scriptstyle{\pm 0.2656}$ \\
    11.0 & $89.28 \scriptstyle{\pm 2.96}$ & $0.3709 \scriptstyle{\pm 0.2551}$ \\
    12.0 & $96.74 \scriptstyle{\pm 2.14}$ & $0.6881 \scriptstyle{\pm 0.4381}$ \\
    13.0 & $96.98 \scriptstyle{\pm 0.72}$ & $0.9812 \scriptstyle{\pm 0.0178}$ \\
    14.0 & $93.32 \scriptstyle{\pm 2.99}$ & $0.7236 \scriptstyle{\pm 0.2095}$ \\
    15.0 & $91.89 \scriptstyle{\pm 2.20}$ & $0.5346 \scriptstyle{\pm 0.0607}$ \\
    16.0 & $93.52 \scriptstyle{\pm 0.91}$ & $0.2104 \scriptstyle{\pm 0.1597}$ \\
    17.0 & $90.28 \scriptstyle{\pm 3.33}$ & $0.2779 \scriptstyle{\pm 0.1739}$ \\
    21.0 & $89.00 \scriptstyle{\pm 4.13}$ & $0.0006 \scriptstyle{\pm 0.0006}$ \\
    \bottomrule
    \end{tabular}
    }
\end{minipage}

\begin{figure}[h]
  \centering
  \includegraphics[width=0.8\textwidth]{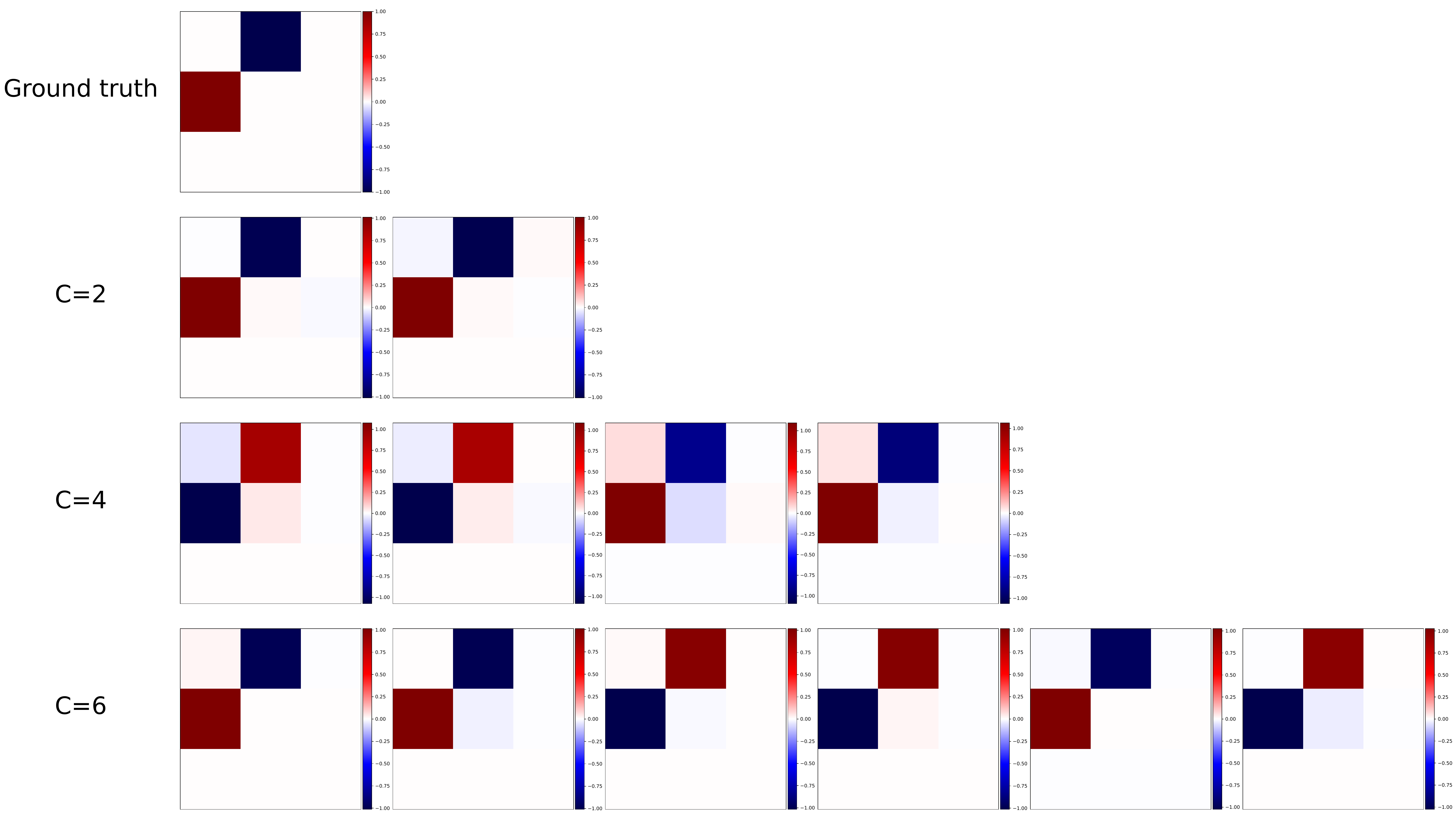}
  \caption{LieAugmenter is able to correctly discover a representation of the underlying symmetry even when the cardinality of the Lie algebra basis is misspecified in the RotatedMNIST dataset.}
  \label{fig:ablation_cardinality}
\end{figure}

Finally, we also explore the symmetry discovery performance of LieAugmenter under a misspecified Lie algebra basis cardinality $C$ in Figure~\ref{fig:ablation_cardinality}. As we can see, under the slight misspecification of setting $C=2$, we observe redundancy across the different elements of the learned basis, which shows that our model is able to robustly discover the correct basis. 
In the more extreme cases of setting $C=4$ or $C=6$, we start observing a slight degradation in the discovered basis, which is somewhat noisier. In addition, while some elements in the basis are inverses of one another, their linear combination would still lead to a valid rotation augmentation.
Therefore, while knowing the correct (or a close approximation) of the basis cardinality a priori helps the symmetry discovery performance of the model, we have shown that it can be robust in scenarios where that knowledge may not be available.

\subsection{Sample efficiency}

\begin{minipage}{0.5\textwidth}
We explore the sample efficiency of LieAugmenter when compared to the other considered prediction approaches by measuring their test performance when trained with a varying number of samples in the training set. As we can see in Figure~\ref{fig:sample_efficiency_rotmnist}, LieAugmenter matches the performance of the CNN trained with ground truth augmentations up to training with 50\% of the data, after which its performance decreases but remains higher than that of the other baselines. 
\end{minipage}
\hfill
\begin{minipage}{0.4\textwidth}
  \centering
  \includegraphics[width=\linewidth]{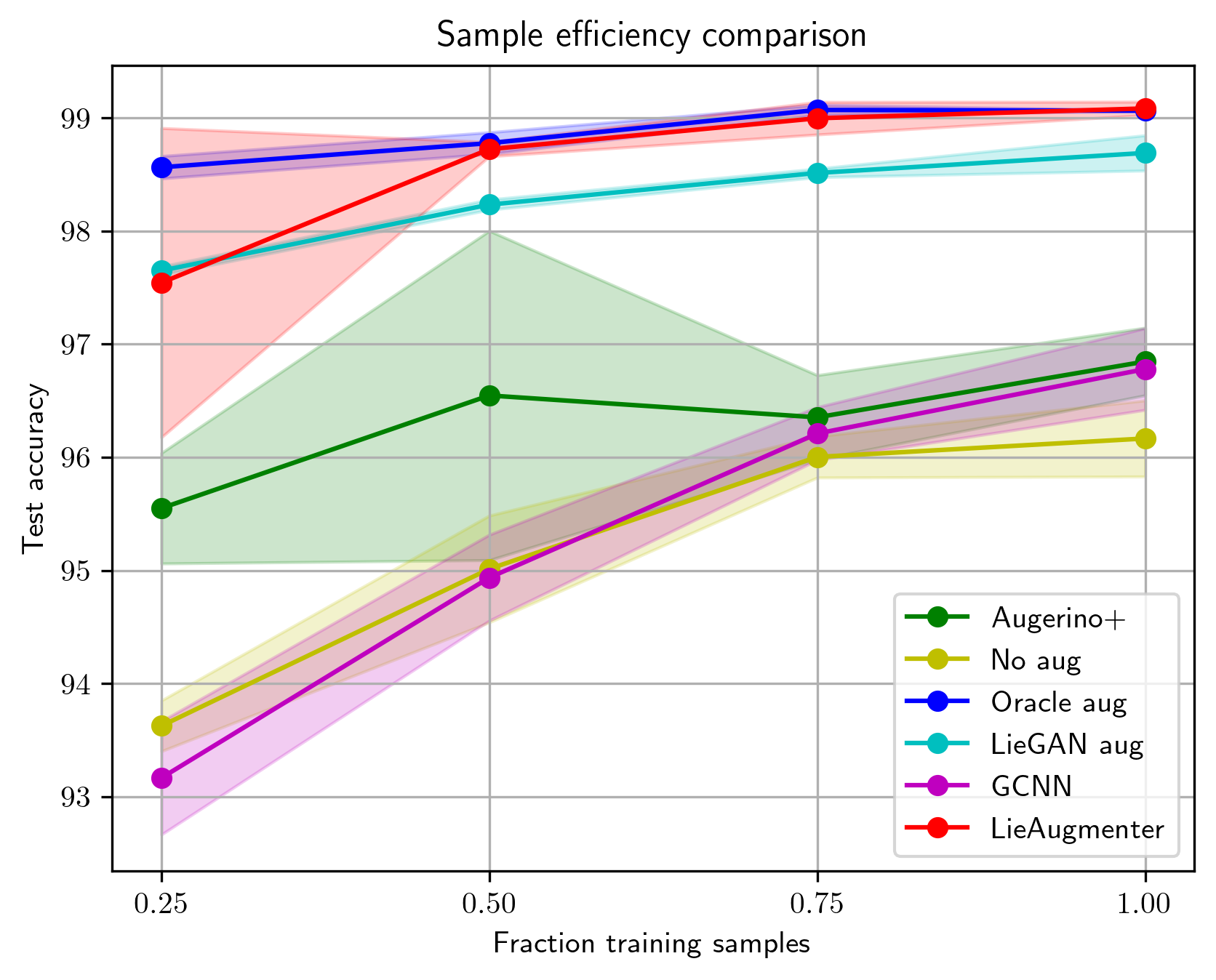}
  \captionof{figure}{Comparison of the sample efficiency for the equivariant-architecture baseline (GCNN, $p4$) and CNN models trained (i) without augmentation (No aug), (ii) with ground truth rotation augmentation (Oracle aug), (iii) with symmetry-discovery baselines (Augerino+ and LieGAN aug), and (iv) with LieAugmenter (\emph{Ours}), which are trained with different fractions of the total training set samples of the RotatedMNIST dataset.}
  \label{fig:sample_efficiency_rotmnist}
\end{minipage}

\subsection{$N$-Body Dynamics Prediction}
\label{app:nbody}

In this section, we provide a deeper exploration of the performance of LieAugmenter and the baseline methods for the task of $N$-body dynamics prediction. 
On one hand, we analyze the symmetry discovery behavior under more relaxed assumptions regarding the search space for the symmetry in the form of different masks for the learnable Lie algebra basis.
On the other hand, we study the more challenging scenario of having access to deeper time horizon for both the inputs and the predictions, in particular consisting of three timesteps per sample, instead of the single timestep considered in our main results. 

\subsubsection{$1$-Timestep Prediction}

As described in the paper, previous symmetry discovery methods considered some auxiliary knowledge about this task in order to restrict the search space for the symmetry. In particular, such prior assumptions led to the definition of a $2 \times 2$ block diagonal mask that restricts which entries of the parameterized Lie algebra basis are learnable.

We denote that mask as $\mathrm{Mask}\_2$ and study the effect of considering more relaxed prior assumptions on the symmetry search space.
First, we consider a mask that allows interactions between only positions and only momenta across the different bodies ($\mathrm{Mask}\_4$). 
In addition, we explore the effect of removing all prior constraints on the search space ($\mathrm{Mask}\_0$). 
We add the same numerical suffixes to LieAugmenter and the baseline models to indicate which mask is applied in each case.

\begin{figure}[h]
  \centering
  \includegraphics[width=0.45\textwidth]{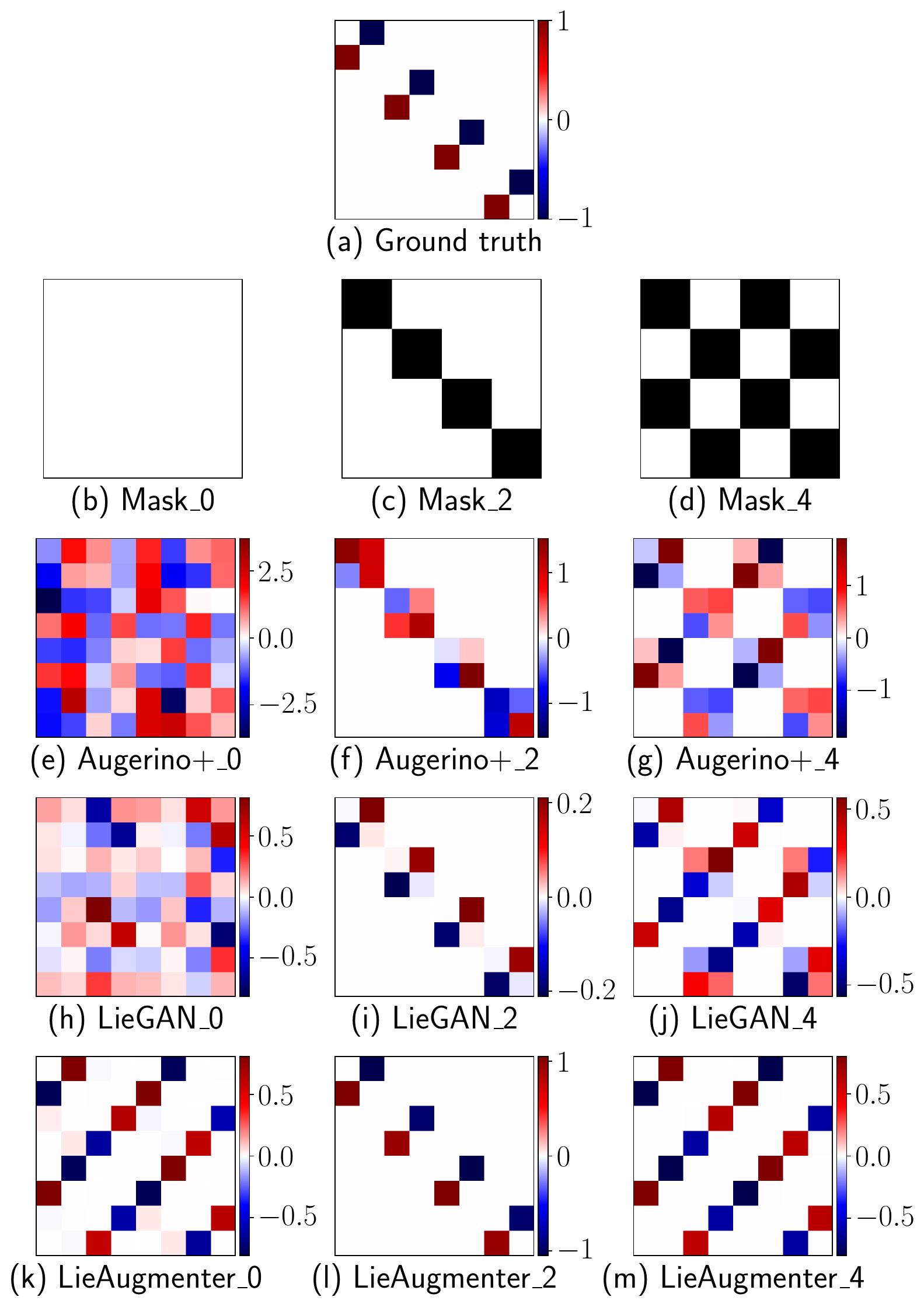}
  \includegraphics[width=0.45\textwidth]{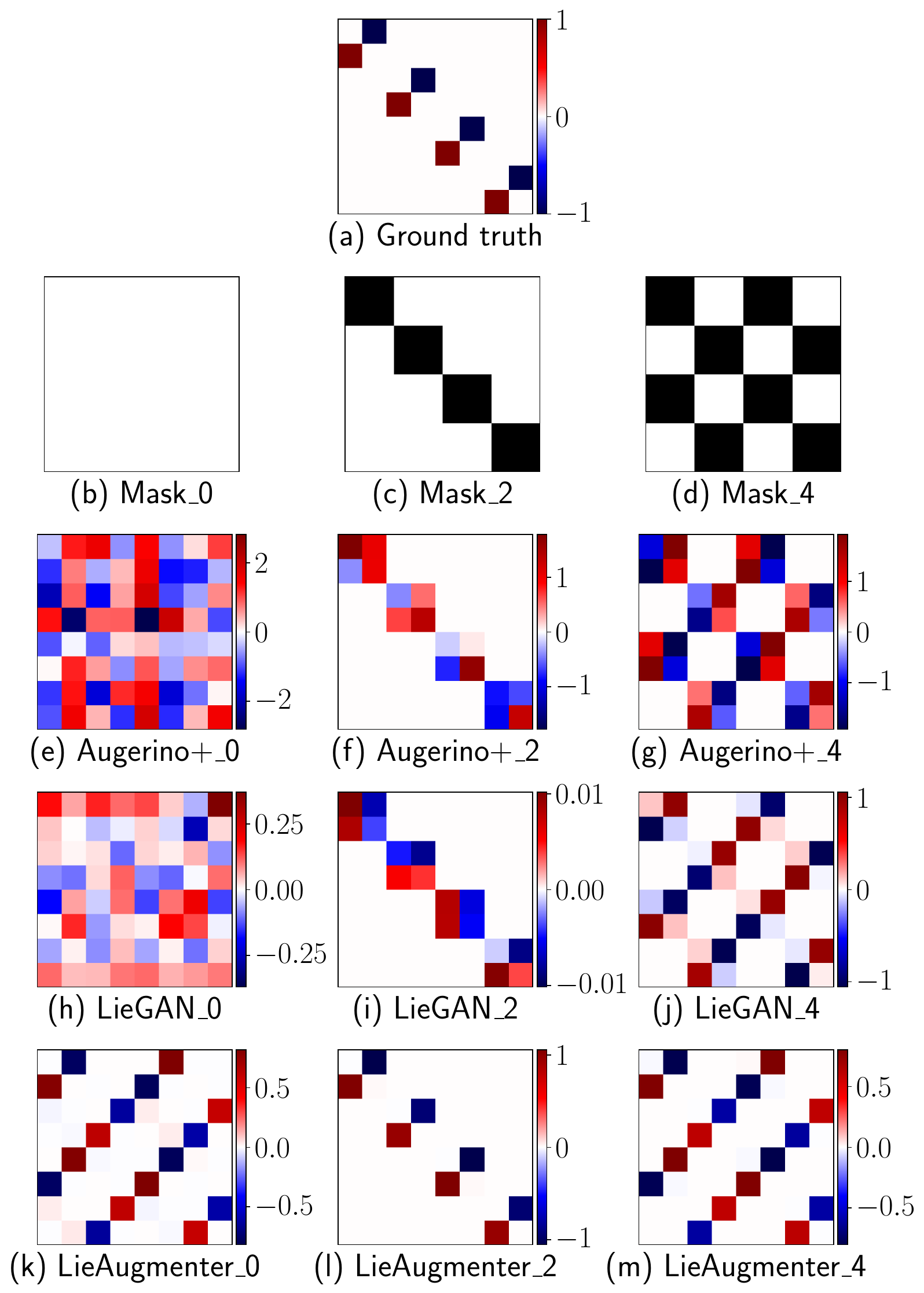}
  \caption{
   Ground-truth Lie algebra basis, masks of learnable entries, and bases learned by baseline symmetry discovery methods and LieAugmenter for 2-body dynamics prediction with one input and one output timestep. Results are shown for the ID (\textit{left}) and OOD settings (\textit{right}).
  }
  \label{fig:lie_algebras_nbody2_all}
\end{figure}

\paragraph{{Symmetry Discovery.}}  Figure~\ref{fig:lie_algebras_nbody2_all} shows a visualization of the different masks, as well as the learned Lie algebra basis by each of the models in the in-distribution and out-of-distribution scenarios.
As we can observe, when relaxing the constraints on the symmetry search space, LieAugmenter learns a Lie algebra basis that differs from the canonical ground truth that is identified under the $\mathrm{Mask}\_2$ setup. However, we can show that for the physical parameters considered in the definition of this dataset, that basis actually corresponds to an alternative representation of the rotational symmetry defined by the ground truth. Please refer to Appendix~\ref{sec:alternative_rep_nbody2} for further details.
As a result, we can conclude that LieAugmenter successfully recovers the correct symmetry under all constraint masks and also the lack thereof in both in-distribution and out-of-distribution scenarios.

\paragraph{{Predictive Performance.}} 
Similarly, we extend our previous performance analysis by including the models that incorporate the symmetry discovery results obtained with different masks. In addition, we also add the results for an equivariant model, EMLP \citep{finzi2021practical}, which is defined to be equivariant with respect to the ground truth Lie algebra basis.
Table~\ref{tab:results_nbody2_app} reports the MSE and equivariance error of the different approaches in the ID and OOD settings. As we can observe, regardless of the mask, LieAugmenter achieves the best performance only (closely) behind the equivariant model, but still surpassing the methods that consider oracle augmentation and augmentation based on discovered symmetries. In addition, note that LieAugmenter is considerably faster to train than the equivariant model (requiring almost a third of the training time), as we present in Table~\ref{tab:times_nbody2}. 

\begin{table}[h!]
  \caption{$2$-body dynamics results for ID and OOD evaluations under different masks restrictions on the symmetry search spacef for one input and output timestep. We report MSE and equivariance error for an equivariant architecture (EMLP) and for MLP models trained (i) without augmentation (\emph{Trivial}), (ii) with \emph{Oracle} rotation augmentation (ground-truth symmetry), (iii) with symmetry-discovery baselines (Augerino+ and LieGAN Aug.), and (iv) with LieAugmenter (\emph{Ours}). All values are reported as $\mathrm{mean}\pm\mathrm{std}$ over three random seeds.}
  \label{tab:results_nbody2_app}
  \centering
\resizebox{\textwidth}{!}{%
    \begin{tabular}{clcccc}
\toprule
& & \multicolumn{2}{c}{In-distribution} & \multicolumn{2}{c}{Out-of-distribution} \\
\cmidrule{3-6}
& & MSE $(\downarrow)$ & Equiv. Error $(\downarrow)$ & MSE $(\downarrow)$ & Equiv. Error $(\downarrow)$ \\
\midrule
\multirow{1}{*}{\makecell{Equiv. architecture}} & EMLP & $\text{1.74e-06} \scriptstyle{{\pm \text{1.50e-06}}}$ & $\text{5.32e-05} \scriptstyle{{\pm \text{1.16e-05}}}$ & $\text{4.64e-06} \scriptstyle{{\pm \text{4.86e-06}}}$ & $\text{5.27e-05} \scriptstyle{{\pm \text{1.10e-05}}}$ \\
\midrule
\multirow{2}{*}[-0.15em]{\makecell{Data\\augmentation}} & Trivial & $\text{5.16e-03} \scriptstyle{{\pm \text{2.79e-03}}}$ & $\text{2.48e-02} \scriptstyle{{\pm \text{7.37e-03}}}$ & $\text{1.86e-02} \scriptstyle{{\pm \text{3.74e-03}}}$ & $\text{3.33e-02} \scriptstyle{{\pm \text{2.56e-03}}}$ \\
& Oracle & $\text{2.13e-05} \scriptstyle{{\pm \text{1.56e-05}}}$ & $\text{3.03e-03} \scriptstyle{{\pm \text{1.03e-03}}}$ & $\text{3.03e-05} \scriptstyle{{\pm \text{8.81e-06}}}$ & $\text{3.70e-03} \scriptstyle{{\pm \text{1.14e-03}}}$ \\
\midrule
\multirow{9}{*}[-0.15em]{\makecell{Symmetry\\discovery}} & Augerino+\_0 & $\text{5.28e+00} \scriptstyle{{\pm \text{4.23e-01}}}$ & $\text{1.90e-01} \scriptstyle{{\pm \text{1.92e-03}}}$ & $\text{3.26e+00} \scriptstyle{{\pm \text{7.30e-01}}}$ & $\text{1.94e-01} \scriptstyle{{\pm \text{4.15e-03}}}$ \\
& LieGAN\_0 aug & $\text{3.23e-03} \scriptstyle{{\pm \text{6.25e-04}}}$ & $\text{1.91e-02} \scriptstyle{{\pm \text{1.47e-03}}}$ & $\text{1.15e-02} \scriptstyle{{\pm \text{3.00e-03}}}$ & $\text{2.84e-02} \scriptstyle{{\pm \text{3.04e-03}}}$ \\
& LieAugmenter\_0 & $\text{1.39e-05} \scriptstyle{{\pm \text{6.08e-07}}}$ & $\text{2.78e-03} \scriptstyle{{\pm \text{4.88e-04}}}$ & $\text{2.38e-05} \scriptstyle{{\pm \text{9.04e-06}}}$ & $\text{2.59e-03} \scriptstyle{{\pm \text{9.89e-05}}}$ \\
\cmidrule{2-6}
& Augerino+\_2 & $\text{9.67e+03} \scriptstyle{{\pm \text{1.37e+04}}}$ & $\text{2.37e-01} \scriptstyle{{\pm \text{7.53e-02}}}$ & $\text{1.48e+05} \scriptstyle{{\pm \text{2.09e+05}}}$ & $\text{1.95e-01} \scriptstyle{{\pm \text{9.23e-03}}}$ \\
& LieGAN\_2 aug & $\text{1.10e-03} \scriptstyle{{\pm \text{6.98e-04}}}$ & $\text{1.48e-02} \scriptstyle{{\pm \text{4.80e-03}}}$ & $\text{8.94e-03} \scriptstyle{{\pm \text{9.31e-04}}}$ & $\text{2.73e-02} \scriptstyle{{\pm \text{1.10e-03}}}$ \\
& LieAugmenter\_2 & $\text{1.30e-05} \scriptstyle{{\pm \text{5.58e-06}}}$ & $\text{3.28e-03} \scriptstyle{{\pm \text{9.36e-04}}}$ & $\text{3.73e-05} \scriptstyle{{\pm \text{2.97e-05}}}$ & $\text{5.96e-03} \scriptstyle{{\pm \text{2.82e-03}}}$ \\
\cmidrule{2-6}
& Augerino+\_4 & $\text{1.79e+00} \scriptstyle{{\pm \text{2.77e-02}}}$ & $\text{1.87e-01} \scriptstyle{{\pm \text{2.92e-04}}}$ & $\text{1.55e+00} \scriptstyle{{\pm \text{8.59e-03}}}$ & $\text{1.86e-01} \scriptstyle{{\pm \text{5.90e-04}}}$ \\
& LieGAN\_4 aug & $\text{2.85e-05} \scriptstyle{{\pm \text{1.19e-05}}}$ & $\text{3.02e-03} \scriptstyle{{\pm \text{5.08e-04}}}$ & $\text{5.12e-05} \scriptstyle{{\pm \text{8.47e-06}}}$ & $\text{2.78e-03} \scriptstyle{{\pm \text{3.62e-04}}}$ \\
& LieAugmenter\_4 & $\text{1.45e-05} \scriptstyle{{\pm \text{6.15e-06}}}$ & $\text{3.15e-03} \scriptstyle{{\pm \text{9.78e-04}}}$ & $\text{1.61e-05} \scriptstyle{{\pm \text{1.22e-06}}}$ & $\text{3.13e-03} \scriptstyle{{\pm \text{2.79e-04}}}$ \\
\bottomrule
\end{tabular}
}
\end{table}

\subsubsection{$3$-Timestep Prediction}

In this section, we follow the same analysis as in the previous one but we study the more challenging task of predicting three future timesteps from three past observations of the trajectory of the bodies.

\begin{figure}[h]
  \centering
  \includegraphics[width=0.45\textwidth]{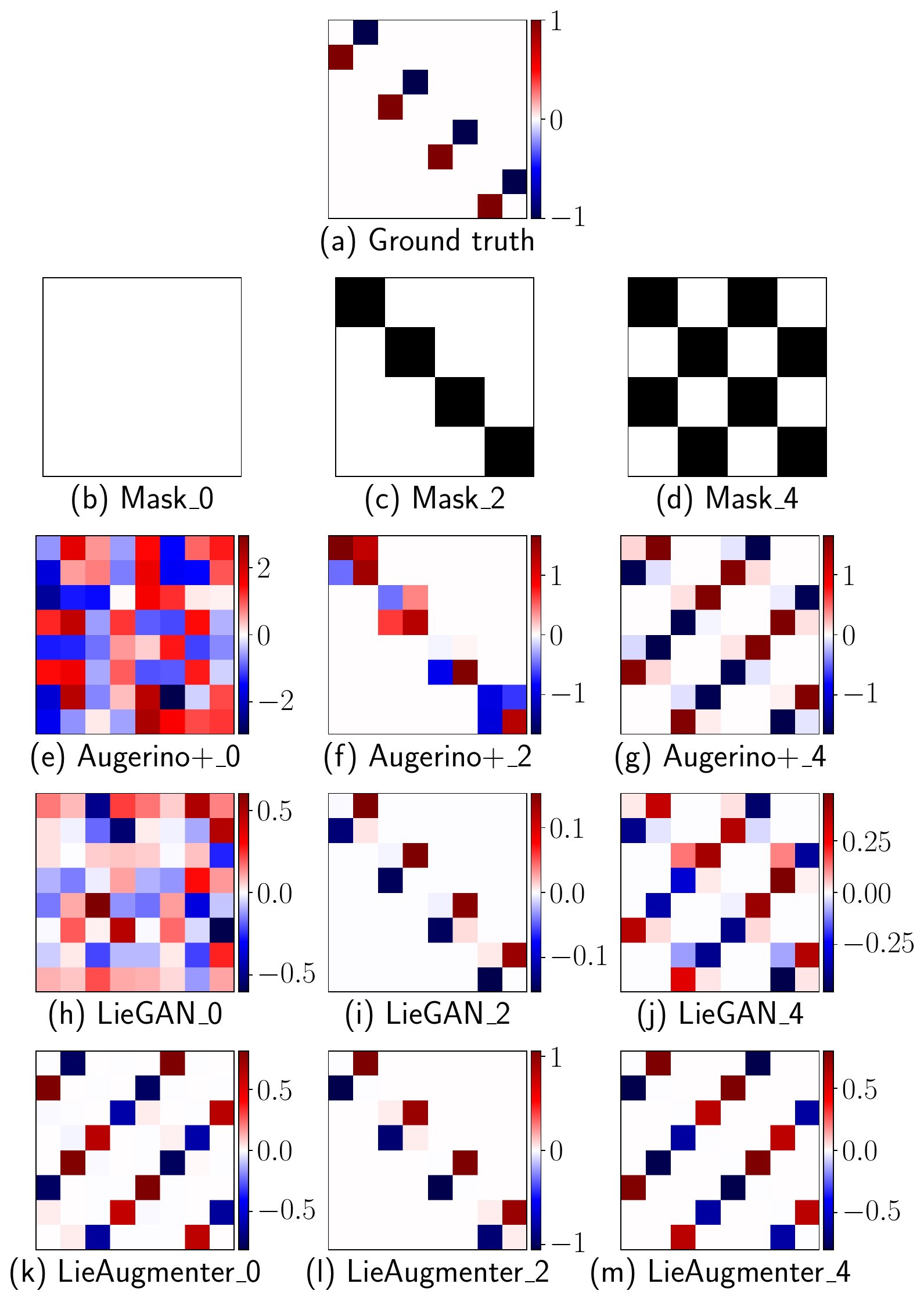}
  \includegraphics[width=0.45\textwidth]{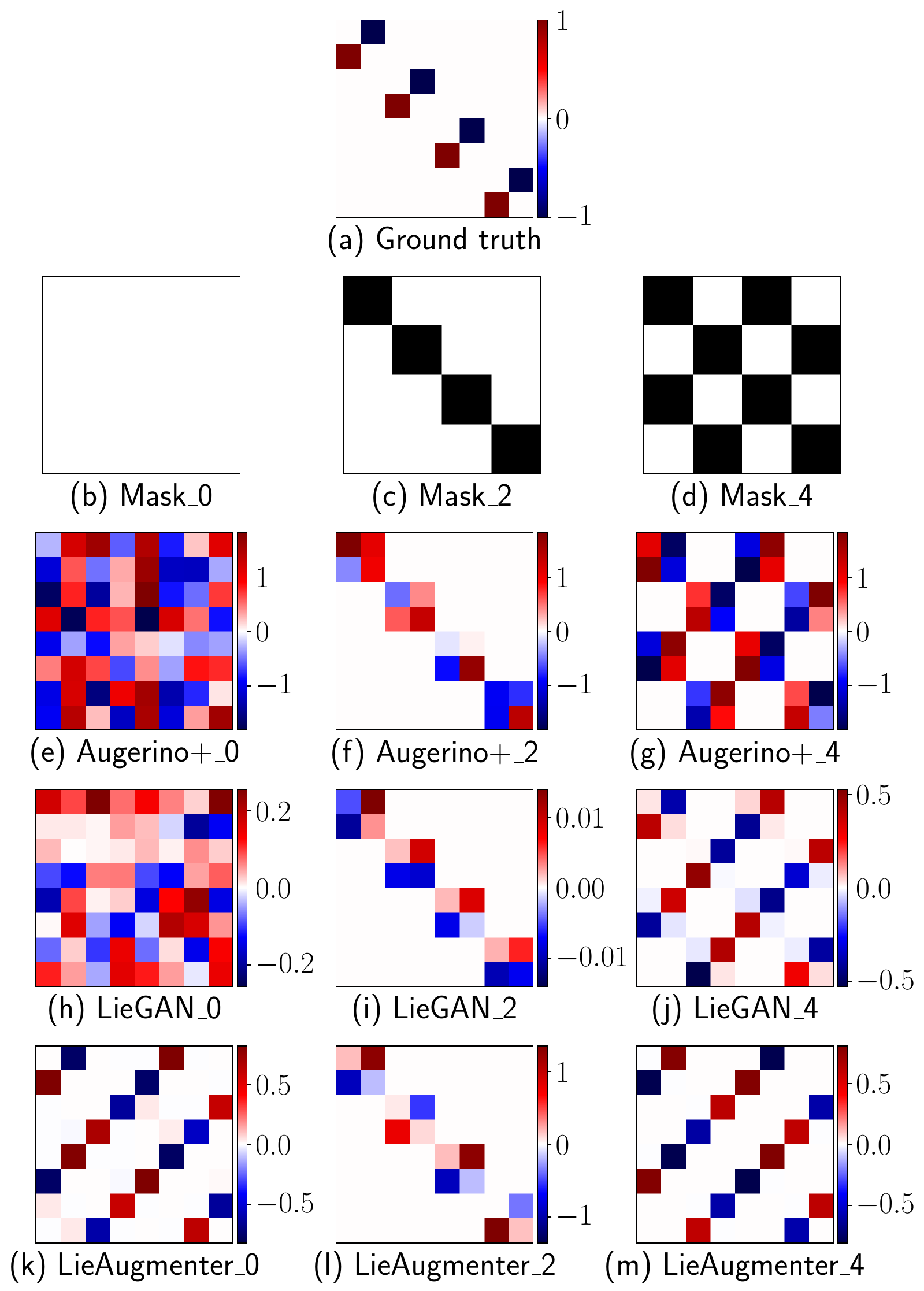}
  \caption{
   Ground-truth Lie algebra basis, masks of learnable entries, and bases learned by baseline symmetry discovery methods and LieAugmenter for 2-body dynamics prediction with three input and three output timesteps. Results are shown for the ID (\textit{left}) and OOD settings (\textit{right}).
  }
  \label{fig:lie_algebras_nbody2_all_t3}
\end{figure}

\paragraph{{Symmetry Discovery.}} Figure~\ref{fig:lie_algebras_nbody2_all_t3} shows a visualization of the ground truth Lie algebra basis and masks, together with the bases learned by LieAugmenter and each of the baseline symmetry discovery approaches.
As we can observe, once again, LieAugmenter provides a strong and robust performance across both the ID and OOD settings irrespective of the applied mask, learning either the canonical representation of the basis or the alternative one that we saw previously.

\paragraph{{Predictive Performance.}} 
Table~\ref{tab:results_nbody2_masks_t3} reports the MSE and equivariance error of the different approaches in the ID and OOD settings. 
Regardless of the mask, LieAugmenter again achieves the best performance only (closely) behind the equivariant model, but still surpassing the methods that consider oracle augmentation and augmentation based on discovered symmetries. 
As previously noted, we remark the faster training speed of LieAugmenter over the equivariant model (requiring almost a third of the training time), as we present in Table~\ref{tab:times_nbody2}. 

\begin{table}[h]
  \caption{$2$-body dynamics results for ID and OOD evaluation under different masks restrictions on the symmetry search space for three input and output timesteps. We report MSE and equivariance error for an equivariant architecture (EMLP) and for MLP models trained (i) without augmentation (\emph{Trivial}), (ii) with \emph{Oracle} rotation augmentation (ground-truth symmetry), (iii) with symmetry-discovery baselines (Augerino+ and LieGAN Aug.), and (iv) with LieAugmenter (\emph{Ours}). All values are reported as $\mathrm{mean}\pm\mathrm{std}$ over three random seeds.}
  \label{tab:results_nbody2_masks_t3}
  \centering
\resizebox{\textwidth}{!}{%
    \begin{tabular}{clcccc}
\toprule
& & \multicolumn{2}{c}{In-distribution} & \multicolumn{2}{c}{Out-of-distribution} \\
\cmidrule{3-6}
& & MSE $(\downarrow)$ & Equiv. Error $(\downarrow)$ & MSE $(\downarrow)$ & Equiv. Error $(\downarrow)$ \\
\midrule
\multirow{1}{*}{\makecell{Equiv. architecture}} & EMLP & $\text{4.54e-06} \scriptstyle{{\pm \text{2.20e-06}}}$ & $\text{2.91e-05} \scriptstyle{{\pm \text{4.71e-06}}}$ & $\text{6.30e-06} \scriptstyle{{\pm \text{3.68e-06}}}$ & $\text{2.90e-05} \scriptstyle{{\pm \text{4.79e-06}}}$ \\
\midrule
\multirow{2}{*}[-0.15em]{\makecell{Data\\augmentation}} & Trivial & $\text{9.62e-03} \scriptstyle{{\pm \text{4.32e-03}}}$ & $\text{3.58e-02} \scriptstyle{{\pm \text{7.57e-03}}}$ & $\text{1.45e-02} \scriptstyle{{\pm \text{1.30e-03}}}$ & $\text{3.51e-02} \scriptstyle{{\pm \text{1.39e-03}}}$ \\
& Oracle & $\text{3.96e-05} \scriptstyle{{\pm \text{1.75e-05}}}$ & $\text{5.08e-03} \scriptstyle{{\pm \text{1.19e-03}}}$ & $\text{4.92e-05} \scriptstyle{{\pm \text{2.25e-05}}}$ & $\text{3.88e-03} \scriptstyle{{\pm \text{6.66e-04}}}$ \\
\midrule
\multirow{9}{*}[-0.15em]{\makecell{Symmetry\\discovery}} & Augerino+\_0 & $\text{4.03e+00} \scriptstyle{{\pm \text{3.49e-01}}}$ & $\text{1.91e-01} \scriptstyle{{\pm \text{2.17e-03}}}$ & $\text{2.68e+00} \scriptstyle{{\pm \text{4.94e-01}}}$ & $\text{1.98e-01} \scriptstyle{{\pm \text{9.13e-03}}}$ \\
& LieGAN\_0 aug & $\text{8.83e-03} \scriptstyle{{\pm \text{2.72e-03}}}$ & $\text{3.59e-02} \scriptstyle{{\pm \text{5.83e-03}}}$ & $\text{1.88e-02} \scriptstyle{{\pm \text{4.80e-03}}}$ & $\text{4.15e-02} \scriptstyle{{\pm \text{5.98e-03}}}$ \\
& LieAugmenter\_0 & $\text{5.98e-05} \scriptstyle{{\pm \text{2.17e-05}}}$ & $\text{4.07e-03} \scriptstyle{{\pm \text{1.50e-03}}}$ & $\text{1.59e-04} \scriptstyle{{\pm \text{1.93e-04}}}$ & $\text{4.90e-03} \scriptstyle{{\pm \text{1.88e-03}}}$ \\
\cmidrule{2-6}
& Augerino+\_2 & $\text{8.70e+00} \scriptstyle{{\pm \text{1.00e+01}}}$ & $\text{1.96e-01} \scriptstyle{{\pm \text{8.81e-03}}}$ & $\text{1.09e+01} \scriptstyle{{\pm \text{1.33e+01}}}$ & $\text{1.87e-01} \scriptstyle{{\pm \text{8.77e-03}}}$ \\
& LieGAN\_2 aug & $\text{2.71e-03} \scriptstyle{{\pm \text{2.39e-04}}}$ & $\text{2.26e-02} \scriptstyle{{\pm \text{1.16e-03}}}$ & $\text{1.83e-02} \scriptstyle{{\pm \text{4.37e-03}}}$ & $\text{4.05e-02} \scriptstyle{{\pm \text{3.98e-03}}}$ \\
& LieAugmenter\_2 & $\text{1.75e-05} \scriptstyle{{\pm \text{6.22e-06}}}$ & $\text{4.58e-03} \scriptstyle{{\pm \text{4.07e-04}}}$ & $\text{3.67e-04} \scriptstyle{{\pm \text{1.10e-04}}}$ & $\text{7.97e-03} \scriptstyle{{\pm \text{1.45e-03}}}$ \\
\cmidrule{2-6}
& Augerino+\_4 & $\text{1.72e+00} \scriptstyle{{\pm \text{1.03e-01}}}$ & $\text{1.88e-01} \scriptstyle{{\pm \text{1.83e-03}}}$ & $\text{1.64e+00} \scriptstyle{{\pm \text{7.75e-02}}}$ & $\text{1.88e-01} \scriptstyle{{\pm \text{2.73e-03}}}$ \\
& LieGAN\_4 aug & $\text{8.86e-05} \scriptstyle{{\pm \text{7.58e-05}}}$ & $\text{4.24e-03} \scriptstyle{{\pm \text{1.10e-03}}}$ & $\text{7.94e-05} \scriptstyle{{\pm \text{2.55e-05}}}$ & $\text{4.78e-03} \scriptstyle{{\pm \text{7.64e-04}}}$ \\
& LieAugmenter\_ 4 & $\text{2.85e-05} \scriptstyle{{\pm \text{3.92e-06}}}$ & $\text{3.97e-03} \scriptstyle{{\pm \text{6.82e-04}}}$ & $\text{5.32e-05} \scriptstyle{{\pm \text{1.28e-05}}}$ & $\text{4.22e-03} \scriptstyle{{\pm \text{1.75e-04}}}$ \\
\bottomrule
\end{tabular}
}
\end{table}

\clearpage

\subsubsection{Alternative representation of the rotation symmetry}
\label{sec:alternative_rep_nbody2}

Throughout our experiments with this $2$-body dynamics dataset, we have observed what we have claimed to be an alternative representation of the correct symmetry in the context of the specific values for the physical parameters of this dataset. In that sense, following \citet{yangGenerativeAdversarialSymmetry2023}, we present the derivation that confirms that claim.

Firstly, we can write this alternative representation of the Lie algebra basis in the following way to highlight its structure after removing the slightly noisy entries:
\begin{align*}
    R &= \begin{bmatrix}
        0 & -1 \\
        1 & 0
    \end{bmatrix} \\
    L &= \begin{bmatrix}
        R & 0 & -R & 0\\
        0 & R & 0 & -R \\
        -R & 0 & R & 0 \\
        0 & -R & 0 & R \\
    \end{bmatrix}
\end{align*}

Then, we can express the result of applying the matrix exponential to a weighted version of that basis as follows:

\begin{align*}
    \exp (w L) &= I + \begin{bmatrix}
        G(w) & 0 & -G(w) & 0 \\
        0 & G(w) & 0 & -G(w) \\
        -G(w) & 0 & G(w) & 0 \\
        0 & -G(w) & 0 & G(w) \\
    \end{bmatrix} \\
    G(w) &= \sum\limits_{n=0}^\infty \dfrac{(-1)^n 2^{2n} w^{2n + 1}}{(2n + 1)!} R + \sum\limits_{n=1}^\infty \dfrac{(-1)^n 2^{2n-1} w^{2n}}{(2n)!} I 
\end{align*}

As a result, we can define its action on the input data as follows:

\begin{equation*}
    \exp(w L) \begin{bmatrix}
        \mathbf{q}_1 \\ \mathbf{p}_1 \\ \mathbf{q}_2 \\ \mathbf{p}_2
        \end{bmatrix}
    = \text{diag} (I + 2 G(w)) \begin{bmatrix}
        \mathbf{q}_1 \\ \mathbf{p}_1 \\ \mathbf{q}_2 \\ \mathbf{p}_2
        \end{bmatrix},
\end{equation*}
where we have used our knowledge about the dataset, i.e., the origin is at the center of mass, and we have $m_1 = m_2$, $\mathbf{q}_1 = -\mathbf{q}_2$, and $\mathbf{p_1} = - \mathbf{p}_2$.

Lastly, we can notice the following relation between the resulting expression and the ground truth rotation symmetry:

\begin{align*}
    I + 2 G(w) &= \sum\limits_{n=0}^\infty \dfrac{(-1)^n 2^{2n+1} w^{2n + 1}}{(2n + 1)!} R + \sum\limits_{n=0}^\infty \dfrac{(-1)^n 2^{2n} w^{2n}}{(2n)!} I  \\
    &= \begin{bmatrix}
        \cos 2w & -\sin 2w \\
        \sin 2w & \cos 2w
    \end{bmatrix}
\end{align*}

\subsection{Behavior in the Absence of Symmetry}
\label{app:nosymmetry}

In this section, we complement our analysis of the behavior of LieAugmenter and symmetry discovery baselines in the absence of symmetries in the data.

First, we compute two metrics that quantify how close to the identity the learned symmetry is for each of the considered methods and present the results in Table~\ref{tab:metrics_no_symmetry}. In this way, we seek to more closely analyze whether the previously visualized Lie algebra bases actually encode any behavior similar to the expected identity transformations.
From the results, we can observe that Augerino+ learns augmentations that significantly differ from the identity, and that LieGAN learns augmentations that are closest to the identity but still considerably differ from it. For LieAugmenter, the augmentations are similarly close to the identity as for LieGAN, and the high variability of the results matches our previous analysis of the visualizations of the learned Lie algebra basis.

Secondly, we present the performance of LieAugmenter and different augmentation approaches in Table~\ref{tab:performance_no_symmetry}, aiming to verify that our method does not lead to a degradation of the performance even when there is no continuous symmetry to discover in the data.
As we can observe, the performance of LieAugmenter is indeed similar to that provided by the other augmentation approaches, and even slightly better. This indicates that our proposed method can still be successfully applied despite our goal of discovering underlying symmetries not being supported by the data.

\begin{minipage}{0.5\textwidth}
    \centering
    \captionof{table}{Evaluation of the proximity of the learned augmentations to the identity in a dataset without continuous symmetries for LieAugmenter and baseline symmetry discovery approaches.}
    \resizebox{\linewidth}{!}{%
    \begin{tabular}{lcc}
\toprule
& $\mathbb{E}_{x, g} \left[ \lvert\lvert \rho_{\mathcal{X}}(g) x - x\rvert\rvert_2\right]$ & $\mathbb{E} \left[\lvert\lvert g - I \rvert\rvert_F\right]$ \\
\midrule
Augerino+ & $262.8388 \scriptstyle{\pm 199.6667}$ & $176.3353 \scriptstyle{\pm 192.0428}$ \\
LieGAN & $6.9031 \scriptstyle{\pm 0.0805}$ & $3.4250 \scriptstyle{\pm 0.1813}$ \\
LieAugmenter & $10.6570 \scriptstyle{\pm 4.4892}$ & $5.5936 \scriptstyle{\pm 0.0123}$ \\
\bottomrule
\end{tabular}
    \label{tab:metrics_no_symmetry}
    }
\end{minipage}
\hfill
\begin{minipage}{0.48\textwidth}
    \centering
    \captionof{table}{Evaluation of the predictive performance in a dataset without continuous symmetries. We report MSE for MLP models trained (i) without augmentation (\emph{Trivial}), (ii) with \emph{Oracle} augmentation (identity transformations), (iii) with symmetry-discovery baselines (Augerino+ and LieGAN Aug.), and (iv) with LieAugmenter. All values are reported as $\mathrm{mean}\pm\mathrm{std}$ over three random seeds.}
    \label{tab:performance_no_symmetry}
    \resizebox{0.9\linewidth}{!}{%
\begin{tabular}{llc}
\toprule
& & MSE $(\downarrow)$ \\ 
\midrule
\multirow{2}{*}[-0.15em]{Data augmentation} & No aug. & $\text{2.29e-04} \scriptstyle{{\pm \text{1.86e-04}}}$ \\ 
& Oracle aug. & $\text{6.31e-04} \scriptstyle{{\pm \text{4.24e-04}}}$ \\ 
\midrule
\multirow{3}{*}[-0.15em]{\makecell{Symmetry discovery}} & Augerino+ & $\underline{\text{2.22e-04}} \scriptstyle{{\pm \text{1.12e-04}}}$ \\ 
& LieGAN Aug. & $\text{6.30e-04} \scriptstyle{{\pm \text{4.31e-04}}}$ \\ 
& LieAugmenter & $\textbf{\text{1.99e-04}} \scriptstyle{{\pm \text{7.23e-05}}}$ \\ 
\bottomrule
\end{tabular}
}
\end{minipage}

\subsection{Rotations in Colorectal Cancer Detection from Images}

As another more complex experiment for analyzing the performance of our proposal, we consider the task of human colorectal cancer (CRC) detection from stained histological images of human tissues. In particular, we consider the datasets NCT-CRC-HE-100K for training and CRC-VAL-HE-7K for testing \citep{kather_2018_1214456}, each containing 100,000 and 7,000 samples, respectively, of histological slices of human colorectal cancer and normal tissue, represented as 224x224 RGB images. Rotational invariance has been shown to be particularly important for this classification task \citep{gerkenEmergentEquivarianceDeep2024}.

The main goal of this experiment is to explore the applicability of our approach to a more complex and relevant computer vision scenario, where the images are larger and in color. In addition, we also seek to study the effect of applying LieAugmenter to more complex architectures besides simple CNNs and MLPs, e.g., Vision Transformer (ViT) \citep{dosovitskiy2020image} and EfficientNet \citep{tan2019efficientnet}.

Note that in this experiment, we do not seek to attain the highest possible performance for the considered models in this dataset, but rather to observe the effect of applying LieAugmenter in contrast to using trivial and ground truth augmentations on each of the base models. As a result, we do not tune the hyperparameters, which we set similarly to previous experiments: $\alpha=1.0$, $\beta=7.0$, $\lambda=0.1$, $\eta=0.0$, $\nu=0.01$, $\gamma=3.0$, and $K=10$. 
We train all models for $25$ epochs with a batch size of $64$.

\begin{figure}[h!]
  \centering
  \includegraphics[width=0.6\textwidth]{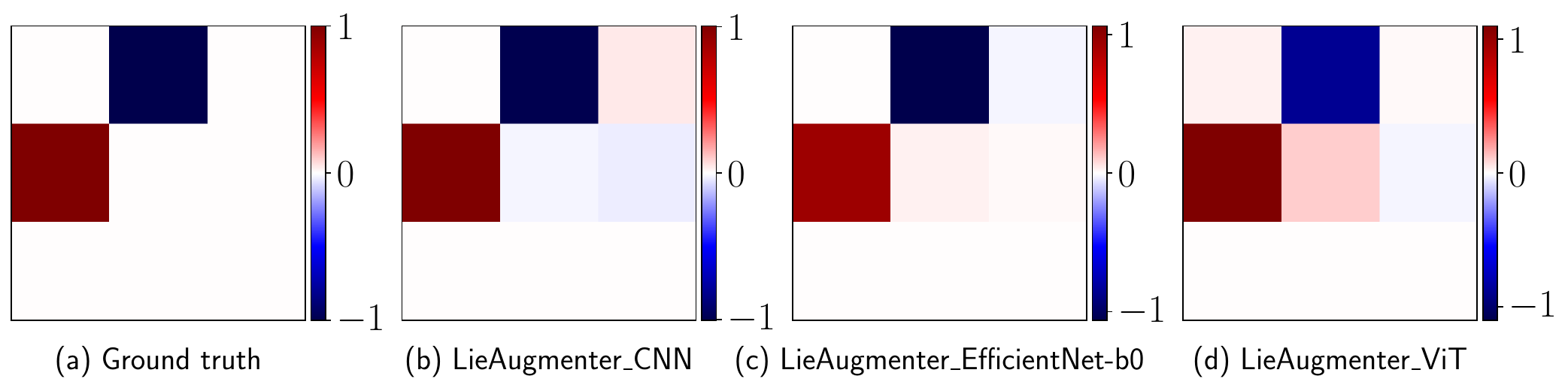}
  \caption{
   Ground-truth Lie algebra basis, and bases learned by LieAugmenter trained with CNN, EfficientNet-b0, and ViT as prediction networks for the CRC dataset.
  }
  \label{fig:crc_lie_algebras}
\end{figure}

\paragraph{{Symmetry Discovery.}} 
In terms of symmetry discovery, we follow the same procedure as for the RotatedMNIST dataset and set the cardinality of the Lie algebra basis to $C=1$ while searching over the six-dimensional space of 2D affine transformations, which are defined to act on the pixel coordinates of the images. 
With that definition, LieAugmenter is able to learn the correct symmetry for all of the considered architectures, as shown in Figure~\ref{fig:crc_lie_algebras}.

\paragraph{{Predictive Performance.}} 
As we can see in the results in Table \ref{tab:results_crc}, LieAugmenter provides an equal or better performance than using ground truth augmentations for all three prediction networks. This observation coincides with and reinforces our previous results, demonstrating once more the benefits and ease of applicability of our approach.

\begin{table}[h!]
  \caption{CRC results for the evaluation of the CNN, EfficientNet-b0, and Vision Transformer models trained (i) without augmentation (No aug.), (ii) with ground truth augmentations (Oracle aug.), and (iii) LieAugmenter. We report accuracy and equivariance error as $\mathrm{mean}\pm\mathrm{std}$ over three random seeds.}
  \label{tab:results_crc}
  \centering
\begin{tabular}{llcc}
Prediction network & Method & Accuracy (\%) ($\uparrow$) & Equiv. Error ($\downarrow$) \\
\toprule
\multirow{3}{*}[-0.15em]{CNN} & No aug. & $84.87 \scriptstyle{\pm 4.44}$ & $2.95 \scriptstyle{\pm 0.15}$ \\
& Oracle aug. & $\underline{85.27} \scriptstyle{\pm 3.34}$ & $\mathbf{1.75} \scriptstyle{\pm 0.29}$ \\
& LieAugmenter & $\mathbf{89.42} \scriptstyle{\pm 0.89}$ & $\underline{1.86} \scriptstyle{\pm 0.21}$ \\
\midrule
\multirow{3}{*}[-0.15em]{EfficientNet-b0} & No aug. & $\underline{93.04} \scriptstyle{\pm 1.18}$ & $2.55 \scriptstyle{\pm 0.24}$ \\
& Oracle aug & $\mathbf{93.35} \scriptstyle{\pm 1.93}$ & $\underline{1.49} \scriptstyle{\pm 0.14}$ \\
& LieAugmenter & $92.74 \scriptstyle{\pm 2.55}$ & $\mathbf{1.21} \scriptstyle{\pm 0.10}$ \\
\midrule
\multirow{3}{*}[-0.15em]{Vision Transformer} & No aug & $75.69 \scriptstyle{\pm 3.90}$ & $2.02 \scriptstyle{\pm 0.42}$ \\
& Oracle aug & $\underline{81.16} \scriptstyle{\pm 2.67}$ & $\underline{1.09} \scriptstyle{\pm 0.07}$ \\
& LieAugmenter & $\mathbf{85.63} \scriptstyle{\pm 2.93}$ & $\mathbf{0.82} \scriptstyle{\pm 0.15}$ \\
\bottomrule
\end{tabular}
\end{table}

\subsection{Discovering Discrete Symmetries}
\label{app:discrete_symmetries}

Next, we also explore the ability of LieAugmenter to identify the correct Lie group when the symmetry of the data corresponds to discrete subgroups of a Lie group instead of continuous groups, as we studied in the main experiments. To that effect, we construct two synthetic regression datasets based on some scenarios explored in \citet{yangGenerativeAdversarialSymmetry2023}.

For both of these datasets, we set the following hyperparameter values for our model: $\alpha=1.0$, $\beta=5.0$, $\lambda=0.1$, $\eta=0.0$, $\nu=0.001$, $\gamma=3.0$, and $K=10$. Once more, $\eta$ has no effect on this particular scenario, because the cardinality of the bases is $C=1$. 
We generate $54,000$ samples for the training set, $6,000$ for the validation set, and $10,000$ for the test set.
The prediction network for LieAugmenter and the different augmentation approaches is defined in both problems as a simple MLP, consisting of four layers and ReLU activations. 

In order to attempt to model the discrete nature of the studied groups, we modify the sampling strategy of the coefficients of the linear combination of the elements of the Lie algebra basis by sampling from a uniform integer grid instead of a uniform distribution. We would expect such modification to possibly lead to slight predictive performance improvements, but it could also degrade the quality of the symmetry discovery results due to the restricted nature of the generated augmentations. 

Since prior knowledge about the discrete or continuous nature of the underlying group is not always readily available, we also explore the performance of LieAugmenter with the previous uniform sampling strategy to show its robustness to such a possible misspecification. We distinguish the results by the nomenclature `LieAugmenter (discrete)' and `LieAugmenter (continuous)'. Overall, we observe that both sampling strategies provide similarly strong performance (especially for symmetry discovery), with the discrete formulation providing slightly higher task performance in our experiments.

\subsubsection{Discrete Rotations}

As the first synthetic dataset, we consider the synthetic regression problem given by $f (x, y, z) = z/(1 + \arctan \frac{y}{x} \mod \frac{2\pi}{k})$, which is invariant to rotations of multiples of $2\pi/k$ in the $xy$ plane, thus corresponding to a discrete cyclic subgroup of $\mathrm{SO}(2)$ with size $k$. Specifically, we consider the case where $k=6$ and $x,y,z \sim \mathcal{N}(0, I)$.

\begin{figure}[h]
  \centering
  \includegraphics[width=\textwidth]{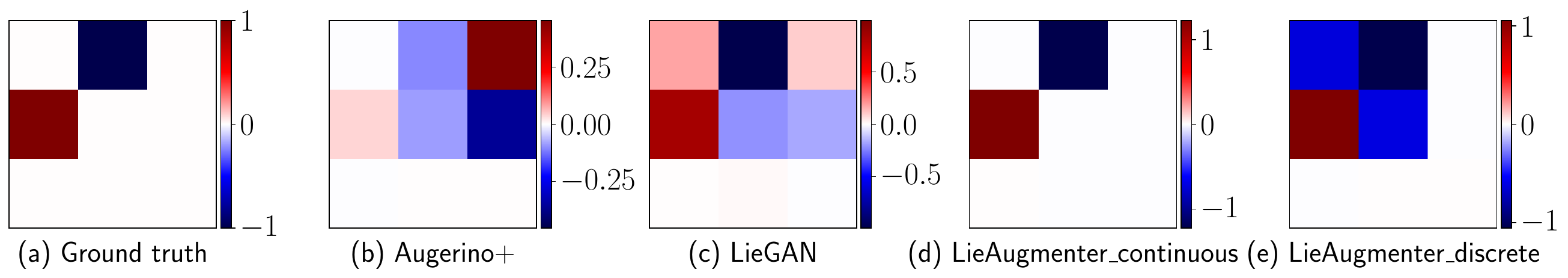}
  \caption{Ground-truth Lie algebra basis, and bases learned by the different sampling strategies for LieAugmenter and baseline symmetry discovery methods in the discrete rotation dataset. The sign of the bases is aligned to simplify their interpretation, since it is irrelevant for their correctness.}
  \label{fig:lie_algebras_discrete_rotation}
\end{figure}

\begin{minipage}{0.5\textwidth}
\paragraph{{Symmetry Discovery.}} 
From Figures~\ref{fig:lie_algebras_discrete_rotation} and Figure~\ref{fig:lie_comparison_discreterot}, we can see that LieAugmenter is capable of correctly learning a Lie algebra basis for the $\mathrm{SO}(2)$ supergroup of the underlying discrete rotation symmetry irrespectively of the discreteness of the distribution selected for sampling.
It seems, however, that discretely sampling from a uniform integer grid leads to a slightly worse recovery of the basis, as we had expected.
\end{minipage}
\hfill
\begin{minipage}{0.45\textwidth}
  \centering
  \includegraphics[width=\linewidth]{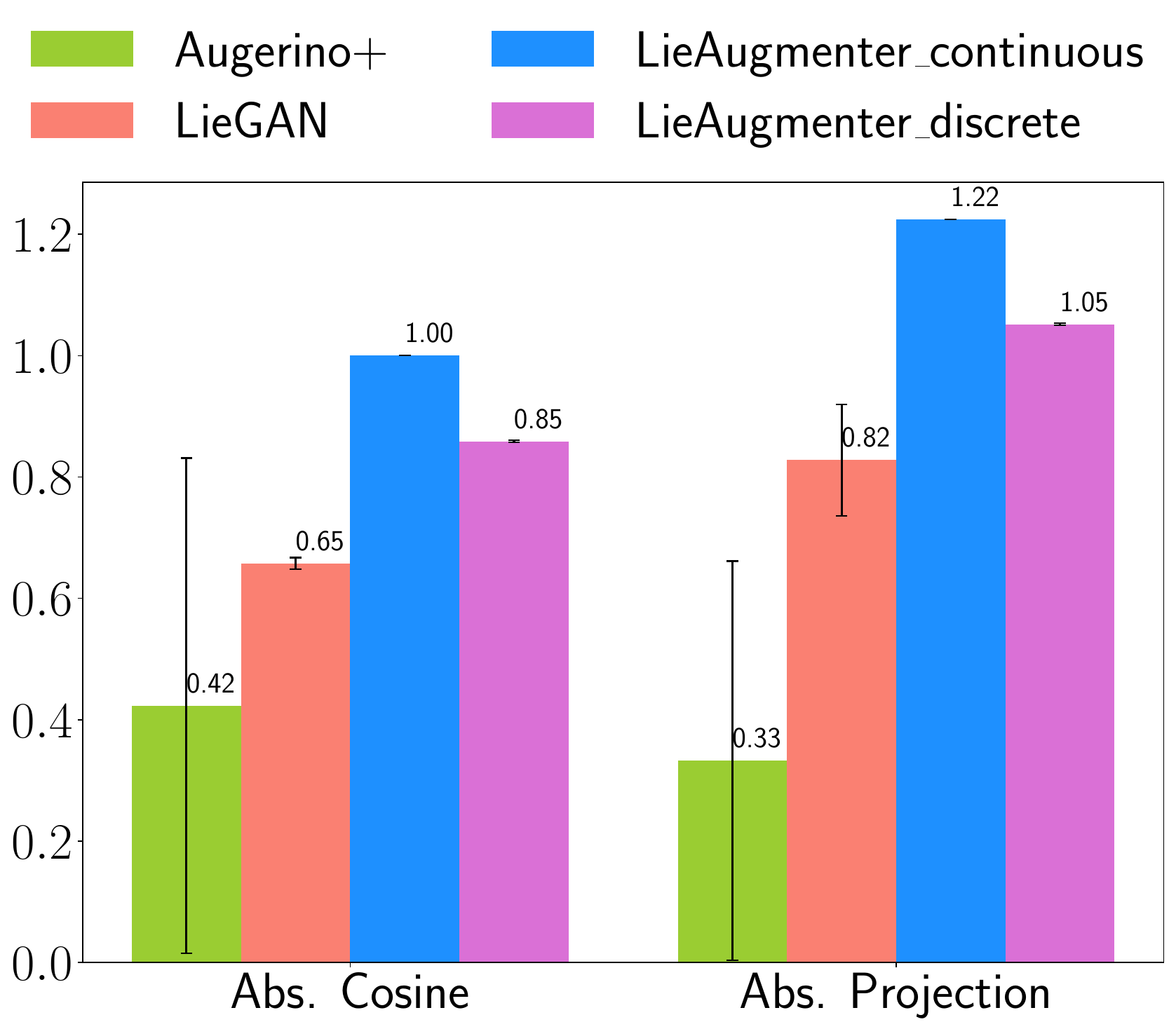}
  \captionof{figure}{Absolute cosine similarity and absolute Frobenius projection for Lie algebra basis recovery on the discrete rotation dataset ($\mathrm{mean}\pm\mathrm{std}$ over three random seeds).}
  \label{fig:lie_comparison_discreterot}
\end{minipage}

\paragraph{{Predictive Performance.}} 
Table~\ref{tab:results_discrete_rotation} reports the MSE and equivariance error of the different prediction approaches.
We can observe that the discrete sampling version of LieAugmenter provides the strongest performance, with an error that is notably much lower than the one from applying our standard continuous sampling strategy.
This result thus indicates that despite more accurately recovering the correct subgroup of the underlying symmetry, enforcing invariance in a manner that does not exactly match the nature of such symmetry can be detrimental to the performance.
As a result, this motivates a future exploration of the possibility of making the sampling distribution learnable to dynamically adapt to the characteristics of the considered data.

\begin{table}[h]
  \caption{Discrete rotation evaluation results. We report MSE and equivariance error for MLP models trained (i) without augmentation, (ii) with \emph{Oracle} rotation augmentation (ground-truth symmetry), (iii) with symmetry-discovery baselines (Augerino+ and LieGAN augmentation), and (iv) with the discrete and continuous sampling strategies for LieAugmenter (\emph{Ours}). All values are reported as $\mathrm{mean}\pm\mathrm{std}$ over three random seeds.}
  \label{tab:results_discrete_rotation}
  \centering
  \resizebox{0.8\linewidth}{!}{%
\begin{tabular}{llcc}
\toprule
& & MSE $(\downarrow)$ & Equiv. Error $(\downarrow)$ \\
\toprule
\multirow{2}{*}[-0.15em]{Data augmentation} & No augmentation & $\underline{\text{4.06e-03}} \scriptstyle{{\pm \text{3.02e-04}}}$ & $\text{9.57e-02} \scriptstyle{{\pm \text{1.54e-03}}}$ \\
& Oracle augmentation & $\text{2.10e-02} \scriptstyle{{\pm \text{1.19e-03}}}$ & $\textbf{\text{2.31e-03}} \scriptstyle{{\pm \text{2.36e-04}}}$ \\
\midrule
\multirow{4}{*}[-0.15em]{\makecell{Symmetry discovery}} & Augerino+ & $\text{2.09e-02} \scriptstyle{{\pm \text{5.56e-04}}}$ & $\text{1.08e-02} \scriptstyle{{\pm \text{4.22e-03}}}$ \\
& LieGAN augmentation & $\text{2.10e-02} \scriptstyle{{\pm \text{1.10e-03}}}$ & $\text{3.28e-03} \scriptstyle{{\pm \text{3.45e-04}}}$ \\
& LieAugmenter (continuous) & $\text{2.12e-02} \scriptstyle{{\pm \text{1.24e-03}}}$ & $\underline{\text{2.46e-03}} \scriptstyle{{\pm \text{3.29e-04}}}$ \\
& LieAugmenter (discrete) & $\textbf{\text{3.36e-03}} \scriptstyle{{\pm \text{4.56e-04}}}$ & $\text{3.09e-02} \scriptstyle{{\pm \text{1.80e-03}}}$ \\
\bottomrule
\end{tabular}
}
\end{table}

\subsubsection{Partial Permutations}

Next, we seek to show that LieAugmenter is able to learn discrete symmetry groups beyond those concerning rotations. In particular, we consider the function $f (x) = x_1 + x_2 + x_3 +x_4^2 - x_5^2, \ x \in \mathbb{R}^5$ with $x \sim \mathcal{U}[-100, 100]$. This function is invariant to partial permutations, with the output remaining the same if we permute the first three dimensions of the input, but changing if we permute the last two. 

\begin{figure}[h!]
  \centering
  \includegraphics[width=\textwidth]{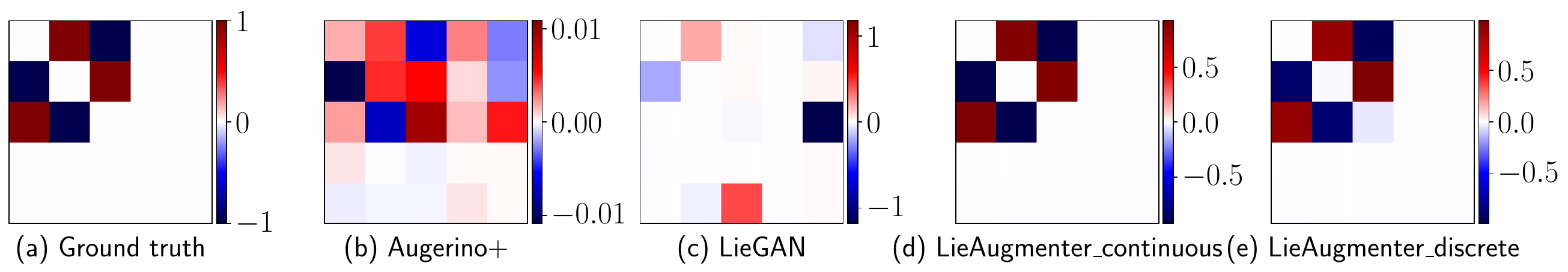}
  \caption{Ground-truth Lie algebra basis, and bases learned by the different sampling strategies for LieAugmenter and baseline symmetry discovery methods in the partial permutation dataset. The sign of the bases is aligned to simplify their interpretation, since it is irrelevant for their correctness.}
  \label{fig:lie_algebras_partial_permutation}
\end{figure}

\begin{minipage}{0.5\textwidth}
\paragraph{{Symmetry Discovery.}} 
From Figures~\ref{fig:lie_algebras_partial_permutation} and Figure~\ref{fig:lie_comparison_partialperm}, we can see that LieAugmenter is capable of correctly learning a Lie algebra basis for the underlying discrete partial permutation symmetry irrespectively of the discreteness of the distribution selected for sampling.\\

Although LieGAN underperforms in our runs, the original work introducing the method \citep{yangGenerativeAdversarialSymmetry2023} reports successful recovery of this symmetry. Our implementation follows the available description and code as closely as possible, but we did not match the reported performance for this particular setting.

\end{minipage}
\hfill
\begin{minipage}{0.45\textwidth}
  \centering
  \includegraphics[width=\linewidth]{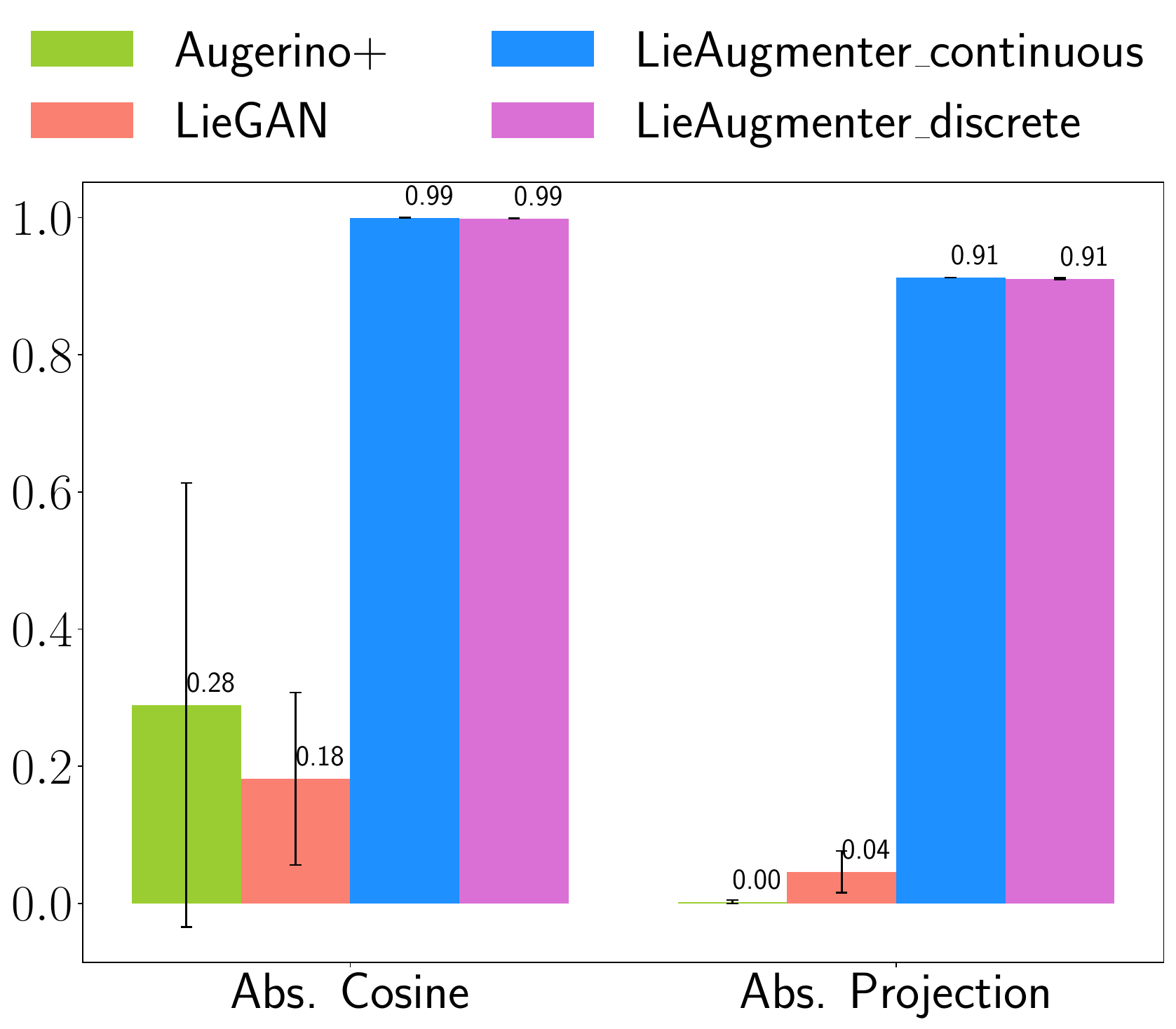}
  \captionof{figure}{Absolute cosine similarity and absolute Frobenius projection for Lie algebra basis recovery on the partial permutation dataset ($\mathrm{mean}\pm\mathrm{std}$ over three random seeds).}
  \label{fig:lie_comparison_partialperm}
\end{minipage}

\paragraph{{Predictive Performance.}} 
From the results in Table~\ref{tab:results_partial_permutation}, we can observe that once more the discrete sampling strategy for LieAugmenter provides the strongest results, outperforming the continuous approach to sampling.
However, differently from the previous discrete rotation experiment, in this case the performance difference is much smaller, which suggests that our standard continuous sampling setting can still perform quite robustly not only in terms of symmetry discovery but also for the generation of predictions despite the misalignment with the level of discreteness of the symmetry that underlies the data.

\begin{table}[h!]
  \caption{Partial permutation evaluation results. We report MSE and equivariance error for MLP models trained (i) without augmentation, (ii) with \emph{Oracle} rotation augmentation (ground-truth symmetry), (iii) with symmetry-discovery baselines (Augerino+ and LieGAN augmentation), and (iv) with the discrete and continuous sampling strategies for LieAugmenter (\emph{Ours}). All values are reported as $\mathrm{mean}\pm\mathrm{std}$ over three random seeds.}
  \label{tab:results_partial_permutation}
  \centering
  \resizebox{0.8\linewidth}{!}{%
\begin{tabular}{llcc}
\toprule
& & MSE $(\downarrow)$ & Equiv. Error $(\downarrow)$ \\
\toprule
\multirow{2}{*}[-0.15em]{Data augmentation}  & No augmentation & $\text{1.97e+03} \scriptstyle{{\pm \text{1.20e+03}}}$ & $\text{1.23e+01} \scriptstyle{{\pm \text{3.21e+00}}}$ \\
& Oracle augmentation & $\underline{\text{1.69e+03}} \scriptstyle{{\pm \text{8.44e+02}}}$ & $\text{1.41e+01} \scriptstyle{{\pm \text{2.06e+00}}}$ \\
\midrule
\multirow{4}{*}[-0.15em]{\makecell{Symmetry discovery}} & Augerino+ & $\text{3.38e+03} \scriptstyle{{\pm \text{8.96e+02}}}$ & $\text{2.52e+01} \scriptstyle{{\pm \text{4.52e+00}}}$ \\
&LieGAN augmentation & $\text{1.24e+04} \scriptstyle{{\pm \text{2.90e+03}}}$ & $\textbf{\text{1.05e+01}} \scriptstyle{{\pm \text{2.12e+00}}}$ \\
&LieAugmenter (continuous) & $\text{1.77e+03} \scriptstyle{{\pm \text{3.57e+02}}}$ & $\text{1.37e+01} \scriptstyle{{\pm \text{4.55e+00}}}$ \\
&LieAugmenter (discrete) & $\textbf{\text{1.49e+03}} \scriptstyle{{\pm \text{6.55e+01}}}$ & $\underline{\text{1.19e+01}} \scriptstyle{{\pm \text{2.76e+00}}}$ \\
\bottomrule
\end{tabular}
}
\end{table}

\subsection{Lorentz Symmetries in Quantum Field Theory Amplitude Predictions}

As an additional experiment, we study the applicability of LieAugmenter to the task of defining a neural surrogate for quantum field theoretical amplitudes \citep{spinner2024lorentz, aylett2021optimising, badger2020using}. This is a problem of great importance in the physics community due to it being central to the theoretical predictions that LHC measurements are compared to. In particular, these amplitudes describe the unnormalized probability of interactions of fundamental particles as a function of their four-momenta, which are exactly Lorentz-invariant.
Following \citet{spinner2024lorentz}, we study the setting $q \bar{q} \rightarrow Z + n g$, i.e., the production of a Z boson with $n = 1, \ldots, 4$ additional gluons from a quark-antiquark pair. In particular, we focus on the scenario where $n = 2$ and we generate and split the dataset as in that work.

\begin{figure}[h]
  \centering
  \includegraphics[width=0.9\textwidth]{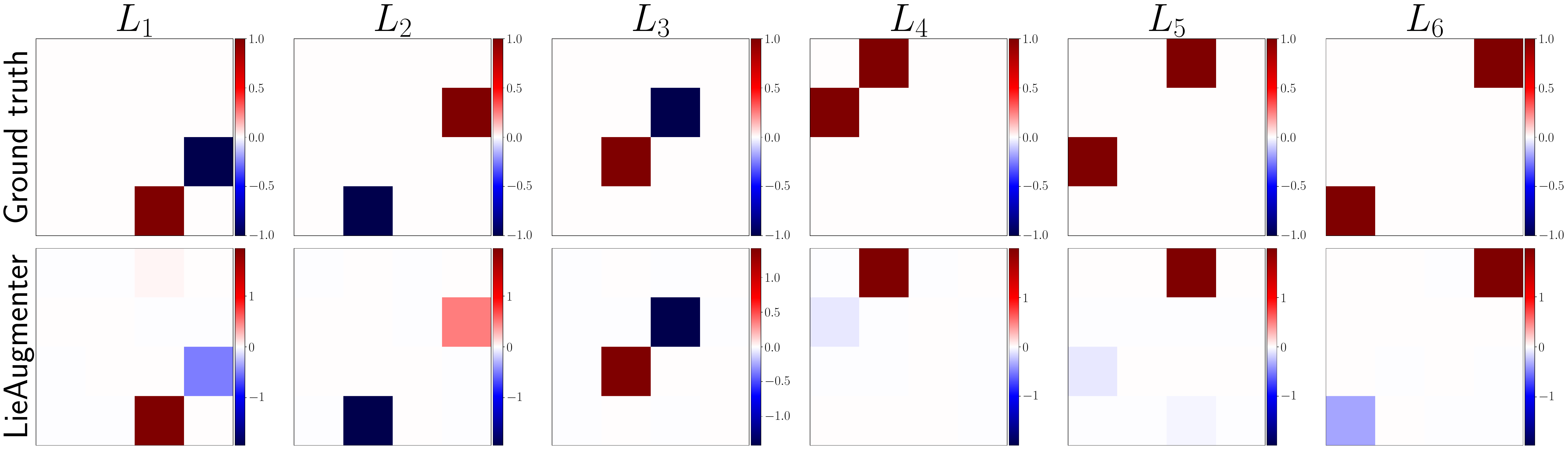}
    \caption{Comparison of canonical ground truth Lie algebra basis for $\mathrm{SO}^+(1,3)$ group (\textit{top}) and the generators learned by LieAugmenter (\textit{bottom}) on the QFT amplitude regression task. 
    For interpretability, we match the learned generators to the closest canonical basis.
  }
  \label{fig:lie_algebras_qft}
\end{figure}

We train LieAugmenter using the baseline Transformer model defined in \citet{spinner2024lorentz} as the prediction network, and with a learnable Lie algebra basis of cardinality $C=6$ corresponding to that of the restricted Lorentz group $\text{SO}^+(1,3)$, i.e., the connected identity component of the Lorentz group. 
From Figure~\ref{fig:lie_algebras_qft}, we can observe that LieAugmenter is able to approximately learn the correct group. 
Even though some of the elements of the learned Lie algebra basis (especially those corresponding to boosts) are not exactly equivalent to their ground truth counterparts, we believe that this is more of a side-effect caused by the symmetries that exist in this particular dataset rather than a partial imprecision of our method.

Interestingly, that observation of certain generators of the group being more strongly represented in the data aligns with the particular characteristics of many LHC applications, where often symmetry is only realized to a subset of the full Lorentz group \citep{favaro2025lorentz}.
In particular, detector effects and the beam axis direction reduce the symmetry group to $\mathrm{SO}(2)_{\text{beam}}$, i.e., rotations around the beam axis.
For the specific dataset we consider, the beam for the collision is aligned to the z-axis, so we could expect the generator of rotations on the xy plane to be the one most strongly recovered, and we observe that to be the case in Figure~\ref{fig:lie_algebras_qft}. 

\begin{figure}[h]
  \centering
  \includegraphics[width=0.5\textwidth]{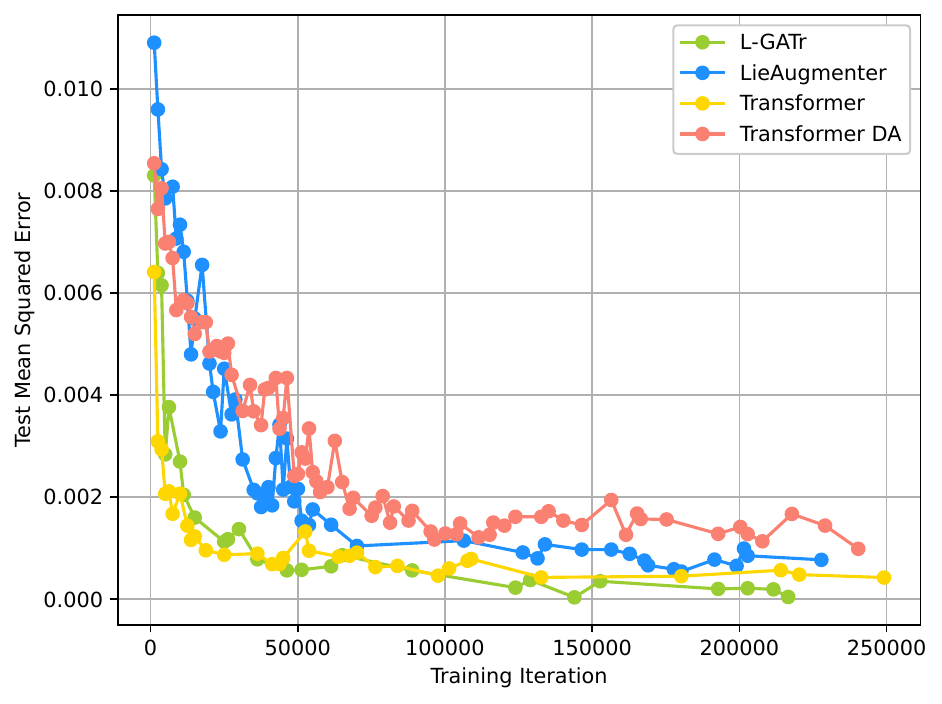}
  \caption{Test MSE during training for the equivariant model L-GATr, a Transformer prediction network trained with and without data augmentation as well using our LieAugmenter method in the QFT amplitude regression dataset.}
  \label{fig:mse_qft}
\end{figure}

Due to the existence of those symmetry breaking effects in the dataset, a model trained with data augmentation from the full $\mathrm{SO}^+(1,3)$ group would not generally be able to achieve optimal performance in this task. As a result, the flexibility and task-adaptiveness of LieAugmenter allows for a more effective data augmentation approach in this scenario. 
We verify this expectation through the prediction performance shown in Figure~\ref{fig:mse_qft}, where we compare L-GATr \citep{spinner2024lorentz}, which is equivariant to the full Lorentz group, with Transformer prediction networks trained with and without data augmentation from the $\mathrm{SO}^+(1,3)$ group, as well as using our LieAugmenter approach.
As we can observe, the performance of LieAugmenter is indeed better than that of the Transformer trained with data augmentation, and it is close to the one obtained with the Transformer without data augmentation and L-GATr despite it simultaneously discovering the symmetry group.

\subsection{Illustration of Augmented Samples}

\begin{figure}[h]
  \centering
  \includegraphics[width=0.9\textwidth]{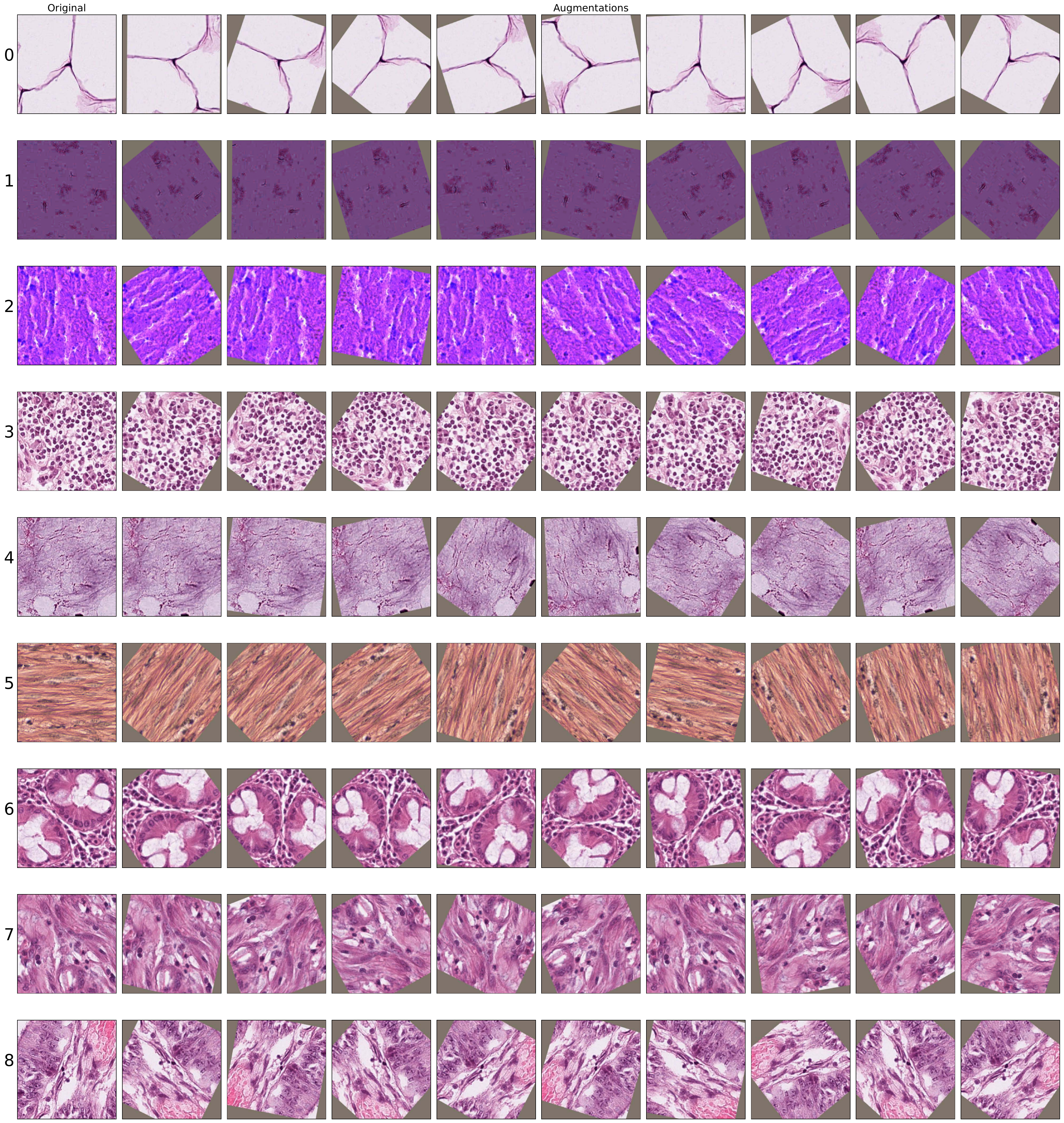}
  \caption{Example of augmentations generated by LieAugmenter for the CRC dataset. The first column shows the original samples, and the subsequent ones correspond to the generated augmentations.}
  \label{fig:augmentations_crc}
\end{figure}

\begin{minipage}{0.43\textwidth}
In order to provide increased interpretability into the way the symmetries learned by LieAugmenter act on the data in order to create the augmentations, we present a visualization of augmented samples generated by our model for RotatedMNIST (Figure~\ref{fig:augmentations_rotmnist}), $N$-body dynamics (Figure~\ref{fig:augmentations_nbody2}), and CRC (Figure~\ref{fig:augmentations_crc}).

\vspace{2em}

 \begin{center}
  \includegraphics[width=\linewidth, height=20em]{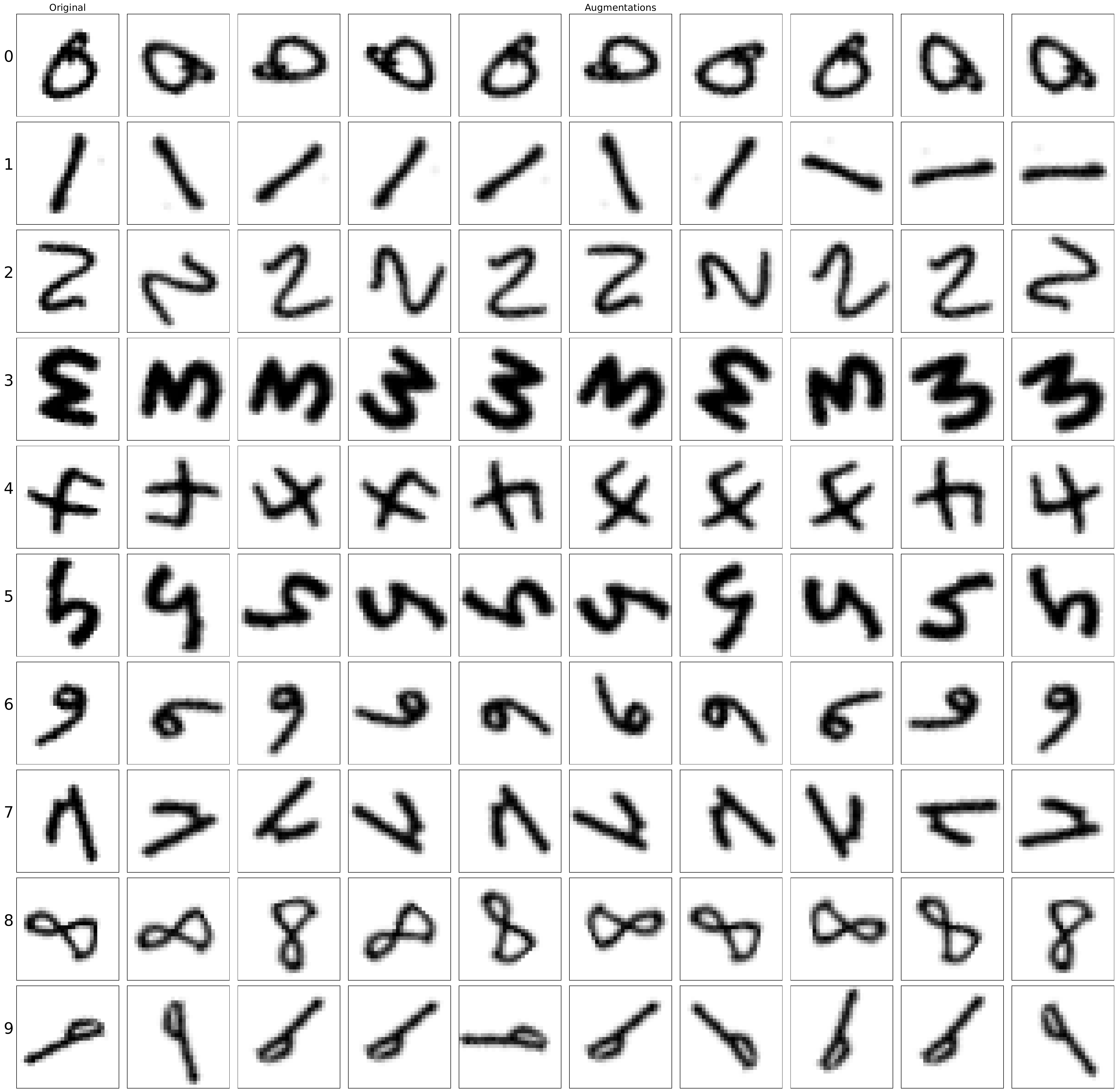}
  \captionof{figure}{Example of augmentations generated by LieAugmenter for the RotatedMNIST dataset. The first column shows the original samples, and the subsequent ones correspond to the generated augmentations.}
  \label{fig:augmentations_rotmnist}
 \end{center}
\end{minipage}
\hspace{0.5em}
\begin{minipage}{0.55\textwidth}
  \centering
  \includegraphics[width=\linewidth]{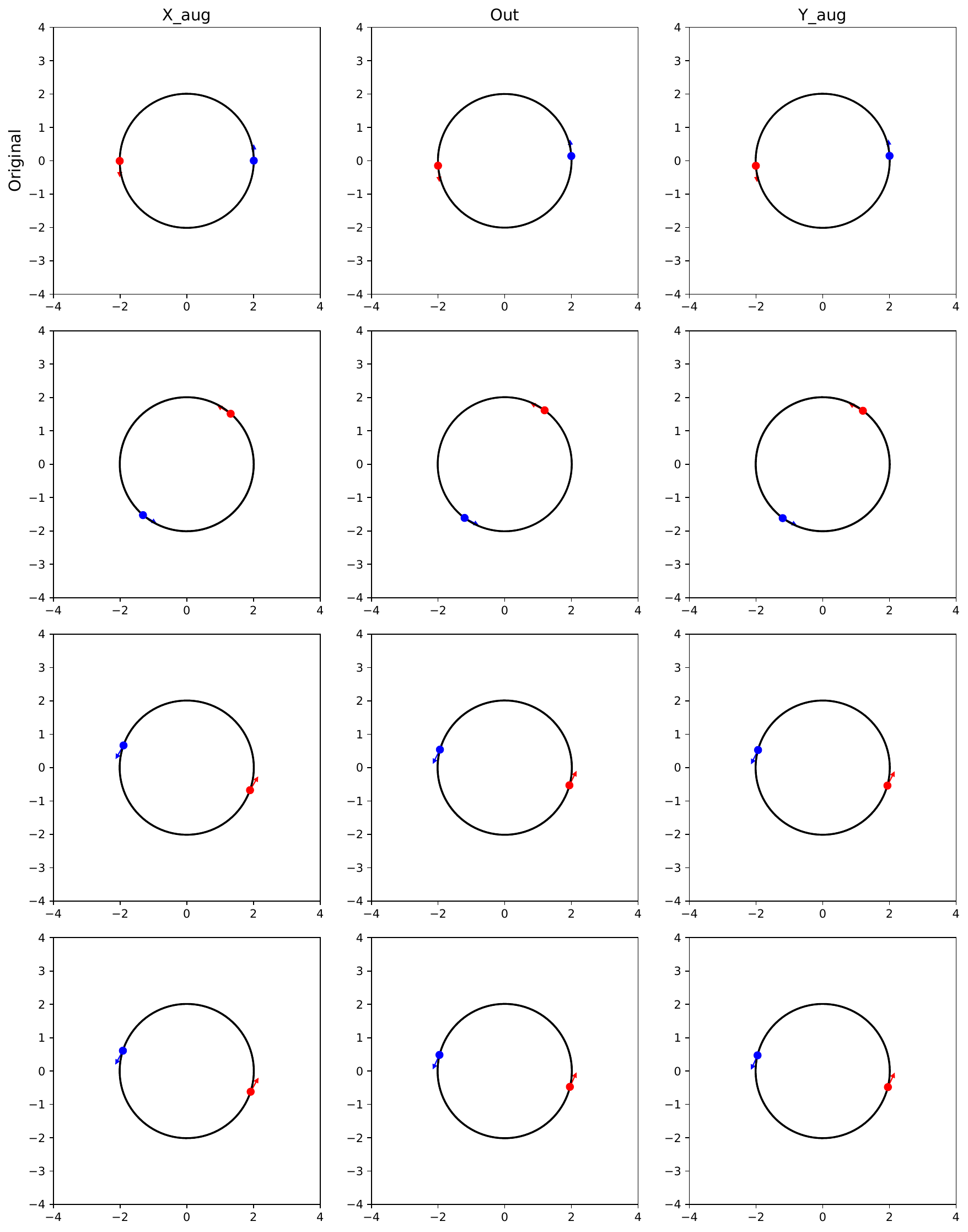}
  \captionof{figure}{Example of augmentations generated by LieAugmenter for the $2$-body dataset. The top row shows (from left to right) the original input sample, the output of the model for that sample, and the target prediction. The rows below show different augmented versions.}
  \label{fig:augmentations_nbody2}
\end{minipage}

\subsection{Runtime Analysis}

Next, we complement the analysis of the predictive performance in the different datasets that we considered by providing measurements of the train and test time associated with each model's prediction results. 

Although these measurements can be slightly influenced by the concurrent allocations of the GPUs we used for their execution, they exemplify the potential computational impact of our approach with respect to the different baselines.
In addition, note that we have not attempted to optimize our implementation in detail, so it could still potentially be made more efficient.

Importantly, the time measurements for the prediction model that uses LieGAN augmentations corresponds to only the training and inference of the prediction network itself, and, thus, it would be considerably slower if we incorporated the additional time required to train LieGAN itself. 
That additional runtime for two-step equivariant prediction approaches can pose
a significant limitation in practice, and it is one of the motivations behind our proposal of a single end-to-end method.

We present the runtime measurements for RotatedMNIST in Tables~\ref{tab:time_rotmnist} and \ref{tab:time_rotmnist_alt_inf}, for $N$-body dynamics in Tables~\ref{tab:times_nbody2} and \ref{tab:times_nbody2_t3}, for QM9 in Table~\ref{tab:results_qm9_app}, for the no-symmetry experiment in Table~\ref{tab:time_nosymmetry}, for the discrete rotation dataset in Table~\ref{tab:results_discrete_rotation_app}, and for partial permutation in Table~\ref{tab:results_partial_permutation_app}. 

From these results we can see the generally longer runtime of LieAugmenter over other augmentation baselines, which is mainly due to its additional loss and trainable parameters.
In contrast, LieAugmenter can be trained much faster than hard equivariant models that consider a priori known symmetries, e.g., EMLP. 
Overall, we generally observe a trade-off between the flexibility and performance of the approaches and the length of their training time.
This trade-off could be potentially minimized in the case of our approach, however, which more careful hyperparameter tuning and a more optimized implementation.

\begin{table*}[h]
\caption{RotatedMNIST runtime measurements (MM:SS) for in-distribution (ID) and out-of-distribution (OOD) evaluation of the different approaches. We report the mean over three random seeds.}

  \label{tab:time_rotmnist}
  \centering
  \resizebox{0.9\textwidth}{!}{%
  \addtolength{\tabcolsep}{-0.3em}
  \begin{tabular}{lcccccc}
    \toprule
    & \multicolumn{1}{c}{Equivariant architecture} & \multicolumn{1}{c}{No augmentation} & \multicolumn{1}{c}{Oracle augmentation} & \multicolumn{3}{c}{Symmetry discovery} \\
    \cmidrule(lr){2-2}\cmidrule(lr){3-3}\cmidrule(lr){4-4}\cmidrule(lr){5-7}
    & GCNN ($p4$) & Trivial & Rotation & Augerino+ & LieGAN Aug. & LieAugmenter \\
    \midrule
    ID train time  & $03:26$                  & $01:52$                 & $06:55$                          & $04:30$                       & $06:15$                              & $07:47$                     \\ 
    ID test time  & $00:00$                  & $00:00$                 & $00:02$                          & $00:03$                       & $00:01$                              & $00:01$                     \\ 
    \midrule
    OOD train time  & $03:56$                  & $01:50$                 & $06:57$                          & $04:47$                       & $05:03$                              & $07:04$                     \\ 
    OOD test time   & $00:00$                  & $00:00$                 & $00:02$                          & $00:03$                       & $00:01$                              & $00:01$                     \\
    \bottomrule
  \end{tabular}
  }
\end{table*}

\begin{table*}[h]
\caption{RotatedMNIST runtime measurements (MM:SS) for in-distribution (ID) and out-of-distribution (OOD) evaluation using the alternative inference strategy described in Appendix~\ref{app:Exploring_Alternative_Inference_Strategy}. We report the mean over three random seeds.}

  \label{tab:time_rotmnist_alt_inf}
  \centering
  \resizebox{0.6\textwidth}{!}{%
  \addtolength{\tabcolsep}{-0.3em}
\begin{tabular}{lccc} 
\toprule
               & Oracle augmentation & \multicolumn{2}{c}{Symmetry discovery}  \\ 
\cmidrule(lr){2-2}\cmidrule(l){3-4}
               & Rotation            & LieGAN Aug. & LieAugmenter              \\ 
\midrule
ID train time  & $03:03$             & $09:45$     & $10:44$                   \\
ID test time   & $00:00$             & $00:00$     & $00:00$                   \\ 
\midrule
OOD train time & $06:57$             & $06:18$     & $10:48$                   \\
OOD test time  & $00:00$             & $00:00$     & $00:00$                   \\
\bottomrule
\end{tabular}
  }

\end{table*}

\begin{table}[h!]
\caption{$2$-body dynamics runtime measurements (MM:SS) for in-distribution (ID) and out-of-distribution (OOD) of the different approaches for one input and output timesteps. We report the mean over three random seeds.}
  \label{tab:times_nbody2}
  \centering
\resizebox{0.8\textwidth}{!}{%
\begin{tabular}{clcccc} 
\toprule
& & \multicolumn{2}{c}{In-distribution} & \multicolumn{2}{c}{Out-of-distribution}  \\ 
\cmidrule{3-4} \cmidrule{5-6}
&                  & Train time & Test time              & Train time & Test time                   \\ 
\midrule
\multirow{1}{*}{Equiv. architecture} & EMLP             & $21:34$    & $00:02$                & $21:13$    & $00:02$                     \\
\midrule
\multirow{2}{*}{\makecell{Data\\augmentation}}        & Trivial          & $02:00$    & $00:00$                & $02:01$    & $00:00$                     \\
& Oracle           & $02:48$    & $00:00$                & $03:10$    & $00:00$                     \\ 
\midrule
\multirow{9}{*}{\makecell{Symmetry\\discovery}}       & Augerino+\_0     & $05:25$    & $00:02$                & $05:25$    & $00:02$                     \\
& LieGAN\_0 aug    & $05:29$    & $00:00$                & $05:25$    & $00:00$                     \\
& LieAugmenter\_0  & $08:18$    & $00:00$                & $08:20$    & $00:00$                     \\ 
\cmidrule{2-6}
& Augerino+\_2     & $05:40$    & $00:02$                & $05:32$    & $00:02$                     \\
& LieGAN\_2 aug    & $05:19$    & $00:00$                & $05:15$    & $00:00$                     \\
& LieAugmenter\_2  & $08:28$    & $00:00$                & $08:26$    & $00:00$                     \\ 
\cmidrule{2-6}
& Augerino+\_4     & $05:33$    & $00:02$                & $05:30$    & $00:02$                     \\
& LieGAN\_4 aug    & $05:22$    & $00:00$                & $05:26$    & $00:00$                     \\
& LieAugmenter\_4  & $08:27$    & $00:00$                & $08:26$    & $00:00$                     \\
\bottomrule
\end{tabular}
}
\end{table}

\begin{table}[h]
\caption{$2$-body dynamics runtime measurements (MM:SS) for in-distribution (ID) and out-of-distribution (OOD) of the different approaches for three input and output timesteps. We report the mean over three random seeds.}
  \label{tab:times_nbody2_t3}
  \centering
\resizebox{0.8\textwidth}{!}{%
\begin{tabular}{clcccc} 
\toprule
&                  & \multicolumn{2}{c}{In-distribution} & \multicolumn{2}{c}{Out-of-distribution}  \\ 
\cmidrule{3-6}
&                  & Train time & Test time              & Train time & Test time                   \\ 
\midrule
\multirow{1}{*}{\makecell{Equiv. architecture}} & EMLP             & $17:32$    & $00:02$                & $16:55$    & $00:02$                     \\
\midrule
\multirow{2}{*}{\makecell{Data\\augmentation}}        & Trivial          & $01:36$    & $00:00$                & $01:32$    & $00:00$                     \\
& Oracle           & $02:34$    & $00:00$                & $02:33$    & $00:00$                     \\ 
\midrule
\multirow{9}{*}{\makecell{Symmetry\\discovery}}       & Augerino+\_0     & $04:20$    & $00:01$                & $04:20$    & $00:01$                     \\
& LieGAN\_0 aug    & $04:22$    & $00:00$                & $04:21$    & $00:00$                     \\
& LieAugmenter\_0  & $06:37$    & $00:00$                & $06:35$    & $00:00$                     \\ 
\cmidrule{2-6}
& Augerino+\_2     & $04:20$    & $00:01$                & $04:23$    & $00:01$                     \\
& LieGAN\_2 aug    & $04:16$    & $00:00$                & $04:16$    & $00:00$                     \\
& LieAugmenter\_2  & $06:43$    & $00:00$                & $06:40$    & $00:00$                     \\ 
\cmidrule{2-6}
& Augerino+\_4     & $04:21$    & $00:01$                & $04:25$    & $00:01$                     \\
& LieGAN\_4 aug    & $04:17$    & $00:00$                & $04:13$    & $00:00$                     \\
& LieAugmenter\_4  & $06:46$    & $00:00$                & $06:48$    & $00:00$                     \\
\bottomrule
\end{tabular}
}
\end{table}

\begin{table}[h]
    \centering
\caption{Runtime measurements (MM:SS) for the prediction results in the dataset without continuous symmetries. We report the mean over three random seeds.}

  \label{tab:time_nosymmetry}
\begin{tabular}{llcc}
\toprule
&& Train time & Test time               \\ 
\midrule
\multirow{2}{*}[-0.15em]{Data augmentation} & No augmentation & $00:44$    & $00:00$                 \\
& Oracle augmentation & $01:12$    & $00:00$                 \\
\midrule
\multirow{3}{*}[-0.15em]{\makecell{Symmetry discovery}} & Augerino+    & $02:00$    & $00:01$                 \\
& LieGAN augmentation & $03:28$ & $00:00$ \\
& LieAugmenter & $03:06$    & $00:00$                 \\
\bottomrule
\end{tabular}
\end{table}

\begin{table}[h]
\centering
\caption{Train and test time (HH:MM:SS) of LieAugmenter for the different considered prediction targets of the QM9 dataset.} 
\label{tab:results_qm9_app}
\begin{tabular}{lcccc} 
\toprule
             & \multicolumn{2}{c}{HOMO} & \multicolumn{2}{c}{LUMO}  \\ 
\cmidrule(l){2-5}
             & Train time & Test time   & Train time & Test time    \\ 
\midrule
LieAugmenter & $15:17:12$ & $00:00:04$  & $15:19:15$ & $00:00:04$   \\
\bottomrule
\end{tabular}
\end{table}

\begin{table}[h]
\centering
\caption{Discrete rotation runtimes (MM:SS) for the different prediction approaches. We report the mean over three random seeds.}
\label{tab:results_discrete_rotation_app}
\begin{tabular}{llcc}
\toprule
&& Train time & Test time               \\ 
\midrule
\multirow{2}{*}[-0.15em]{Data augmentation} & No augmentation & $00:03:31$ & $00:00:00$              \\ 
& Oracle augmentation                & $00:05:35$ & $00:00:00$              \\ 
\midrule
\multirow{4}{*}[-0.15em]{\makecell{Symmetry discovery}} &Augerino+                 & $00:09:29$ & $00:00:01$              \\ 
&LieGAN aug                & $00:09:21$ & $00:00:00$              \\ 
&LieAugmenter (continuous) & $00:14:24$ & $00:00:00$              \\ 
&LieAugmenter (discrete)   & $00:14:51$ & $00:00:00$              \\
\bottomrule
\end{tabular}
\end{table}

\clearpage

\begin{table}[h]
\centering
\caption{Partial permutation runtimes (MM:SS) for the different prediction approaches. We report the mean over three random seeds.}
  \label{tab:results_partial_permutation_app}
\begin{tabular}{llcc}
\toprule
&& Train time & Test time               \\ 
\midrule
\multirow{2}{*}[-0.15em]{Data augmentation} & No augmentation            & $00:03:40$ & $00:00:00$              \\ 
&Oracle augmentation                & $00:05:33$ & $00:00:00$              \\ 
\midrule
\multirow{4}{*}[-0.15em]{\makecell{Symmetry discovery}} & Augerino+                 & $00:09:55$ & $00:00:01$              \\ 
&LieGAN augmentation                & $00:09:54$ & $00:00:00$              \\ 
&LieAugmenter (continuous) & $00:15:09$ & $00:00:00$              \\ 
&LieAugmenter (discrete)   & $00:15:04$ & $00:00:00$              \\
\bottomrule
\end{tabular}
\end{table}

\subsection{Additional Equivariance Error Measure}
\label{app:equiv_error2}

Lastly, we expand our analysis of the equivariance error of the different prediction approaches by studying the behavior of another similar metric that differently measures the consistency of the generated predictions under augmentations from the ground truth symmetry group: 
$\frac{1}{N} \sum_{i=1}^N \frac{1}{K} \sum\limits_{j=1}^K \left\lVert \Psi(\rho_{\mathcal{X}}(g_j)(x_i)) - \rho_{\mathcal{Y}} (g_j) (\Psi(x_i)) \right\rVert$.

We present the associated results for RotatedMNIST in Tables~\ref{tab:equiv_err_rotmnist} and \ref{tab:equiv_err_rotmnist_alt_inf}, for $N$-body dynamics in Tables~\ref{tab:equiv_error_nbody} and \ref{tab:equiv_error_nbody_t3}, for QM9 in Table~\ref{tab:equiv_error_qm9}, for the discrete rotation dataset in Table~\ref{tab:equiv_error_discreterot}, and for partial permutation in Table~\ref{tab:equiv_error_partialperm}. 

As we can observe, the measurements for this metric are very close to ones we obtained with the previous equivariance error and follow similar trends in terms of which approaches provide a better result, thus reinforcing our previous analysis.
In particular, we can once more see that LieAugmenter always provides among the lowest equivariance error measurements showing its strong and robust ability to encode the relevant equivariance into the prediction network.

\begin{table*}[h]
\caption{Additional equivariance error measure in the RotatedMNIST dataset for the in-distribution (ID) and out-of-distribution (OOD) evaluation of the different approaches. All values are reported as $\mathrm{mean}\pm\mathrm{std}$ over three random seeds.}

  \label{tab:equiv_err_rotmnist}
  \centering
  \resizebox{0.9\textwidth}{!}{%
  \addtolength{\tabcolsep}{-0.3em}
  \begin{tabular}{lcccccc}
    \toprule
    & \multicolumn{1}{c}{Equivariant architecture} & \multicolumn{1}{c}{No augmentation} & \multicolumn{1}{c}{Oracle augmentation} & \multicolumn{3}{c}{Symmetry discovery} \\
    \cmidrule(lr){2-2}\cmidrule(lr){3-3}\cmidrule(lr){4-4}\cmidrule(lr){5-7}
    & GCNN ($p4$) & Trivial & Rotation & Augerino+ & LieGAN Aug. & LieAugmenter \\
    \midrule
ID Equiv. Error ($\downarrow$) & $\text{6.51} \scriptstyle{\pm \text{1.41}}$ & $\text{6.69} \scriptstyle{\pm \text{0.24}}$ & $\underline{\text{4.86}} \scriptstyle{\pm \text{0.36}}$    & $\text{6.35} \scriptstyle{\pm \text{0.98}}$ & $\text{6.12} \scriptstyle{\pm \text{0.25}}$        & $\textbf{\text{3.82}} \scriptstyle{\pm \text{0.38}}$  \\ 
\midrule
OOD Equiv. Error ($\downarrow$)       & $\text{7.16} \scriptstyle{\pm \text{1.52}}$ & $\text{8.74} \scriptstyle{\pm \text{0.61}}$ & $\underline{\text{4.75}} \scriptstyle{\pm \text{0.38}}$    & $\text{5.26} \scriptstyle{\pm \text{0.72}}$ & $\text{7.17} \scriptstyle{\pm \text{0.50}}$        & $\textbf{\text{4.12}} \scriptstyle{\pm \text{0.75}}$  \\
    \bottomrule
  \end{tabular}
  }
\end{table*}

\begin{table*}[h]
\caption{Additional equivariance error measure in the RotatedMNIST dataset for the in-distribution (ID) and out-of-distribution (OOD) evaluation of the different approaches with the alternative inference strategy. All values are reported as $\mathrm{mean}\pm\mathrm{std}$ over three random seeds.}

  \label{tab:equiv_err_rotmnist_alt_inf}
  \centering
  \resizebox{0.9\textwidth}{!}{%
  \addtolength{\tabcolsep}{-0.3em}
  \begin{tabular}{lcccccc}
    \toprule
    & \multicolumn{1}{c}{Equivariant architecture} & \multicolumn{1}{c}{No augmentation} & \multicolumn{1}{c}{Oracle augmentation} & \multicolumn{3}{c}{Symmetry discovery} \\
    \cmidrule(lr){2-2}\cmidrule(lr){3-3}\cmidrule(lr){4-4}\cmidrule(lr){5-7}
    & GCNN ($p4$) & Trivial & Rotation & Augerino+ & LieGAN Aug. & LieAugmenter \\
    \midrule
ID Equiv. Error ($\downarrow$) & $6.51 \scriptstyle{\pm 1.41}$ & $6.69 \scriptstyle{\pm 0.24}$ & $\underline{4.72} \scriptstyle{\pm 0.27}$    & $6.35 \scriptstyle{\pm 0.98}$ & $5.95 \scriptstyle{\pm 0.87}$        & $\mathbf{4.54} \scriptstyle{\pm 0.17}$  \\  
\midrule
OOD Equiv. Error ($\downarrow$)   & $7.16 \scriptstyle{\pm 1.52}$ & $8.74 \scriptstyle{\pm 0.61}$ & $\underline{4.96} \scriptstyle{\pm 0.13}$    & $5.26 \scriptstyle{\pm 0.72}$ & $8.35 \scriptstyle{\pm 0.97}$        & $\mathbf{4.29} \scriptstyle{\pm 0.56}$  \\
    \bottomrule
  \end{tabular}
  }
\end{table*}

\begin{table}[h]
\centering
\caption{$2$-body dynamics alternative equivariant error measurements for in-distribution (ID) and out-of-distribution (OOD) of the different approaches for one input and output timesteps. We report the mean over three random seeds.}
\label{tab:equiv_error_nbody}
\begin{tabular}{clcc} 
\toprule
&                  & In-distribution                                       & Out-of-distribution                                    \\ 
\cmidrule{3-4}
&                  & Equiv. Error  $(\downarrow)$                                      & Equiv. Error $(\downarrow)$                                              \\ 
\midrule
\multirow{1}{*}{\makecell{Equiv. architecture}} & EMLP             & $\text{8.08e-05} \scriptstyle{{\pm \text{1.76e-05}}}$ & $\text{8.02e-05} \scriptstyle{{\pm \text{1.72e-05}}}$  \\
\midrule
\multirow{2}{*}{\makecell{Data\\augmentation}}        & Trivial          & $\text{4.84e-02} \scriptstyle{{\pm \text{1.58e-02}}}$ & $\text{8.91e-02} \scriptstyle{{\pm \text{6.24e-03}}}$  \\
& Oracle           & $\text{6.35e-03} \scriptstyle{{\pm \text{1.44e-03}}}$ & $\text{6.53e-03} \scriptstyle{{\pm \text{1.25e-03}}}$  \\ 
\midrule
\multirow{9}{*}{\makecell{Symmetry\\discovery}}       & Augerino+\_0     & $\text{4.20e-01} \scriptstyle{{\pm \text{1.85e-03}}}$ & $\text{4.21e-01} \scriptstyle{{\pm \text{6.31e-03}}}$  \\
& LieGAN\_0 aug    & $\text{3.93e-02} \scriptstyle{{\pm \text{1.73e-03}}}$ & $\text{7.25e-02} \scriptstyle{{\pm \text{1.01e-02}}}$  \\
& LieAugmenter\_0  & $\text{5.75e-03} \scriptstyle{{\pm \text{2.26e-04}}}$ & $\text{4.53e-03} \scriptstyle{{\pm \text{2.52e-04}}}$  \\ 
\cmidrule{2-4}
& Augerino+\_2     & $\text{4.30e-01} \scriptstyle{{\pm \text{5.04e-02}}}$ & $\text{4.00e-01} \scriptstyle{{\pm \text{6.40e-02}}}$  \\
& LieGAN\_2 aug    & $\text{2.83e-02} \scriptstyle{{\pm \text{8.22e-03}}}$ & $\text{6.68e-02} \scriptstyle{{\pm \text{2.19e-03}}}$  \\
& LieAugmenter\_2  & $\text{5.34e-03} \scriptstyle{{\pm \text{1.35e-03}}}$ & $\text{9.61e-03} \scriptstyle{{\pm \text{4.93e-03}}}$  \\ 
\cmidrule{2-4}
& Augerino+\_4     & $\text{4.16e-01} \scriptstyle{{\pm \text{1.64e-03}}}$ & $\text{4.16e-01} \scriptstyle{{\pm \text{3.58e-05}}}$  \\
& LieGAN\_4 aug    & $\text{6.03e-03} \scriptstyle{{\pm \text{8.00e-04}}}$ & $\text{7.08e-03} \scriptstyle{{\pm \text{1.67e-03}}}$  \\
& LieAugmenter\_4  & $\text{6.17e-03} \scriptstyle{{\pm \text{1.61e-03}}}$ & $\text{5.79e-03} \scriptstyle{{\pm \text{7.28e-04}}}$  \\
\bottomrule
\end{tabular}
\end{table}

\begin{table}[h]
\centering
\caption{$2$-body dynamics alternative equivariant error measurements for in-distribution (ID) and out-of-distribution (OOD) of the different approaches for three input and output timesteps. We report the mean over three random seeds.}
\label{tab:equiv_error_nbody_t3}
\begin{tabular}{clcc} 
\toprule
&                  & In-distribution                                       & Out-of-distribution                                    \\ 
\cmidrule{3-4}
&                  & Equiv. Error    $(\downarrow)$                                         & Equiv. Error   $(\downarrow)$                                           \\ 
\midrule
\multirow{1}{*}{\makecell{Equiv. architecture}} & EMLP             & $\text{4.47e-05} \scriptstyle{{\pm \text{7.24e-06}}}$ & $\text{4.46e-05} \scriptstyle{{\pm \text{7.31e-06}}}$  \\
\midrule
\multirow{2}{*}{\makecell{Data\\augmentation}}        & Trivial          & $\text{7.17e-02} \scriptstyle{{\pm \text{1.53e-02}}}$ & $\text{8.36e-02} \scriptstyle{{\pm \text{3.20e-03}}}$  \\
& Oracle           & $\text{1.10e-02} \scriptstyle{{\pm \text{3.29e-03}}}$ & $\text{8.62e-03} \scriptstyle{{\pm \text{8.36e-04}}}$  \\ 
\midrule
\multirow{9}{*}{\makecell{Symmetry\\discovery}}       & Augerino+\_0     & $\text{4.22e-01} \scriptstyle{{\pm \text{3.07e-03}}}$ & $\text{4.22e-01} \scriptstyle{{\pm \text{8.34e-03}}}$  \\
& LieGAN\_0 aug    & $\text{7.20e-02} \scriptstyle{{\pm \text{1.13e-02}}}$ & $\text{9.55e-02} \scriptstyle{{\pm \text{1.45e-02}}}$  \\
& LieAugmenter\_0  & $\text{7.62e-03} \scriptstyle{{\pm \text{1.95e-03}}}$ & $\text{1.04e-02} \scriptstyle{{\pm \text{4.38e-03}}}$  \\ 
\cmidrule{2-4}
& Augerino+\_2     & $\text{4.13e-01} \scriptstyle{{\pm \text{1.26e-02}}}$ & $\text{4.13e-01} \scriptstyle{{\pm \text{1.53e-03}}}$  \\
& LieGAN\_2 aug    & $\text{4.36e-02} \scriptstyle{{\pm \text{2.40e-03}}}$ & $\text{9.53e-02} \scriptstyle{{\pm \text{1.06e-02}}}$  \\
& LieAugmenter\_2  & $\text{8.50e-03} \scriptstyle{{\pm \text{8.16e-04}}}$ & $\text{2.13e-02} \scriptstyle{{\pm \text{2.66e-03}}}$  \\ 
\cmidrule{2-4}
& Augerino+\_4     & $\text{4.19e-01} \scriptstyle{{\pm \text{4.20e-03}}}$ & $\text{4.20e-01} \scriptstyle{{\pm \text{3.19e-03}}}$  \\
& LieGAN\_4 aug    & $\text{8.10e-03} \scriptstyle{{\pm \text{1.21e-03}}}$ & $\text{9.62e-03} \scriptstyle{{\pm \text{3.32e-03}}}$  \\
& LieAugmenter\_4  & $\text{8.37e-03} \scriptstyle{{\pm \text{9.29e-04}}}$ & $\text{8.11e-03} \scriptstyle{{\pm \text{3.96e-04}}}$  \\
\bottomrule
\end{tabular}
\end{table}

\begin{table}[h]
\centering
\caption{Alternative equivariant error measurements of LieAugmenter for the different considered prediction targets of the QM9 dataset.}
\label{tab:equiv_error_qm9}
\begin{tabular}{lcc} 
\toprule
             & HOMO          & LUMO           \\ 
\cmidrule(l){2-3}
             & Equiv. Error $(\downarrow)$ & Equiv. Error $(\downarrow)$ \\ 
\midrule
LieAugmenter & 0.0045        & 0.0019         \\
\bottomrule
\end{tabular}
\end{table}

\begin{table}[h]
\centering
\caption{Alternative equivariant error measurements for the discrete rotation dataset. We report the mean over three random seeds.}
\label{tab:equiv_error_discreterot}
\begin{tabular}{llc}
\toprule
              & & Equiv. Error   $(\downarrow)$                                       \\ 
\midrule
\multirow{2}{*}[-0.15em]{Data augmentation} & No augmentation & $\text{1.24e-01} \scriptstyle{{\pm \text{2.39e-03}}}$  \\ 
& Oracle augmentation                & $\textbf{\text{2.91e-03}} \scriptstyle{{\pm \text{2.35e-04}}}$  \\ 
\midrule
\multirow{4}{*}[-0.15em]{\makecell{Symmetry discovery}} & Augerino+                 & $\text{1.36e-02} \scriptstyle{{\pm \text{5.32e-03}}}$  \\ 
& LieGAN augmentation                & $\text{4.21e-03} \scriptstyle{{\pm \text{4.20e-04}}}$  \\ 
&LieAugmenter (continuous) & $\underline{\text{3.18e-03}} \scriptstyle{{\pm \text{4.07e-04}}}$  \\ 
&LieAugmenter (discrete)   & $\text{4.41e-02} \scriptstyle{{\pm \text{2.06e-03}}}$  \\
\bottomrule
\end{tabular}
\end{table}

\begin{table}[h]
\centering
\caption{Alternative equivariant error measurements for the partial permutation dataset. We report the mean over three random seeds.}
\label{tab:equiv_error_partialperm}
\begin{tabular}{llc}
\toprule
              & & Equiv. Error   $(\downarrow)$                                       \\ 
\toprule
\multirow{2}{*}[-0.15em]{Data augmentation} & No augmentation                 & $\text{1.58e+01} \scriptstyle{{\pm \text{4.25e+00}}}$  \\ 
& Oracle augmentation              & $\text{1.80e+01} \scriptstyle{{\pm \text{2.68e+00}}}$  \\ 
\midrule
\multirow{4}{*}[-0.15em]{\makecell{Symmetry discovery}} & Augerino+                 & $\text{3.20e+01} \scriptstyle{{\pm \text{5.35e+00}}}$  \\ 
&LieGAN augmentation                & $\textbf{\text{1.34e+01}} \scriptstyle{{\pm \text{2.59e+00}}}$  \\ 
&LieAugmenter (continuous) & $\text{1.75e+01} \scriptstyle{{\pm \text{5.80e+00}}}$  \\ 
&LieAugmenter (discrete)   & $\underline{\text{1.49e+01}} \scriptstyle{{\pm \text{3.45e+00}}}$  \\
\bottomrule
\end{tabular}
\end{table}

\end{document}